\pdfoutput=1
\documentclass[11pt]{article}

\usepackage[T1]{fontenc}
\usepackage[utf8]{inputenc}
\usepackage{lmodern}
\usepackage[margin=1in]{geometry}
\usepackage{microtype}
\usepackage{amsmath,amssymb,amsfonts,amsthm}
\usepackage{graphicx}
\usepackage{booktabs}
\usepackage{xcolor}
\usepackage[round,authoryear]{natbib}
\usepackage{authblk}

\usepackage{amsmath,amssymb,amsfonts,bm}
\usepackage{algorithmic}
\usepackage{graphicx}

\def\x{{\mathbf{x}}}
\def\bb{{\mathbf{b}}}
\def\tt{{\mathbf{t}}}
\def\BB{{\mathbf{B}}}
\def\A{{\mathbf{A}}}

\def\f{{\mathbf{f}}}

\def\s{{\mathbf{s}}}
\def\h{{\mathbf{h}}}

\def\p{{\mathbf{p}}}

\newcommand{\R}{\mathbb{R}}

\DeclareMathOperator*{\argmin}{argmin}

\usepackage{hyperref}
\definecolor{mydarkgreen}{rgb}{0,0.6,0.30}
\definecolor{mydarkblue}{rgb}{0,0.30,0.65}
\hypersetup{
    colorlinks=true,    %
    linkcolor=mydarkblue,     %
    citecolor=mydarkgreen,    %
    urlcolor=mydarkblue       %
}

\usepackage{url}
\usepackage{enumitem}
\usepackage{subcaption}
\DeclareCaptionFont{sevenpt}{\fontsize{7}{8.5}\selectfont}
\usepackage{wrapfig}
\usepackage{multirow}

\usepackage{tikz}
\usetikzlibrary{positioning,calc,arrows,backgrounds,fit,shapes.geometric,decorations.pathreplacing,spy}
\definecolor{gtfill}{HTML}{E7A6A6}
\definecolor{gtline}{HTML}{A85252}
\definecolor{trfill}{HTML}{F7E6E6}
\definecolor{trline}{HTML}{7A2E2E}
\definecolor{pefill}{HTML}{BFDBEF}
\definecolor{peline}{HTML}{2E6E9E}
\definecolor{rbox}{HTML}{E11010}
\definecolor{ssstroke}{HTML}{555555}

\title{TomoTransformer: Towards a Foundation Model for CT Reconstruction}

\author{AmirEhsan Khorashadizadeh\thanks{Swiss Data Science Center (SDSC) in Paul Scherrer Institute (PSI), Villigen, Switzerland.\newline Email: \texttt{amirehsan.khorashadizadeh@psi.ch}}\hspace{0.5em}and Benjam\'in B\'ejar\thanks{Swiss Data Science Center (SDSC) in Paul Scherrer Institute (PSI), Villigen, Switzerland.\newline Email: \texttt{benjamin.bejar@psi.ch}}}

\date{}

\newcommand{\rev}[1]{{#1}}
\newenvironment{revblock}{}{}
\newcommand{\arxivnew}[1]{{#1}}

\begin{document}

\maketitle

\begin{abstract}
Supervised deep learning has advanced sparse-view tomographic reconstruction. However, conventional models, which typically map filtered back-projection (FBP) images or sinograms to clean reconstructions, are brittle under distribution shifts. Because they require retraining whenever projection counts and angles, detector resolutions, or data distributions change, their deployment in real-world applications remains limited. To address this, we introduce TomoTransformer, a transformer-based architecture that treats each \textit{local} filtered projection as an individual token and predicts missing views via self-attention. Crucially, TomoTransformer operates in a \emph{back-projection space} that separates projections across spatial locations, making view interpolation geometrically well-posed and invariant to detector size. This design yields a single foundation model that can process any number of input projections, at arbitrary angular locations and detector dimensions, and query any number of target angles without retraining. Trained on a large-scale dataset spanning diverse medical CT anatomies and natural images, TomoTransformer generalizes effectively across anatomies, materials, and resolutions. Extensive evaluations on several benchmark sparse-view datasets show that TomoTransformer significantly outperforms concurrent multi-purpose models like ViewTrans while remaining fully agnostic to the number of input and target projections. \arxivnew{We further introduce ray-consistent stitching, an inference procedure that combines the predictions of all patches crossed by the same ray into a single filtered projection, enforcing global consistency over the entire image without any retraining. Because TomoTransformer predicts projections, it can also be adapted to a new scan by self-supervised fine-tuning on the measured projections alone, without ground truth.} Finally, TomoTransformer demonstrates strong generalization to real experimental nanoscale brain and chip data acquired at an X-ray synchrotron, showcasing its practical utility for real-world applications.

\end{abstract}

\section{Introduction}
\label{sec:introduction}

Computed tomography (CT) \citep{kak2001principles} is among the most widely used imaging modalities for volumetric reconstruction, with many applications in biology \citep{eisenstein2023parallel,bosch2025nondestructive}, material science \citep{holler2017high}, and medicine \citep{de2014industrial}, to name a few. The objective of tomographic imaging is to reconstruct a 3D volume from a set of 2D projections acquired at different angles. Under dense angular sampling and noise-free conditions, filtered back-projection (FBP)~\citep{kak1988principles} is optimal and provides a closed-form inverse via the Fourier slice theorem. In practice, CT projections are acquired at discrete angular locations and are inherently affected by measurement noise. Constraints on radiation dose and scanning time often limit the number of acquired projections, motivating the even more challenging scenario of \emph{sparse-view} CT. In that setting, the incomplete noisy measurements make image reconstruction an ill-posed inverse problem, where FBP can produce streaking artifacts and lose structural and fine-scale details.

A large body of work has addressed sparse-view CT through model-based iterative reconstruction with hand-crafted priors, including total variation~\citep{sidky2008image} and wavelet sparsity, which partially suppress artifacts at the expense of iterative (slow) optimization. Over the past decade, supervised deep learning has emerged as the dominant alternative, providing faster inference and higher reconstruction quality. The common approach is post-processing of an initial estimate (e.g., an FBP reconstruction from a sparse sinogram) into a clean image~\citep{kang2017deep, adler2018learned, han2018framing, wu2018learn, wang2022dudotrans, yang2023attention, ayad2024qnmixer}.

Despite their empirical success, existing supervised methods commonly have three fundamental limitations: \emph{(i)} each model is trained for a specific acquisition geometry (i.e., a fixed projection count and angular range); \emph{(ii)} each model is designed for a fixed image and detector size; and \emph{(iii)} limited generalizability, a model is trained on a single, narrow data distribution (e.g., a single anatomical region). For instance, a network trained on 60-view sinograms cannot be deployed on a 90-view acquisition without retraining or architectural changes, and is similarly tied to the image, detector size, and data distribution used at training time. It is therefore, highly desirable to have a general model that could accommodate for and be robust to this variability. In line with the recent trend of developing foundations models, a single model trained at scale and adapted to a wide range of downstream tasks and conditions \citep{bommasani2021opportunities, brown2020language, kirillov2023segment, wasserthal2023totalsegmentator, he2025vista3d, liu2024seislm}, we advocate for a foundation model for CT reconstruction capable of ingesting arbitrary sparse acquisitions, with varying view counts, angular configurations, and detector resolutions. Recent attempts to partially achieve this goal have been made by ViewTrans~\citep{chen2026viewtrans}, which uses a transformer to predict missing views from raw sinograms. However, ViewTrans is still tied to a fixed detector size and image resolution, limiting its practical utility.

We propose \textbf{TomoTransformer}, a transformer-based architecture that treats each local filtered projection as a token and predicts missing views via self-attention. Unlike ViewTrans~\citep{chen2026viewtrans}, which operates on raw sinograms and is tied to a fixed detector size, TomoTransformer acts in a \emph{local disentangled} back-projection space. By decoupling projections across spatial patches, our approach renders view interpolation geometrically well-posed and invariant to detector dimensions. As a result, a single trained model maps any number of input projections of arbitrary detector resolution to any higher number, without retraining or architectural changes. We train this model on a large-scale dataset spanning diverse medical CT anatomies and natural images, and demonstrate its generalization by transferring the pretrained model zero-shot to nanoscale brain tomography. Our experiments show that TomoTransformer significantly outperforms ViewTrans~\citep{chen2026viewtrans} and
matches or exceeds strong single-purpose baselines that are specifically trained for each acquisition protocol, while remaining agnostic to the number of input and queried projections. Our contributions are as follows.
\begin{itemize}[leftmargin=1.5em]
    \item We introduce \textbf{TomoTransformer}, a CT reconstruction model that is simultaneously \emph{agnostic to the number and  angular configuration of projections, and detector size}, removing the protocol dependence that limits existing supervised methods.
    \item We formulate sinogram completion in a disentangled space which makes view interpolation geometrically natural, yielding a flexible and scalable architecture.
    \item We train TomoTransformer on a large-scale dataset of medical CT anatomies and natural images, showing that a single set of weights generalizes across anatomies, materials, and resolutions, including zero-shot transfer to real experimental data of nanoscale ptychography X-ray tomography acquired at Swiss Light Source (SLS), \rev{mouse brain sample~\citep{bosch2025nondestructive} and processor chip~\citep{aidukas2024high}.}
    \item We evaluate the robustness of TomoTransformer against noise where our experiments show that a TomoTransformer pre-trained on clean images achieves remarkable zero-shot denoising performance on unseen noisy projections without any fine-tuning, demonstrating its practical utility for real-world imaging tasks.
    \arxivnew{\item We propose ray-consistent stitching (RCS), an inference procedure that enforces global consistency over the entire image for free: every missing view is reconstructed as a single filtered projection shared by all patches crossed by the same ray, without any retraining. RCS effectively removes the blocking artifacts of patch-wise inference. Such global consistency is not achievable with standard patch-based models, whose patches are predicted and stitched independently in the image domain.}
    \rev{\item We develop a self-supervised test-time fine-tuning strategy that allows TomoTransformer to adapt to unseen data, further improving reconstruction quality on real experimental data.}
\end{itemize}

\section{Related work}
\label{sec: related works}

Convolutional neural networks (CNNs) have been the dominant architecture for learning-based tomographic reconstruction. FBPConvNet~\citep{jin2017deep} and tight-frame U-Net variants~\citep{han2018framing} established the U-Net~\citep{ronneberger2015unet} as a strong backbone, and subsequent work extended this idea with directional wavelets~\citep{kang2017deep}, attention gating~\citep{yang2023attention}, and dual-domain pipelines that jointly process sinograms and images~\citep{zhang2018sparse}. \rev{Sinogram-domain methods instead restore the missing views before reconstruction: SIPID~\citep{yuan2018sipid} combines a sinogram interpolation network with an image-domain denoiser, and the convolutional interpolation network (CIN) of \citet{guo2023efficient} predicts the missing views as a residual on their linear interpolation.} A parallel line of work unrolls iterative solvers into learnable networks~\citep{wu2018learn,adler2018learned}, embedding the Radon operator into the architecture and improving data consistency and model generalization. However, these methods are not scalable as they always need to process the whole image, as opposed to U-Net-like architectures that can process local patches. More recently, transformer-based reconstructors like DuDoTrans~\citep{wang2022dudotrans}, CTTR~\citep{shi2022ct}, ViewTrans~\citep{chen2026viewtrans}\rev{, MDST~\citep{li2023mdst}} or TD-STrans~\citep{chen2024tdstrans} replace convolutional backbones to capture the global structure of the sinogram that local convolutions miss\rev{; MDST and TD-STrans combine the sinogram and image domains with the projector inside the network and, like unrolled methods, process the whole image at once}. Despite these advances, all existing models remain tied to a fixed acquisition protocol and detector resolution, requiring full retraining whenever projection counts or angles change thus limiting their practical utility. \arxivnew{Recently, CryoLithe~\citep{kishore2025cryolithe} used a DeepSets architecture to handle a varying number of projections in cryo-electron tomography.}

A second class of methods casts reconstruction as posterior sampling under a learned generative prior. In this paradigm, a diffusion model is often trained exclusively on clean images, and the forward operator is incorporated at inference via data-consistency terms that guide reverse diffusion~\citep{chung2023dps, song2022solving, chung2023diffusionmbir}. Because the generative prior is decoupled from the forward operator, a single trained model can theoretically accommodate any acquisition protocol, a flexibility shared by our approach. However, this comes at the cost of iterative inference, requiring hundreds to thousands of score evaluations per reconstruction, and a prior specific to the imaged distribution and the detector size.

\section{Computed tomography}
\label{sec: computed_tomography}

We consider parallel-beam 2D computed tomography where the attenuation field $f(\x)$ is illuminated by parallel X-ray beams at different angles,
\begin{align}
    s_\theta(u) = \int_{-\infty}^\infty f\big(x_{\theta, u}(t), y_{\theta, u}(t) \big)\,dt,
    \label{eq:radon_def}
\end{align}
where $\x = (x_{\theta, u}(t), y_{\theta, u}(t))$ parameterizes the line at angle $\theta$ and distance $u$ from the origin,
\begin{align}
    x_{\theta, u}(t) = u\cos(\theta) - t\sin(\theta), \quad
    y_{\theta, u}(t) = u\sin(\theta) + t\cos(\theta).
\end{align}
In practice, measurements are acquired over a discrete set of $R$ projection angles $\{\theta_{r}\}_{r=1}^{R}$, each sampled at $M$ equispaced detector pixels: the $r$-th column of the \emph{sinogram} $\s \in \mathbb{R}^{M \times R}$ collects $s_{\theta_r}(\cdot)$ at those $M$ detector positions.

The filtered back-projection (FBP) is the most widely used method for tomographic reconstruction. According to the Fourier slice theorem \citep{kak1988principles}, we first filter each column of the sinogram along the detector dimension, $\s_r * \h$, where $*$ represents 1D convolution and $\h$ is a high-pass ramp kernel. We write $\tilde{\s}_r(\cdot)$ for the linear interpolation of this filtered column, so that it can be evaluated at any real detector coordinate. We then back-project the filtered projections by identifying the measurements associated with each pixel $(x,y)$ in the reconstruction domain, which are given by the sinusoidal trajectory $x\cos(\theta) + y\sin(\theta)$ in the sinogram as shown in Figure \ref{fig: tomotransformer},
\begin{align}
    \bb_{r}(x,y) \;=\; \tilde{\s}_r\big(x\cos(\theta_{r}) + y\sin(\theta_{r}) \big),
    \label{eq:single_view_bp}
\end{align}
evaluated on the $N \times N$ reconstruction grid to give the discrete single-view back-projection $\bb_{r} \in \mathbb{R}^{N \times N}$, where $N$ determines the dimension of the reconstructed image. The FBP reconstruction $\f^{\text{FBP}} \in \mathbb{R}^{N \times N}$ is simply the angular average over the filtered back-projections,
\begin{align}
    \f^\text{FBP} = \frac{1}{R}\sum_{r=1}^{R} \bb_r.
    \label{eq:FBP}
\end{align}

Equation~\eqref{eq:single_view_bp} is central to our TomoTransformer architecture, as it isolates the contribution of view $r$ alone at reconstruction coordinate $(x,y)$. Stacking these single-view back-projections across all $R$ angles yields a tensor
\begin{align}
    \BB = \big[\bb_{1}, \bb_{2}, \dots, \bb_{R}\big] \in \mathbb{R}^{N \times N \times R},
    \label{eq:disentangled_space}
\end{align}
which we refer to as the \emph{disentangled back-projection space}. In the next section, we establish our TomoTransformer architecture in this disentangled space.

\section{TomoTransformer: a flexible reconstruction model}
\label{sec: tomotransformers}

\begin{figure}[!h]
    \centering
    \begin{tikzpicture}
      \node[anchor=south west, inner sep=0pt] (tomofig) at (0,0)
            {\includegraphics[width=0.8\textwidth]{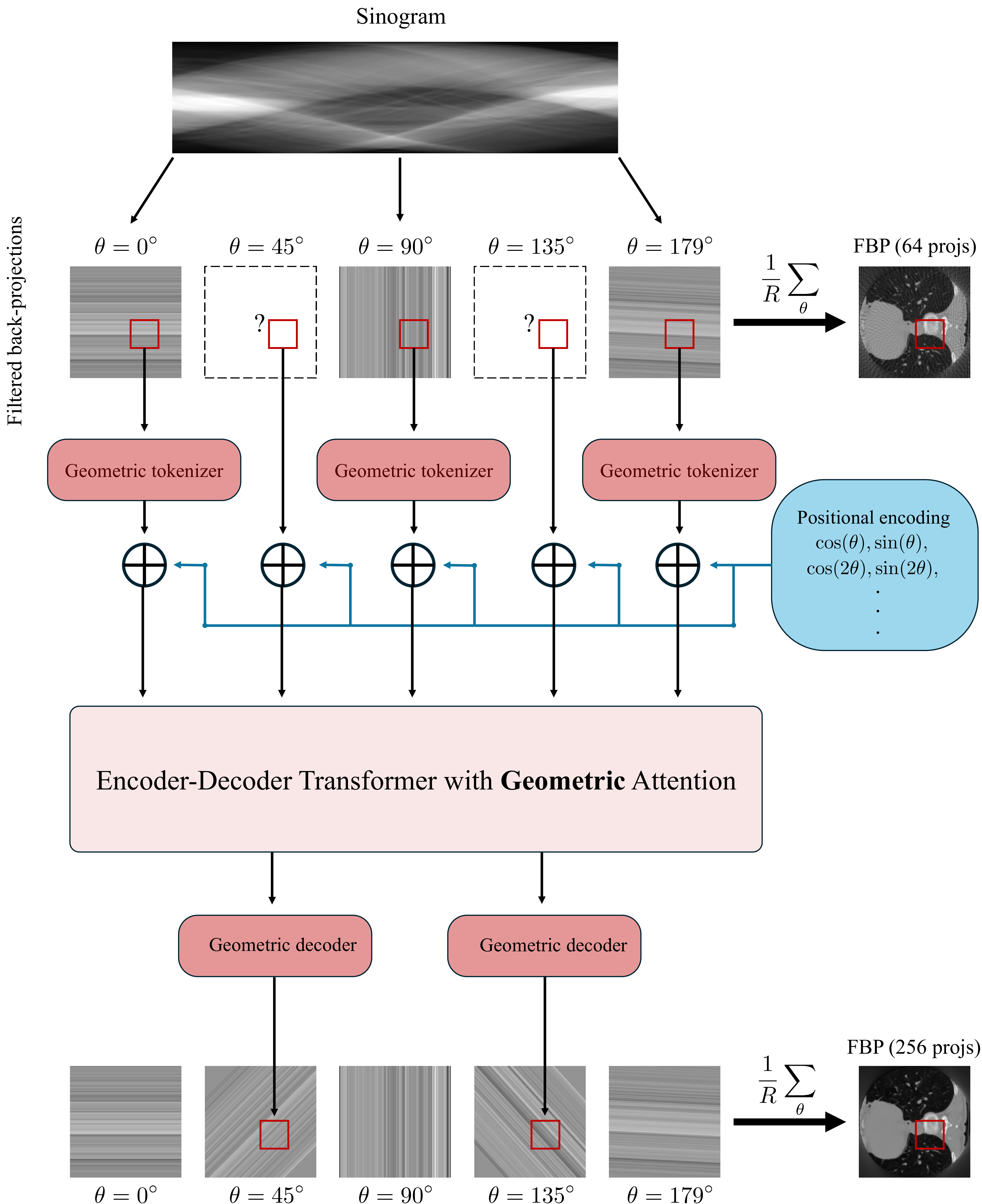}};
      \begin{scope}[x={(tomofig.south east)}, y={(tomofig.north west)}]
        \def\tomoX#1{0.175369+0.0025222*#1}
        \def\tomoS#1{0.918981-0.0099639*cos(#1)-0.0128352*sin(#1)}
        \def\tomoW#1{0.0115741*(abs(cos(#1))+abs(sin(#1)))}
        \fill[rbox, opacity=0.30]
          plot[domain=0:180,samples=181,variable=\t,smooth]
               ({\tomoX{\t}},{\tomoS{\t}+\tomoW{\t}})
          -- plot[domain=180:0,samples=181,variable=\t,smooth]
               ({\tomoX{\t}},{\tomoS{\t}-\tomoW{\t}}) -- cycle;
        \foreach \e in {+,-}{%
          \draw[rbox, line width=0.5pt, densely dashed]
            plot[domain=0:180,samples=181,variable=\t,smooth]
                 ({\tomoX{\t}},{\tomoS{\t}\e\tomoW{\t}});}
        \draw[white, line width=1.9pt, line cap=round]
          plot[domain=0:180,samples=181,variable=\t,smooth]
               ({\tomoX{\t}},{\tomoS{\t}});
        \draw[rbox, line width=0.9pt, line cap=round]
          plot[domain=0:180,samples=181,variable=\t,smooth]
               ({\tomoX{\t}},{\tomoS{\t}});
        \foreach \t in {0,45,90,135,179}{%
          \fill[white] ({\tomoX{\t}},{\tomoS{\t}}) circle (1.5pt);
          \fill[rbox]  ({\tomoX{\t}},{\tomoS{\t}}) circle (1.0pt);}
        \foreach \p in {(0.147181,0.722531), (0.288120,0.722531),
                        (0.421491,0.722531), (0.563375,0.722531),
                        (0.692017,0.722531), (0.946727,0.722531),
                        (0.279417,0.057562), (0.555429,0.057562),
                        (0.946727,0.057562)}{%
          \fill[rbox] \p circle (0.9pt);}
        \node[anchor=west, align=left, font=\scriptsize, inner sep=2pt, text=black]
              at (0.636,0.919)
              {sinogram footprint of the patch:\\%
               $x\cos\theta+y\sin\theta \pm \tfrac{P}{2}(|\cos\theta|+|\sin\theta|)$};
      \end{scope}
    \end{tikzpicture}
    \caption{TomoTransformer operates on the disentangled filtered back-projection space by treating each local filtered projection at angle $\theta$ as a token. It can process a varying number of projections in the input and generate an arbitrary number of local projections.}
    \label{fig: tomotransformer}
\end{figure}

A foundation model for sparse-view tomography must satisfy several core requirements that existing supervised methods satisfy only in isolation:
\textbf{(R1)} Agnosticism to the number and angular positions of input projections; \textbf{(R2)} Agnosticism to detector dimensions and volume resolution; \textbf{(R3)} Scalability to large images; \textbf{(R4)} Generalization across heterogeneous data distributions.

TomoTransformer fulfills all four requirements through three design choices: First, we formulate reconstruction in the \emph{disentangled} back-projection space \eqref{eq:disentangled_space}, casting unseen view prediction as a spatially aligned interpolation problem \textbf{(R1)}. Second, we treat each single-view back-projection as a \emph{token} conditioned on its viewing angle to be processed by a transformer. This design enables a single model to ingest and predict an arbitrary number of views at arbitrary angles and detector resolutions \textbf{(R1--R2)}. Third, we tokenize \emph{local} $P \times P$ patches rather than full back-projections, yielding a \rev{scalable pipeline for large images and volumes of arbitrary size} \textbf{(R3)}. Requirement \textbf{(R4)} is met by two factors: 1) working on local patches rather than full back-projections improves generalization by reducing the complexity of the prediction task and increasing the effective number of training samples \citep{khorashadizadeh2025glimpse, khorashadizadeh2025lofi} and 2) training the model at scale on large-scale heterogeneous datasets.

In this paper, we formulate the missing view prediction in the disentangled space \eqref{eq:disentangled_space}: predicting a missing projection at $\theta_\star$ amounts to predicting a new slice $\bb_\star$ from the observed slices $\{\bb_r\}_{r = 1}^R$. Because every slice lives on the same $N \times N$ image grid and shares the same coordinate system, the interpolation is spatially aligned across views, a property the raw sinogram does not enjoy. As shown in Figure~\ref{fig: tomotransformer}, each back-projection $\bb_r \in \R^{N \times N}$ is treated as a token conditioned on its projection angle $\theta_r$. Through self-attention, TomoTransformer can process a variable number of input tokens at arbitrary angles and predict an arbitrary number of queried views $\{\hat{\bb}_q\}_{q=1}^{Q}$ as output, without any change to the architecture.

TomoTransformer has an encoder-decoder architecture similar to masked autoencoders (MAE) \citep{he2022masked}. The encoder processes only the existing tokens, followed by the decoder, which takes the encoder's output for the existing tokens together with a learnable tensor for the queried tokens to be generated. Unlike standard transformers, token positions in our framework are physical quantities: the position of token $r$ is its projection angle $\theta_r$, a continuous coordinate rather than a discrete index. To represent the viewing geometry, we map each angle $\theta_r$ to Fourier positional embeddings $\bigl\{(\sin(k\theta_r),\cos(k\theta_r))\bigr\}_{k=1}^{F}$. Because these features vary smoothly with $\theta_r$, the embedding is defined at every real angle rather than only on a fixed grid, which is what enables the model to generalize to unseen target angles at test time. These embeddings encode each token's \emph{absolute} angle; the transformer blocks additionally condition attention on the \emph{relative} geometry, adding a learned function of the pairwise angular difference $\theta_r - \theta_{r'}$ to the attention logits. We refer to this as geometric attention and detail it in Appendix~\ref{sec: tomo_arch_details}.

Unlike naive sinogram interpolation, which must process the entire projection simultaneously, operating in the disentangled space enables localized computation: because each pixel's relevant measurements are spatially aligned, tokens do not need to span the full back-projection. As shown in Figure~\ref{fig: tomotransformer} (red boxes), we extract local patches $\tt_r \in \R^{P \times P}$ ($P \ll N$) from the full back-projections $\bb_r$ to serve as input tokens. Processing $P \times P$ patches rather than full $N \times N$ projections drastically reduces memory consumption and computational overhead during training, making the model scalable to large images. Since the network operates exclusively on fixed-size $P \times P$ tokens, it is decoupled from the full image dimension $N$ and the detector resolution. A single model can thus be trained on and reconstruct datasets at their native resolutions without resampling to a uniform grid. Furthermore, predicting a localized back-projection patch from observed views represents a much simpler task than predicting entire back-projections for a fixed set of views. Finally, the model can train on a large number of local patches extracted from each image, improving generalization and robustness to distribution shifts \citep{khorashadizadeh2025lofi}.

\subsection{Geometric tokenizer and decoder}
\label{sec: Geometric tokenizer}

The geometric tokenizer maps a local back-projection patch $\tt_r \in \R^{P \times P}$ to a low-dimensional embedding $\mathbf{z}_r \in \R^{D}$ by exploiting the geometric structure of single-view back-projections, and the geometric detokenizer reconstructs $\hat{\tt}_q \in \R^{P \times P}$ from the predicted embedding $\hat{\mathbf{z}}_q \in \R^{D}$.

\begin{wrapfigure}{r}{0.44\textwidth}
    \vspace{-\intextsep}
    \centering
    \setlength{\fboxsep}{0pt}\setlength{\fboxrule}{0.4pt}
    \begin{tabular}{@{}c@{\hspace{2pt}}c@{\hspace{2pt}}c@{}}
      $\theta=28^\circ$ & $\theta=90^\circ$ & $\theta=135^\circ$ \\[1pt]
      \fbox{\includegraphics[width=0.136\textwidth]{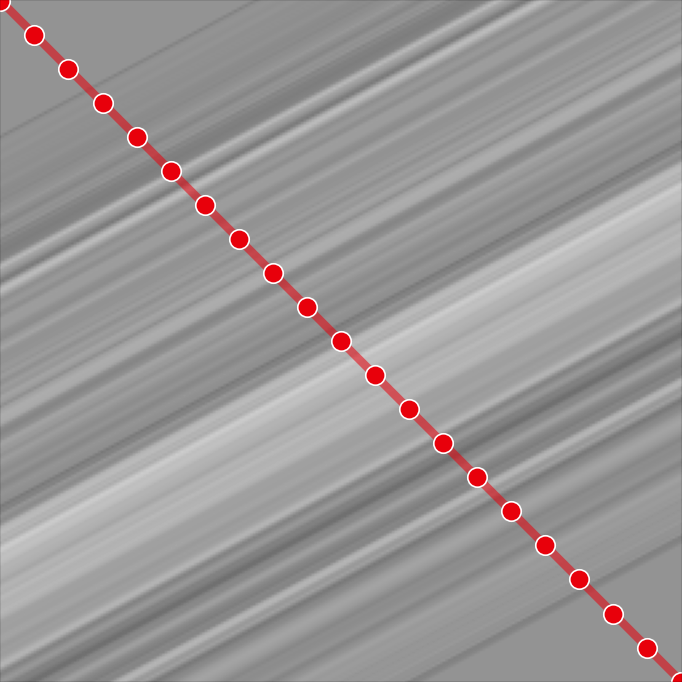}} &
      \fbox{\includegraphics[width=0.136\textwidth]{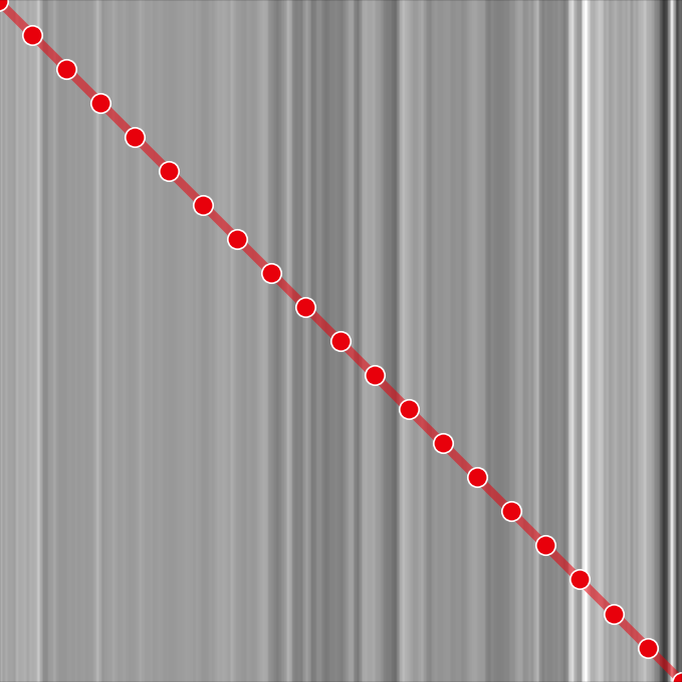}} &
      \fbox{\includegraphics[width=0.136\textwidth]{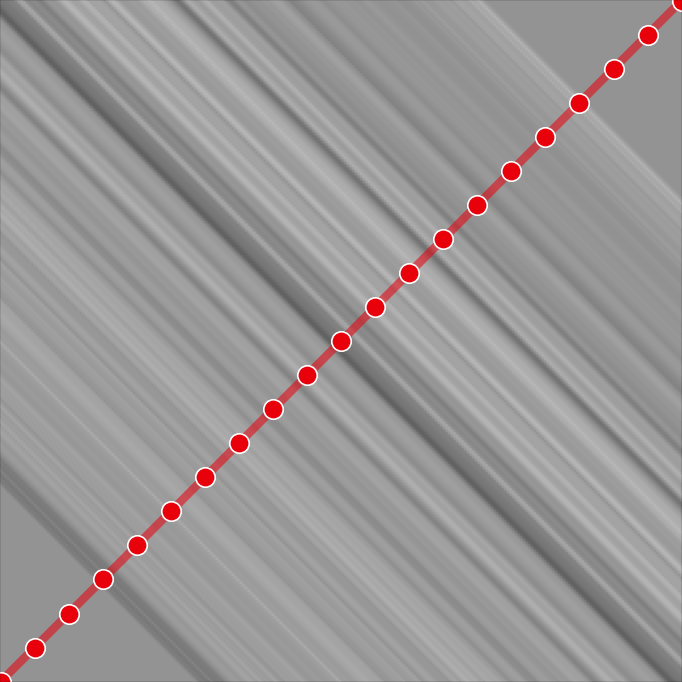}}
    \end{tabular}
    \caption{Geometric tokenizer}
    \vspace{-\intextsep}
    \label{fig: geometric_tokenizer}
\end{wrapfigure}
As illustrated in Figure~\ref{fig: geometric_tokenizer}, a single-view back-projection $\bb_r$ consists of parallel stripes perpendicular to the projection angle $\theta_r$. Any local patch $\tt_r \in \R^{P \times P}$ cropped from it inherits the same structure: the 2D patch is determined by a 1D signal \emph{swept} along the stripe direction. The geometric tokenizer directly extracts this underlying 1D representation by sampling the patch along the diagonal closest to the normal of the stripe direction (red dashed lines in Figure~\ref{fig: geometric_tokenizer}). Sampling $K = \lceil P\sqrt{2}\rceil$ equispaced values along this diagonal yields the \emph{geometric feature} $\mathbf{p}_r \in \R^{K}$, which serves as a concise 1D summary of the patch. Finally, a learnable linear encoder $\mathbf{E}: \R^{K} \to \R^{D}$ maps this feature to the token embedding $\mathbf{z}_r = \mathbf{E}\mathbf{p}_r \in \R^{D}$.

The geometric detokenizer reverses the process; a learnable linear decoder $\mathbf{E}': \R^{D} \to \R^{K}$ turns a predicted embedding $\hat{\mathbf{z}}_q$ back into a $1$-D feature, which is then swept perpendicular to the queried angle $\theta_q$ to fill in the output patch $\hat{\tt}_q$. Further information on the network architecture and training details is provided in Appendices~\ref{sec: tomo_arch_details} and~\ref{sec: training_details}.

\begin{revblock}
\subsection{\texorpdfstring{\rev{Ray-consistent stitching}}{Ray-consistent stitching}}
\label{sec: sino_aggregation}
\begin{wrapfigure}[14]{r}{0.5\textwidth}
    \vspace{-\intextsep}
    \centering
    \includegraphics[width=0.5\textwidth]{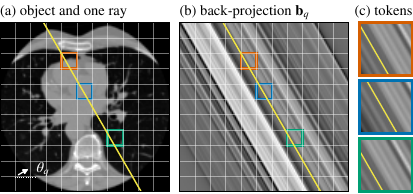}
    \caption{\rev{Ray-consistent stitching (RCS). The predicted back-projection patches (red) are averaged along each ray to enforce consistency across patches.}}
    \label{fig: sino_consistency}
\end{wrapfigure}
We now describe how the predicted tokens are assembled into the final image. The standard approach, patch-based image-domain models, tiles the $N \times N$ image into $P \times P$ patches with stride $P$. In each patch, the predicted tokens fill the missing views, the measured tokens are kept unchanged, and all $R_{\text{total}}$ tokens are averaged as in \eqref{eq:FBP}. However, each patch is predicted independently which can cause inconsistencies and blocking artifacts. To address this issue, we propose a new inference method specifically designed for TomoTransformer, termed ray-consistent stitching (RCS).

As shown in Figure~\ref{fig: sino_consistency}, a single-view back-projection is constant along each ray so all patches crossed by the same ray at angle $\theta_q$ should predict the same value, namely the filtered projection $\tilde{\s}_q$ of that ray. However, this cross-patch consistency is not guaranteed by the network, which predicts each local back-projection patch separately. RCS enforces it by averaging, for every missing angle, the predictions of all patches along each ray. This yields a single filtered projection per missing view; the measured views are kept unchanged and the image is reconstructed by FBP. RCS is a cheap and effective method for killing the blocking artifacts which remarkably improves the reconstruction quality. Please refer to Appendix~\ref{sec: sino_aggregation_appendix} for more details.

\end{revblock}

\begin{revblock}
\subsection{\texorpdfstring{\rev{Supervised and self-supervised fine-tuning}}{Supervised and self-supervised fine-tuning}}
\label{sec: ssl}

A pre-trained TomoTransformer can be applied to a new dataset or scan without any fine-tuning. However, when the test data lies far from the training distribution, for instance a material, contrast or structure with no counterpart in the training set, a domain gap remains. When a small set of ground-truth images from the target distribution is available, the pre-trained model can be specialized to the task at hand by \emph{supervised fine-tuning}: we can fine-tune the pre-trained model with the training loss of Appendix~\ref{sec: training_details}.

In practice, real measurements, however, may come without any ground truth. In contrast to image-domain models that only permits supervised fine-tuning, we can fine-tune the pretrained TomoTransformer to the new scan in a \emph{self-supervised} manner; we randomly mask a subset of the measured projections and ask the model to predict them from the remaining ones. We then minimize the $\ell_1$ loss between the predicted and masked measured views. Since TomoTransformer processes local patches, even a single scan provides a large number of training samples for fine-tuning. For further details on the self-supervised fine-tuning procedure, please refer to Appendix~\ref{sec: ssl_appendix}.

\end{revblock}

\section{Experiments}
\label{sec: experiments}
In this section, we evaluate TomoTransformer on sparse-view CT reconstruction tasks. We first describe the training dataset and the common experimental setup.

\subsection{Experimental setup}
\label{sec: exp_setup}

To train a versatile model across diverse anatomical regions, we build a large heterogeneous dataset of medical and natural images. The medical component comprises chest CTs from LoDoPaB-CT~\citep{leuschner2021lodopab} and the COVID-19 subset of CTSpine1K~\citep{deng2021ctspine1k}, abdominal CTs from KiTS23~\citep{heller2023kits21}, and further thoracic, colonographic, and head-and-neck volumes from the MSD, COLONOG, and HNSCC subsets of CTSpine1K~\citep{deng2021ctspine1k}. To expand structural diversity beyond anatomy, we additionally include natural images from FFHQ~\citep{karras2019style} and ImageNet~\citep{deng2009imagenet} as texture priors. Because TomoTransformer operates on local $P\times P$ tokens, no resampling to a common grid is required, so we use the axial slices of each dataset at their \emph{native} resolutions, normalized per slice into a shared intensity band. In total, the training set comprises $\sim$650k 2D slices: 570k CT slices from 1{,}340 volumes and 80k natural images. Since every slice contributes many patches, each paired with a freshly sampled set of projection angles, the model is exposed to a massive training set, which is essential for robust generalization to unseen anatomies and distributions.

We simulate 2D parallel-beam acquisition following Section~\ref{sec: computed_tomography}. Both dense-view reference images and sparse-view inputs are reconstructed using FBP. We evaluate performance using peak signal-to-noise ratio (PSNR, in dB) and structural similarity index measure (SSIM), computed on reconstructed images against the dense-view targets. \rev{Each test set consists of $32$ slices drawn independently, each from a randomly chosen held-out volume ($6$--$80$ test volumes per dataset).}

\subsection{Fixed sparse-view reconstruction}
\label{sec: fixed_sparsity}

In this section, we evaluate TomoTransformer under a \emph{fixed} sparse-view protocol, where every method, including TomoTransformer, is trained and evaluated at a single sparsity level: $R_{\text{sparse}} = 64$ projections uniformly sampled from a dense $R_{\text{total}} = 256$ projections. We consider post-processing baselines that map sparse-view FBP reconstructions to dense-view target FBPs: U-Net~\citep{jin2017deep}, DRUNet~\citep{zhang2021plug} and NAFNet~\citep{chen2022simple} that use CNN and Restormer~\citep{zamir2022restormer} and CTformer~\citep{wang2023ctformer} that use transformers. \rev{We also include a sinogram-domain baseline, the convolutional interpolation network (CIN) \citep{guo2023efficient}, which predicts the missing views from the measured ones before FBP.} For a fair comparison, all baselines have approximately $12\,\text{M}$ parameters. All models are trained on the same heterogeneous dataset described in Section~\ref{sec: exp_setup}. Please refer to Appendices~\ref{sec: tomo_arch_details}--\ref{sec: baseline_details} for the network architectures and training details of the baselines and TomoTransformer.

\begin{figure*}[t]
  \centering
 \begin{subfigure}[t]{\textwidth}
  \centering
 \begin{minipage}[c]{0.02\textwidth}
  \centering
  \rotatebox{90}{\small\textbf{Chest}}
 \end{minipage}%
 \hspace{0.022\textwidth}\begin{minipage}[c]{0.95\textwidth}
 \begin{subfigure}[t]{0.118\textwidth}
  \captionsetup{labelformat=empty,font=sevenpt}
  \centering
  \begin{tikzpicture}[spy using outlines={circle,orange,line width=0.5pt,magnification=3,size=0.8509cm, connect spies}]
    \node[name=img, inner sep=0pt, line width=0.05mm, draw=white] at (0,0) {\includegraphics[width=\textwidth]{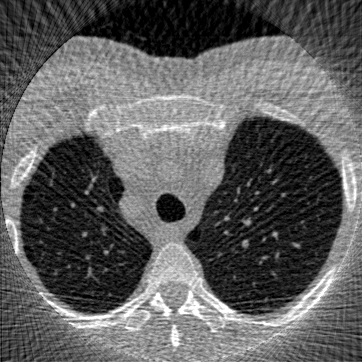}};
    \node[anchor=south west, draw=yellow, line width=0.35pt, fill=black, fill opacity=0.55, text=yellow, text opacity=1, inner sep=1.1pt, font=\tiny]
          at (img.south west) {27.1\,dB};
    \spy on ($(img.center)+(0.2718,-0.1182)$) in node at ($(img.north west)+(0.4491,-0.4491)$);
  \end{tikzpicture}
 \end{subfigure}\hfill
 \begin{subfigure}[t]{0.118\textwidth}
  \captionsetup{labelformat=empty,font=sevenpt}
  \centering
  \begin{tikzpicture}[spy using outlines={circle,orange,line width=0.5pt,magnification=3,size=0.8509cm, connect spies}]
    \node[name=img, inner sep=0pt, line width=0.05mm, draw=white] at (0,0) {\includegraphics[width=\textwidth]{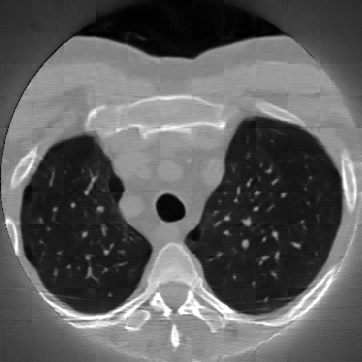}};
    \node[anchor=south west, draw=yellow, line width=0.35pt, fill=black, fill opacity=0.55, text=yellow, text opacity=1, inner sep=1.1pt, font=\tiny]
          at ( img.south west) {34.1\,dB};
    \spy on ($(img.center)+(0.2718,-0.1182)$) in node at ($(img.north west)+(0.4491,-0.4491)$);
  \end{tikzpicture}
 \end{subfigure}\hfill
 \begin{subfigure}[t]{0.118\textwidth}
  \captionsetup{labelformat=empty,font=sevenpt}
  \centering
  \begin{tikzpicture}[spy using outlines={circle,orange,line width=0.5pt,magnification=3,size=0.8509cm, connect spies}]
    \node[name=img, inner sep=0pt, line width=0.05mm, draw=white] at (0,0) {\includegraphics[width=\textwidth]{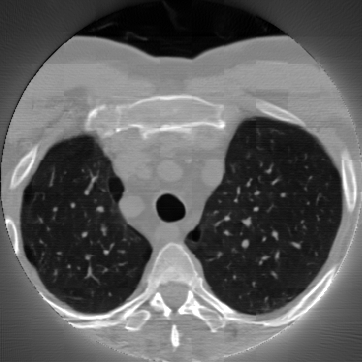}};
    \node[anchor=south west, draw=yellow, line width=0.35pt, fill=black, fill opacity=0.55, text=yellow, text opacity=1, inner sep=1.1pt, font=\tiny]
          at ( img.south west) {35.0\,dB};
    \spy on ($(img.center)+(0.2718,-0.1182)$) in node at ($(img.north west)+(0.4491,-0.4491)$);
  \end{tikzpicture}
 \end{subfigure}\hfill
 \begin{subfigure}[t]{0.118\textwidth}
  \captionsetup{labelformat=empty,font=sevenpt}
  \centering
  \begin{tikzpicture}[spy using outlines={circle,orange,line width=0.5pt,magnification=3,size=0.8509cm, connect spies}]
    \node[name=img, inner sep=0pt, line width=0.05mm, draw=white] at (0,0) {\includegraphics[width=\textwidth]{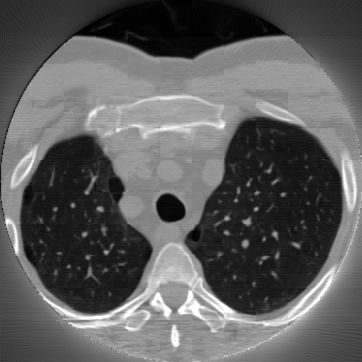}};
    \node[anchor=south west, draw=yellow, line width=0.35pt, fill=black, fill opacity=0.55, text=yellow, text opacity=1, inner sep=1.1pt, font=\tiny]
          at ( img.south west) {35.2\,dB};
    \spy on ($(img.center)+(0.2718,-0.1182)$) in node at ($(img.north west)+(0.4491,-0.4491)$);
  \end{tikzpicture}
 \end{subfigure}\hfill
 \begin{subfigure}[t]{0.118\textwidth}
  \captionsetup{labelformat=empty,font=sevenpt}
  \centering
  \begin{tikzpicture}[spy using outlines={circle,orange,line width=0.5pt,magnification=3,size=0.8509cm, connect spies}]
    \node[name=img, inner sep=0pt, line width=0.05mm, draw=white] at (0,0) {\includegraphics[width=\textwidth]{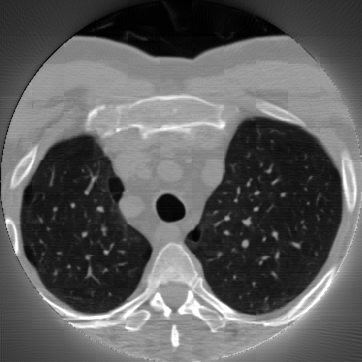}};
    \node[anchor=south west, draw=yellow, line width=0.35pt, fill=black, fill opacity=0.55, text=yellow, text opacity=1, inner sep=1.1pt, font=\tiny]
          at ( img.south west) {35.3\,dB};
    \spy on ($(img.center)+(0.2718,-0.1182)$) in node at ($(img.north west)+(0.4491,-0.4491)$);
  \end{tikzpicture}
 \end{subfigure}\hfill
 \begin{subfigure}[t]{0.118\textwidth}
  \captionsetup{labelformat=empty,font=sevenpt}
  \centering
  \begin{tikzpicture}[spy using outlines={circle,orange,line width=0.5pt,magnification=3,size=0.8509cm, connect spies}]
    \node[name=img, inner sep=0pt, line width=0.05mm, draw=white] at (0,0) {\includegraphics[width=\textwidth]{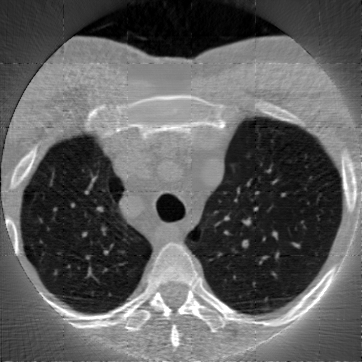}};
    \node[anchor=south west, draw=yellow, line width=0.35pt, fill=black, fill opacity=0.55, text=yellow, text opacity=1, inner sep=1.1pt, font=\tiny]
          at ( img.south west) {33.6\,dB};
    \spy on ($(img.center)+(0.2718,-0.1182)$) in node at ($(img.north west)+(0.4491,-0.4491)$);
  \end{tikzpicture}
 \end{subfigure}\hfill
 \begin{subfigure}[t]{0.118\textwidth}
  \captionsetup{labelformat=empty,font=sevenpt}
  \centering
  \begin{tikzpicture}[spy using outlines={circle,orange,line width=0.5pt,magnification=3,size=0.8509cm, connect spies}]
    \node[name=img, inner sep=0pt, line width=0.05mm, draw=white] at (0,0) {\includegraphics[width=\textwidth]{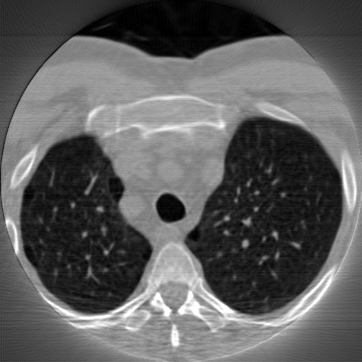}};
    \node[anchor=south west, draw=yellow, line width=0.35pt, fill=black, fill opacity=0.55, text=yellow, text opacity=1, inner sep=1.1pt, font=\tiny]
          at ( img.south west) {35.5\,dB};
    \spy on ($(img.center)+(0.2718,-0.1182)$) in node at ($(img.north west)+(0.4491,-0.4491)$);
  \end{tikzpicture}
 \end{subfigure}\hfill
 \begin{subfigure}[t]{0.118\textwidth}
  \captionsetup{labelformat=empty,font=sevenpt}
  \centering
  \begin{tikzpicture}[spy using outlines={circle,orange,line width=0.5pt,magnification=3,size=0.8509cm, connect spies}]
    \node[name=img, inner sep=0pt, line width=0.05mm, draw=white] at (0,0) {\includegraphics[width=\textwidth]{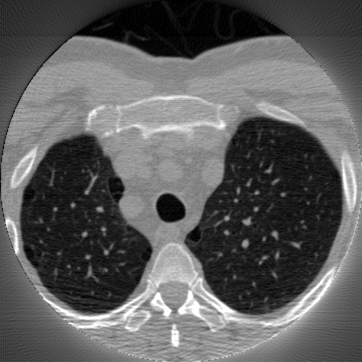}};
    \spy on ($(img.center)+(0.2718,-0.1182)$) in node at ($(img.north west)+(0.4491,-0.4491)$);
  \end{tikzpicture}
 \end{subfigure}\hfill
 \end{minipage}
 \end{subfigure}

 \begin{subfigure}[t]{\textwidth}
  \centering
 \begin{minipage}[c]{0.02\textwidth}
  \centering
  \rotatebox{90}{\small\textbf{Abdomen}}
 \end{minipage}%
 \hspace{0.022\textwidth}\begin{minipage}[c]{0.95\textwidth}
 \begin{subfigure}[t]{0.118\textwidth}
  \captionsetup{labelformat=empty,font=sevenpt}
  \centering
  \begin{tikzpicture}[spy using outlines={circle,orange,line width=0.5pt,magnification=3,size=0.8509cm, connect spies}]
    \node[name=img, inner sep=0pt, line width=0.05mm, draw=white] at (0,0) {\includegraphics[width=\textwidth]{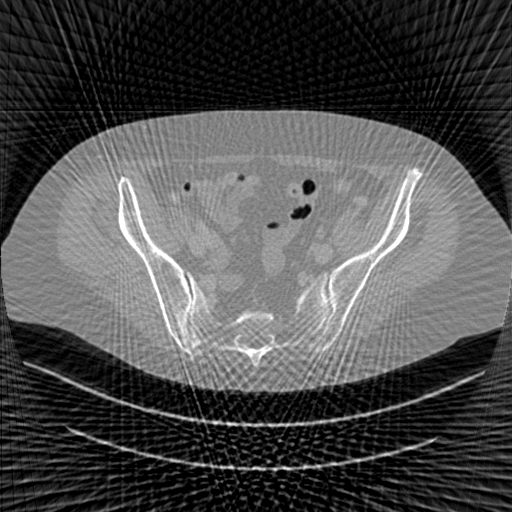}};
    \node[anchor=south west, draw=yellow, line width=0.35pt, fill=black, fill opacity=0.55, text=yellow, text opacity=1, inner sep=1.1pt, font=\tiny]
          at ( img.south west) {26.0\,dB};
    \spy on ($(img.center)+(-0.2482,-0.2127)$) in node at ($(img.north west)+(0.4491,-0.4491)$);
  \end{tikzpicture}
 \end{subfigure}\hfill
 \begin{subfigure}[t]{0.118\textwidth}
  \captionsetup{labelformat=empty,font=sevenpt}
  \centering
  \begin{tikzpicture}[spy using outlines={circle,orange,line width=0.5pt,magnification=3,size=0.8509cm, connect spies}]
    \node[name=img, inner sep=0pt, line width=0.05mm, draw=white] at (0,0) {\includegraphics[width=\textwidth]{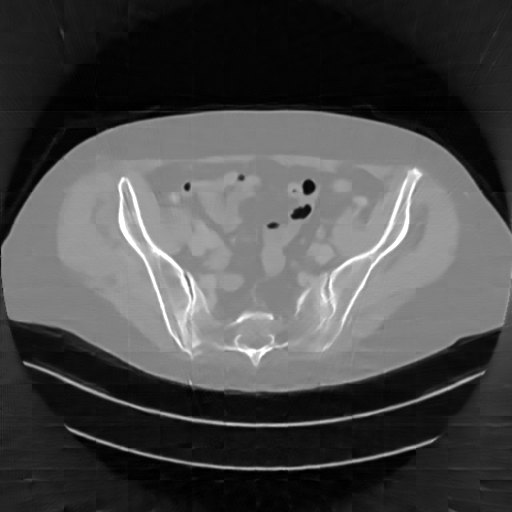}};
    \node[anchor=south west, draw=yellow, line width=0.35pt, fill=black, fill opacity=0.55, text=yellow, text opacity=1, inner sep=1.1pt, font=\tiny]
          at ( img.south west) {39.0\,dB};
    \spy on ($(img.center)+(-0.2482,-0.2127)$) in node at ($(img.north west)+(0.4491,-0.4491)$);
  \end{tikzpicture}
 \end{subfigure}\hfill
 \begin{subfigure}[t]{0.118\textwidth}
  \captionsetup{labelformat=empty,font=sevenpt}
  \centering
  \begin{tikzpicture}[spy using outlines={circle,orange,line width=0.5pt,magnification=3,size=0.8509cm, connect spies}]
    \node[name=img, inner sep=0pt, line width=0.05mm, draw=white] at (0,0) {\includegraphics[width=\textwidth]{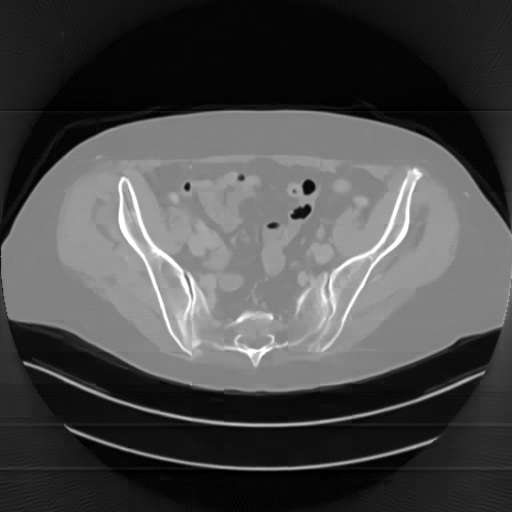}};
    \node[anchor=south west, draw=yellow, line width=0.35pt, fill=black, fill opacity=0.55, text=yellow, text opacity=1, inner sep=1.1pt, font=\tiny]
          at ( img.south west) {41.6\,dB};
    \spy on ($(img.center)+(-0.2482,-0.2127)$) in node at ($(img.north west)+(0.4491,-0.4491)$);
  \end{tikzpicture}
 \end{subfigure}\hfill
 \begin{subfigure}[t]{0.118\textwidth}
  \captionsetup{labelformat=empty,font=sevenpt}
  \centering
  \begin{tikzpicture}[spy using outlines={circle,orange,line width=0.5pt,magnification=3,size=0.8509cm, connect spies}]
    \node[name=img, inner sep=0pt, line width=0.05mm, draw=white] at (0,0) {\includegraphics[width=\textwidth]{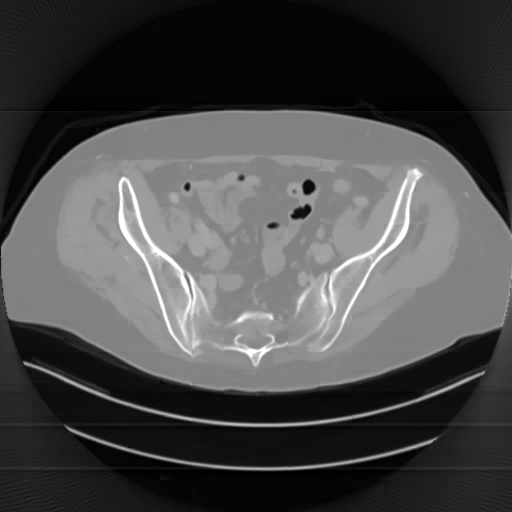}};
    \node[anchor=south west, draw=yellow, line width=0.35pt, fill=black, fill opacity=0.55, text=yellow, text opacity=1, inner sep=1.1pt, font=\tiny]
          at ( img.south west) {42.3\,dB};
    \spy on ($(img.center)+(-0.2482,-0.2127)$) in node at ($(img.north west)+(0.4491,-0.4491)$);
  \end{tikzpicture}
 \end{subfigure}\hfill
 \begin{subfigure}[t]{0.118\textwidth}
  \captionsetup{labelformat=empty,font=sevenpt}
  \centering
  \begin{tikzpicture}[spy using outlines={circle,orange,line width=0.5pt,magnification=3,size=0.8509cm, connect spies}]
    \node[name=img, inner sep=0pt, line width=0.05mm, draw=white] at (0,0) {\includegraphics[width=\textwidth]{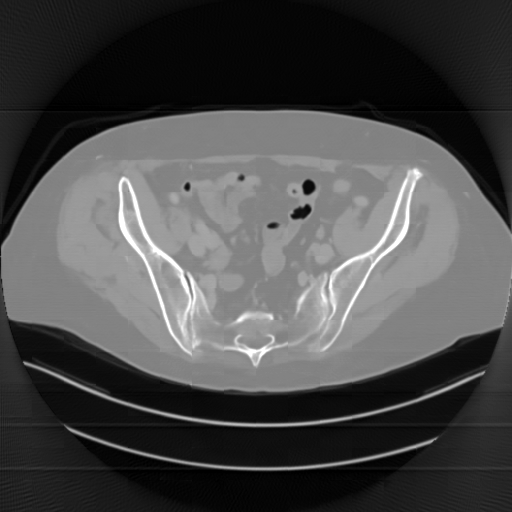}};
    \node[anchor=south west, draw=yellow, line width=0.35pt, fill=black, fill opacity=0.55, text=yellow, text opacity=1, inner sep=1.1pt, font=\tiny]
          at ( img.south west) {42.5\,dB};
    \spy on ($(img.center)+(-0.2482,-0.2127)$) in node at ($(img.north west)+(0.4491,-0.4491)$);
  \end{tikzpicture}
 \end{subfigure}\hfill
 \begin{subfigure}[t]{0.118\textwidth}
  \captionsetup{labelformat=empty,font=sevenpt}
  \centering
  \begin{tikzpicture}[spy using outlines={circle,orange,line width=0.5pt,magnification=3,size=0.8509cm, connect spies}]
    \node[name=img, inner sep=0pt, line width=0.05mm, draw=white] at (0,0) {\includegraphics[width=\textwidth]{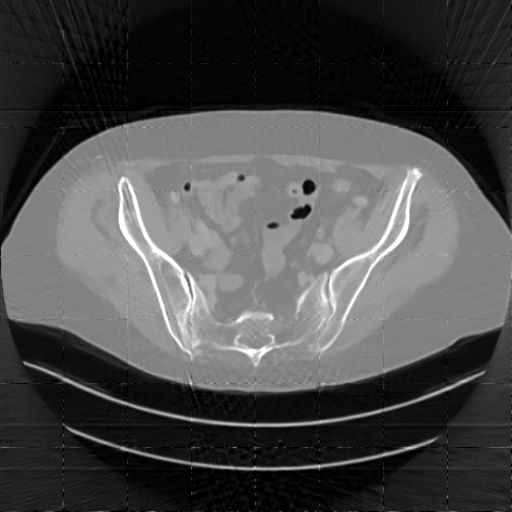}};
    \node[anchor=south west, draw=yellow, line width=0.35pt, fill=black, fill opacity=0.55, text=yellow, text opacity=1, inner sep=1.1pt, font=\tiny]
          at ( img.south west) {36.6\,dB};
    \spy on ($(img.center)+(-0.2482,-0.2127)$) in node at ($(img.north west)+(0.4491,-0.4491)$);
  \end{tikzpicture}
 \end{subfigure}\hfill
 \begin{subfigure}[t]{0.118\textwidth}
  \captionsetup{labelformat=empty,font=sevenpt}
  \centering
  \begin{tikzpicture}[spy using outlines={circle,orange,line width=0.5pt,magnification=3,size=0.8509cm, connect spies}]
    \node[name=img, inner sep=0pt, line width=0.05mm, draw=white] at (0,0) {\includegraphics[width=\textwidth]{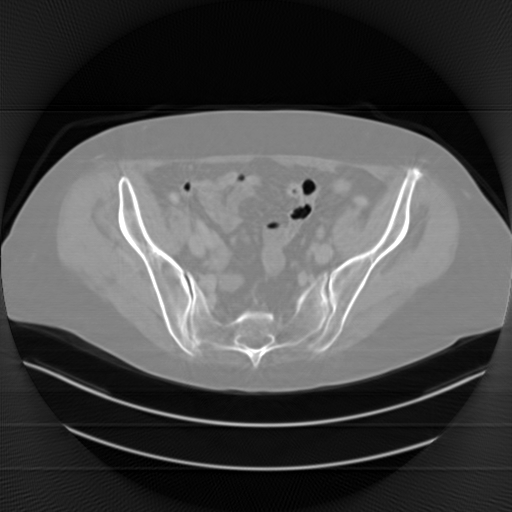}};
    \node[anchor=south west, draw=yellow, line width=0.35pt, fill=black, fill opacity=0.55, text=yellow, text opacity=1, inner sep=1.1pt, font=\tiny]
          at ( img.south west) {44.0\,dB};
    \spy on ($(img.center)+(-0.2482,-0.2127)$) in node at ($(img.north west)+(0.4491,-0.4491)$);
  \end{tikzpicture}
 \end{subfigure}\hfill
 \begin{subfigure}[t]{0.118\textwidth}
  \captionsetup{labelformat=empty,font=sevenpt}
  \centering
  \begin{tikzpicture}[spy using outlines={circle,orange,line width=0.5pt,magnification=3,size=0.8509cm, connect spies}]
    \node[name=img, inner sep=0pt, line width=0.05mm, draw=white] at (0,0) {\includegraphics[width=\textwidth]{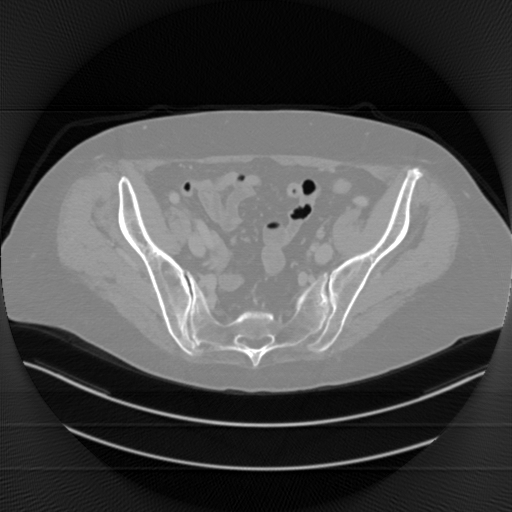}};
    \spy on ($(img.center)+(-0.2482,-0.2127)$) in node at ($(img.north west)+(0.4491,-0.4491)$);
  \end{tikzpicture}
 \end{subfigure}\hfill
 \end{minipage}
 \end{subfigure}

 \begin{subfigure}[t]{\textwidth}
  \centering
 \begin{minipage}[c]{0.02\textwidth}
  \centering
  \raisebox{1em}{\rotatebox{90}{\small\textbf{ImageNet}}}
 \end{minipage}%
 \hspace{0.022\textwidth}\begin{minipage}[c]{0.95\textwidth}
 \begin{subfigure}[t]{0.118\textwidth}
  \captionsetup{labelformat=empty,font=sevenpt}
  \centering
  \begin{tikzpicture}[spy using outlines={circle,orange,line width=0.5pt,magnification=3,size=0.8509cm, connect spies}]
    \node[name=img, inner sep=0pt, line width=0.05mm, draw=white] at (0,0) {\includegraphics[width=\textwidth]{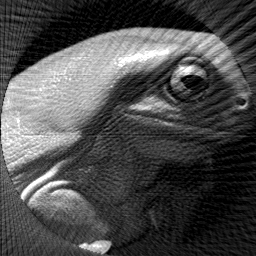}};
    \node[anchor=south west, draw=yellow, line width=0.35pt, fill=black, fill opacity=0.55, text=yellow, text opacity=1, inner sep=1.1pt, font=\tiny]
          at ( img.south west) {27.9\,dB};
    \spy on ($(img.center)+(0.4018,0.4609)$) in node at ($(img.north west)+(0.4491,-0.4491)$);
  \end{tikzpicture}
  \caption*{\makebox[\linewidth][c]{FBP (64 projs)}}
 \end{subfigure}\hfill
 \begin{subfigure}[t]{0.118\textwidth}
  \captionsetup{labelformat=empty,font=sevenpt}
  \centering
  \begin{tikzpicture}[spy using outlines={circle,orange,line width=0.5pt,magnification=3,size=0.8509cm, connect spies}]
    \node[name=img, inner sep=0pt, line width=0.05mm, draw=white] at (0,0) {\includegraphics[width=\textwidth]{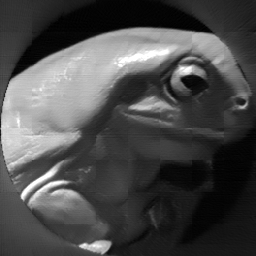}};
    \node[anchor=south west, draw=yellow, line width=0.35pt, fill=black, fill opacity=0.55, text=yellow, text opacity=1, inner sep=1.1pt, font=\tiny]
          at ( img.south west) {33.8\,dB};
    \spy on ($(img.center)+(0.4018,0.4609)$) in node at ($(img.north west)+(0.4491,-0.4491)$);
  \end{tikzpicture}
  \caption*{U-Net}
 \end{subfigure}\hfill
 \begin{subfigure}[t]{0.118\textwidth}
  \captionsetup{labelformat=empty,font=sevenpt}
  \centering
  \begin{tikzpicture}[spy using outlines={circle,orange,line width=0.5pt,magnification=3,size=0.8509cm, connect spies}]
    \node[name=img, inner sep=0pt, line width=0.05mm, draw=white] at (0,0) {\includegraphics[width=\textwidth]{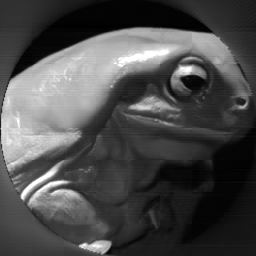}};
    \node[anchor=south west, draw=yellow, line width=0.35pt, fill=black, fill opacity=0.55, text=yellow, text opacity=1, inner sep=1.1pt, font=\tiny]
          at ( img.south west) {35.6\,dB};
    \spy on ($(img.center)+(0.4018,0.4609)$) in node at ($(img.north west)+(0.4491,-0.4491)$);
  \end{tikzpicture}
  \caption*{DRUNet}
 \end{subfigure}\hfill
 \begin{subfigure}[t]{0.118\textwidth}
  \captionsetup{labelformat=empty,font=sevenpt}
  \centering
  \begin{tikzpicture}[spy using outlines={circle,orange,line width=0.5pt,magnification=3,size=0.8509cm, connect spies}]
    \node[name=img, inner sep=0pt, line width=0.05mm, draw=white] at (0,0) {\includegraphics[width=\textwidth]{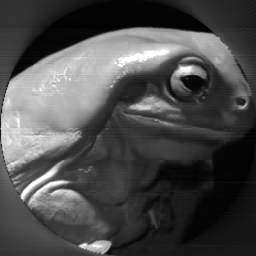}};
    \node[anchor=south west, draw=yellow, line width=0.35pt, fill=black, fill opacity=0.55, text=yellow, text opacity=1, inner sep=1.1pt, font=\tiny]
          at ( img.south west) {36.2\,dB};
    \spy on ($(img.center)+(0.4018,0.4609)$) in node at ($(img.north west)+(0.4491,-0.4491)$);
  \end{tikzpicture}
  \caption*{NAFNet}
 \end{subfigure}\hfill
 \begin{subfigure}[t]{0.118\textwidth}
  \captionsetup{labelformat=empty,font=sevenpt}
  \centering
  \begin{tikzpicture}[spy using outlines={circle,orange,line width=0.5pt,magnification=3,size=0.8509cm, connect spies}]
    \node[name=img, inner sep=0pt, line width=0.05mm, draw=white] at (0,0) {\includegraphics[width=\textwidth]{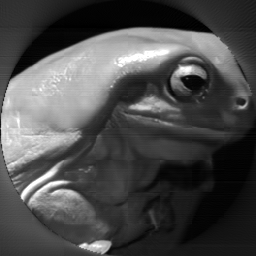}};
    \node[anchor=south west, draw=yellow, line width=0.35pt, fill=black, fill opacity=0.55, text=yellow, text opacity=1, inner sep=1.1pt, font=\tiny]
          at ( img.south west) {36.4\,dB};
    \spy on ($(img.center)+(0.4018,0.4609)$) in node at ($(img.north west)+(0.4491,-0.4491)$);
  \end{tikzpicture}
  \caption*{Restormer}
 \end{subfigure}\hfill
 \begin{subfigure}[t]{0.118\textwidth}
  \captionsetup{labelformat=empty,font=sevenpt}
  \centering
  \begin{tikzpicture}[spy using outlines={circle,orange,line width=0.5pt,magnification=3,size=0.8509cm, connect spies}]
    \node[name=img, inner sep=0pt, line width=0.05mm, draw=white] at (0,0) {\includegraphics[width=\textwidth]{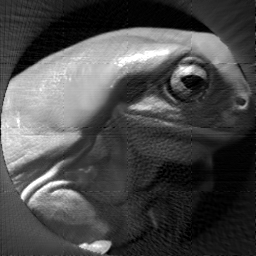}};
    \node[anchor=south west, draw=yellow, line width=0.35pt, fill=black, fill opacity=0.55, text=yellow, text opacity=1, inner sep=1.1pt, font=\tiny]
          at ( img.south west) {33.3\,dB};
    \spy on ($(img.center)+(0.4018,0.4609)$) in node at ($(img.north west)+(0.4491,-0.4491)$);
  \end{tikzpicture}
  \caption*{CTformer}
 \end{subfigure}\hfill
 \begin{subfigure}[t]{0.118\textwidth}
  \captionsetup{labelformat=empty,font=sevenpt}
  \centering
  \begin{tikzpicture}[spy using outlines={circle,orange,line width=0.5pt,magnification=3,size=0.8509cm, connect spies}]
    \node[name=img, inner sep=0pt, line width=0.05mm, draw=white] at (0,0) {\includegraphics[width=\textwidth]{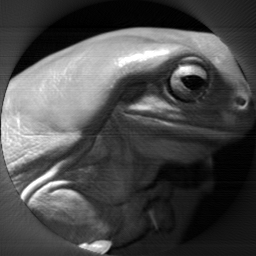}};
    \node[anchor=south west, draw=yellow, line width=0.35pt, fill=black, fill opacity=0.55, text=yellow, text opacity=1, inner sep=1.1pt, font=\tiny]
          at ( img.south west) {37.2\,dB};
    \spy on ($(img.center)+(0.4018,0.4609)$) in node at ($(img.north west)+(0.4491,-0.4491)$);
  \end{tikzpicture}
  \caption*{\makebox[\linewidth][c]{TomoTransformer\hspace*{3mm}}}
 \end{subfigure}\hfill
 \begin{subfigure}[t]{0.118\textwidth}
  \captionsetup{labelformat=empty,font=sevenpt}
  \centering
  \begin{tikzpicture}[spy using outlines={circle,orange,line width=0.5pt,magnification=3,size=0.8509cm, connect spies}]
    \node[name=img, inner sep=0pt, line width=0.05mm, draw=white] at (0,0) {\includegraphics[width=\textwidth]{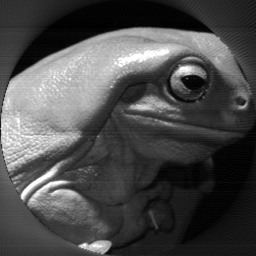}};
    \spy on ($(img.center)+(0.4018,0.4609)$) in node at ($(img.north west)+(0.4491,-0.4491)$);
  \end{tikzpicture}
  \caption*{\makebox[\linewidth][c]{FBP (256 projs)}}
 \end{subfigure}\hfill
 \end{minipage}
 \end{subfigure}

  \caption{Fixed sparse-view reconstruction ($64\!\rightarrow\!256$ projections) across three domains: chest CT (LoDoPaB-CT), abdominal CT (KiTS23), and a natural image (ImageNet). The box in the bottom-left of each panel reports PSNR (dB) against the dense $256$-view reference.}
  \label{fig: fixed_sparsity}
\end{figure*}

Table~\ref{tab: fixed_sparsity_psnr} and Table~\ref{tab: fixed_sparsity_ssim} (Appendix) report per-dataset PSNR and SSIM values, averaged over $32$ test slices, with reconstructions visualized in Figure~\ref{fig: fixed_sparsity}. These results demonstrate that TomoTransformer (fixed) outperforms all competing baselines on average (last column). \rev{By using our proposed iference method, ray-consistent stitching (RCS, Section~\ref{sec: sino_aggregation}), TomoTransformer (fixed, RCS) raises the average PSNR by $0.41$\,dB, making it the best model on every dataset in both metrics.} For reference, we also report TomoTransformer (varied), a single model trained on variable sparsity and applied at $64\!\rightarrow\!256$ without retraining. \rev{Although it is not specialized to this protocol, it is only $0.81$\,dB below the best protocol-specific baseline on average ($38.66$ against $39.47$\,dB, Restormer).} \rev{After supervised fine-tuning (Section~\ref{sec: ssl}) on only $400$ training slices ($50$ per dataset), it outperforms every image-domain baseline, see the last row of Tables~\ref{tab: fixed_sparsity_psnr} and~\ref{tab: fixed_sparsity_ssim}}. \rev{Nevrtheless, we should note that the drawback of the above fixed-protocol models appears as soon as the sparsity pattern changes. Here we apply the same models trained at $64\!\rightarrow\!256$ in Table~\ref{tab: fixed_sparsity_psnr} to a different sparsity level, $128\!\rightarrow\!256$, without retraining. As shown in Table~\ref{tab: fixed_sparsity_128} (Appendix), TomoTransformer (varied), even without fine-tuning, is best on all eight datasets, around $2.0$\,dB above the best fixed-protocol model, Restormer.}

\begin{table*}[t]
\caption{Fixed sparse-view reconstruction ($64\!\rightarrow\!256$): \textbf{PSNR} (dB) per dataset (against the dense $256$-view reference) for TomoTransformer and the baselines. \rev{RCS: using ray-consistent stitching (Section~\ref{sec: sino_aggregation}). + supervised fine-tuning: TomoTransformer (varied, RCS) fine-tuned for $64\!\rightarrow\!256$ on $400$ training slices ($50$ per dataset; Section~\ref{sec: ssl}).} Best per column in \textbf{bold}.}
\centering
\renewcommand\arraystretch{1.25}
\resizebox{\textwidth}{!}{%
\begin{tabular}{c|cccccccc|c}
\hline
Method & LoDoPaB & COVID-19 & KiTS23 & MSD-T10 & COLONOG & HNSCC & FFHQ & ImageNet & Average \\
\hline
FBP~\citep{kak1988principles} & 29.24 & 28.48 & 30.27 & 28.13 & 28.59 & 28.93 & 27.97 & 26.46 & 28.51 \\
U-Net~\citep{ronneberger2015unet} & 36.76 & 36.97 & 40.45 & 38.38 & 36.88 & 40.20 & 33.43 & 31.78 & 36.86 \\
DRUNet~\citep{zhang2021plug} & 37.95 & 39.35 & 42.95 & 40.24 & 38.34 & 43.58 & 35.23 & 33.42 & 38.88 \\
NAFNet~\citep{chen2022simple} & 38.27 & 39.85 & 43.49 & 40.66 & 38.66 & 44.53 & 35.68 & 33.79 & 39.37 \\
Restormer~\citep{zamir2022restormer} & 38.32 & 39.96 & 43.58 & 40.73 & 38.75 & 44.73 & 35.78 & 33.88 & 39.47 \\
CTformer~\citep{wang2023ctformer} & 35.79 & 36.94 & 38.33 & 36.39 & 35.80 & 37.72 & 33.97 & 31.55 & 35.81 \\
\rev{CIN~\citep{guo2023efficient}} & \rev{37.31} & \rev{38.91} & \rev{42.66} & \rev{40.00} & \rev{38.15} & \rev{42.67} & \rev{35.34} & \rev{33.32} & \rev{38.55} \\
\hline
TomoTransformer (fixed) & 38.19 & 40.23 & 43.94 & 41.15 & 38.86 & 44.08 & 36.94 & 33.89 & 39.66 \\
\rev{TomoTransformer (fixed, RCS)} & \rev{\textbf{38.47}} & \rev{\textbf{40.65}} & \rev{\textbf{44.36}} & \rev{\textbf{41.50}} & \rev{\textbf{39.09}} & \rev{\textbf{44.98}} & \rev{\textbf{37.34}} & \rev{\textbf{34.19}} & \rev{\textbf{40.07}} \\
\hline
TomoTransformer (varied) & 37.47 & 39.05 & 42.68 & 40.02 & 38.18 & 39.61 & 36.11 & 33.42 & 38.32 \\
\rev{TomoTransformer (varied, RCS)} & \rev{37.73} & \rev{39.48} & \rev{43.18} & \rev{40.43} & \rev{38.46} & \rev{39.87} & \rev{36.46} & \rev{33.70} & \rev{38.66} \\
\rev{+ supervised fine-tuning} & \rev{38.16} & \rev{40.19} & \rev{43.92} & \rev{41.09} & \rev{38.86} & \rev{43.20} & \rev{36.85} & \rev{34.05} & \rev{39.54} \\
\hline
\end{tabular}}
\label{tab: fixed_sparsity_psnr}
\end{table*}

\subsection{Varied sparse-view reconstruction}
\label{sec: variable_sparsity}

\begin{figure*}[tb]
  \centering
  \captionsetup[subfigure]{skip=1pt}   %
  \begin{minipage}{0.85\textwidth}

 \begin{subfigure}[t]{\textwidth}
  \centering
  \begin{subfigure}[c]{0.2305\textwidth}
    \captionsetup{labelformat=empty,font=sevenpt}\centering
    \begin{tikzpicture}[spy using outlines={circle,orange,line width=0.5pt,magnification=3,size=0.8509cm, connect spies}]
      \node[name=img, inner sep=0pt, line width=0.05mm, draw=white] at (0,0) {\includegraphics[width=\textwidth]{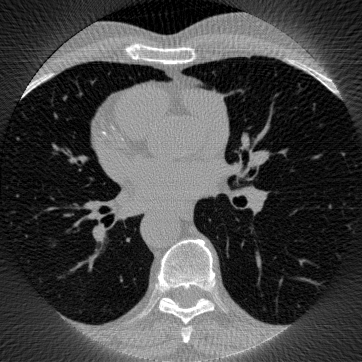}};
      \node[anchor=south west, draw=yellow, line width=0.35pt, fill=black, fill opacity=0.55, text=yellow, text opacity=1, inner sep=1.6pt, font=\footnotesize]
            at (img.south west) {34.5\,dB};
      \spy on ($(img.center)+(-0.6973,-0.3191)$) in node at ($(img.north west)+(0.4491,-0.4491)$);
    \end{tikzpicture}
    \caption*{FBP input ($128$ views)}
  \end{subfigure}%
  \hfill
  \begin{minipage}[c]{0.7435\textwidth}
   \centering
   \begin{subfigure}[c]{0.31\textwidth}
    \captionsetup{labelformat=empty,font=sevenpt}\centering
    \begin{tikzpicture}[spy using outlines={circle,orange,line width=0.5pt,magnification=3,size=0.8509cm, connect spies}]
      \node[name=img, inner sep=0pt, line width=0.05mm, draw=white] at (0,0) {\includegraphics[width=\textwidth]{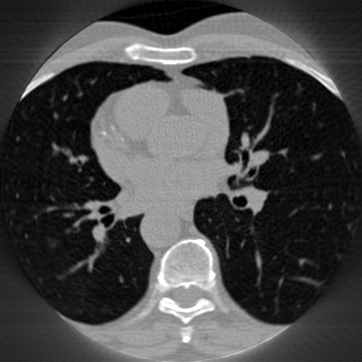}};
      \node[anchor=south west, draw=yellow, line width=0.35pt, fill=black, fill opacity=0.55, text=yellow, text opacity=1, inner sep=1.6pt, font=\footnotesize]
            at (img.south west) {36.1\,dB};
    \spy on ($(img.center)+(-0.6973,-0.3191)$) in node at ($(img.north west)+(0.4491,-0.4491)$);
    \end{tikzpicture}
    \caption*{ViewTrans ($128\!\rightarrow\!256$)}
   \end{subfigure}\hfill
   \begin{subfigure}[c]{0.31\textwidth}
    \captionsetup{labelformat=empty,font=sevenpt}\centering
    \begin{tikzpicture}[spy using outlines={circle,orange,line width=0.5pt,magnification=3,size=0.8509cm, connect spies}]
      \node[name=img, inner sep=0pt, line width=0.05mm, draw=white] at (0,0) {\includegraphics[width=\textwidth]{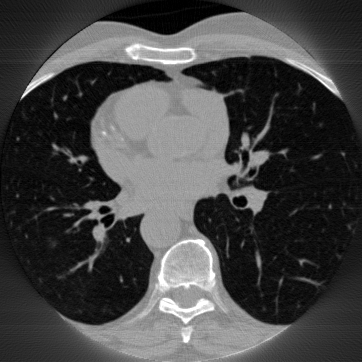}};
      \node[anchor=south west, draw=yellow, line width=0.35pt, fill=black, fill opacity=0.55, text=yellow, text opacity=1, inner sep=1.6pt, font=\footnotesize]
            at (img.south west) {40.3\,dB};
    \spy on ($(img.center)+(-0.6973,-0.3191)$) in node at ($(img.north west)+(0.4491,-0.4491)$);
    \end{tikzpicture}
    \caption*{\makebox[\linewidth][c]{TomoTransformer ($128\!\rightarrow\!256$)}}
   \end{subfigure}\hfill
   \begin{subfigure}[c]{0.31\textwidth}
    \captionsetup{labelformat=empty,font=sevenpt}\centering
    \begin{tikzpicture}[spy using outlines={circle,orange,line width=0.5pt,magnification=3,size=0.8509cm, connect spies}]
      \node[name=img, inner sep=0pt, line width=0.05mm, draw=white] at (0,0) {\includegraphics[width=\textwidth]{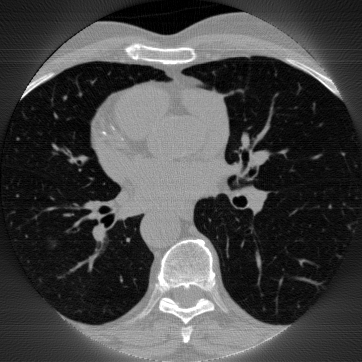}};
    \spy on ($(img.center)+(-0.6973,-0.3191)$) in node at ($(img.north west)+(0.4491,-0.4491)$);
    \end{tikzpicture}
    \caption*{FBP ($256$ views)}
   \end{subfigure}
  \end{minipage}
 \end{subfigure}

  \vspace{2pt}

  \begin{subfigure}[t]{\textwidth}
  \centering
  \begin{subfigure}[c]{0.2305\textwidth}
    \captionsetup{labelformat=empty,font=sevenpt}\centering
    \begin{tikzpicture}[spy using outlines={circle,orange,line width=0.5pt,magnification=3,size=0.8509cm, connect spies}]
      \node[name=img, inner sep=0pt, line width=0.05mm, draw=white] at (0,0) {\includegraphics[width=\textwidth]{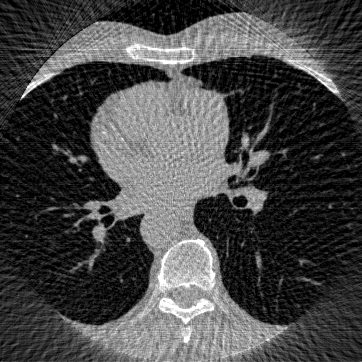}};
      \node[anchor=south west, draw=yellow, line width=0.35pt, fill=black, fill opacity=0.55, text=yellow, text opacity=1, inner sep=1.6pt, font=\footnotesize]
            at (img.south west) {27.5\,dB};
      \spy on ($(img.center)+(-0.6973,-0.3191)$) in node at ($(img.north west)+(0.4491,-0.4491)$);
    \end{tikzpicture}
    \caption*{FBP input ($64$ views)}
  \end{subfigure}%
  \hfill
  \begin{minipage}[c]{0.7435\textwidth}
   \centering
   \begin{subfigure}[c]{0.31\textwidth}
    \captionsetup{labelformat=empty,font=sevenpt}\centering
    \begin{tikzpicture}[spy using outlines={circle,orange,line width=0.5pt,magnification=3,size=0.8509cm, connect spies}]
      \node[name=img, inner sep=0pt, line width=0.05mm, draw=white] at (0,0) {\includegraphics[width=\textwidth]{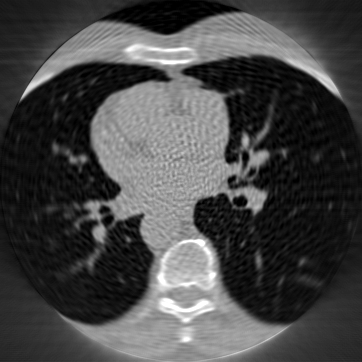}};
      \node[anchor=south west, draw=yellow, line width=0.35pt, fill=black, fill opacity=0.55, text=yellow, text opacity=1, inner sep=1.6pt, font=\footnotesize]
            at (img.south west) {31.5\,dB};
    \spy on ($(img.center)+(-0.6973,-0.3191)$) in node at ($(img.north west)+(0.4491,-0.4491)$);
    \end{tikzpicture}
    \caption*{ViewTrans ($64\!\rightarrow\!256$)}
   \end{subfigure}\hfill
   \begin{subfigure}[c]{0.31\textwidth}
    \captionsetup{labelformat=empty,font=sevenpt}\centering
    \begin{tikzpicture}[spy using outlines={circle,orange,line width=0.5pt,magnification=3,size=0.8509cm, connect spies}]
      \node[name=img, inner sep=0pt, line width=0.05mm, draw=white] at (0,0) {\includegraphics[width=\textwidth]{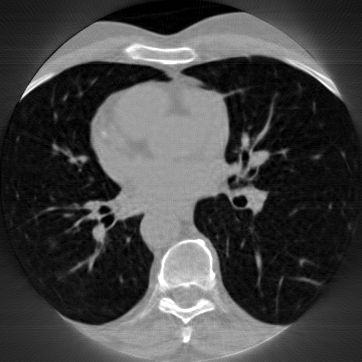}};
      \node[anchor=south west, draw=yellow, line width=0.35pt, fill=black, fill opacity=0.55, text=yellow, text opacity=1, inner sep=1.6pt, font=\footnotesize]
            at (img.south west) {35.9\,dB};
    \spy on ($(img.center)+(-0.6973,-0.3191)$) in node at ($(img.north west)+(0.4491,-0.4491)$);
    \end{tikzpicture}
    \caption*{\makebox[\linewidth][c]{TomoTransformer ($64\!\rightarrow\!256$)}}
   \end{subfigure}\hfill
   \begin{subfigure}[c]{0.31\textwidth}
    \captionsetup{labelformat=empty,font=sevenpt}\centering
    \begin{tikzpicture}[spy using outlines={circle,orange,line width=0.5pt,magnification=3,size=0.8509cm, connect spies}]
      \node[name=img, inner sep=0pt, line width=0.05mm, draw=white] at (0,0) {\includegraphics[width=\textwidth]{figures/sparse_grid_lodo/fbp256dense_lodo_z17.png}};
    \spy on ($(img.center)+(-0.6973,-0.3191)$) in node at ($(img.north west)+(0.4491,-0.4491)$);
    \end{tikzpicture}
    \caption*{FBP ($256$ views)}
   \end{subfigure}
  \end{minipage}
 \end{subfigure}
  \end{minipage}

  \caption{Varied sparse-view reconstruction on LoDoPaB-CT. From $128$ (top) and $64$ (bottom) measured views, TomoTransformer and ViewTrans reconstruct a dense $256$-view target.}
  \label{fig: variable_sparsity}
\end{figure*}

While the previous section evaluated TomoTransformer on a fixed sparse-view setup, we now consider a more realistic and challenging regime: training a single model to handle variable projection counts and arbitrary angular configurations. This flexible paradigm is precisely what TomoTransformer was designed to address \textbf{(R1)}. For this experiment, we compare against ViewTrans~\citep{chen2026viewtrans} as the baselines in the previous section were not designed for variable-view reconstruction. Unlike TomoTransformer, ViewTrans operates directly on the raw sinogram and tokenizes each view into a vector whose dimension equals the detector width; its architecture works only for a single detector size and image resolution and cannot process images with different dimensions. To ensure a fair comparison, we restrict this experiment to the LoDoPaB-CT dataset~\citep{leuschner2021lodopab} at its native $362 \times 362$ resolution across varying sparsity levels. At every training iteration we draw a dense grid size uniformly in $[128, 512]$ and almost uniformly $[20\%, 80\%]$ of them for the measured views, with a small random angular perturbation. TomoTransformer and ViewTrans have similar capacity ($\sim$12M parameters).

We next evaluate the pretrained models using two input sparsity levels: $128$ and $64$ measured projections, reconstructing a $256$-view target. Representative reconstructions are shown in Figure~\ref{fig: variable_sparsity}, and full quantitative results (averaged over $32$ test slices against the dense $256$-view FBP reference) are reported in Table~\ref{tab: variable_sparsity} (Appendix). When given $128$ input views, TomoTransformer significantly outperforms ViewTrans by a wide margin. The contrast becomes even stronger at $64$ input views, an extremely sparse regime that poses a severe challenge for both models. While TomoTransformer maintains a strong quantitative lead, the visual difference in Figure~\ref{fig: variable_sparsity} is even more dramatic. At $64$ views, ViewTrans introduces heavy swirling washing-machine artifacts that do not exist in the reference. This failure mode in ViewTrans highlights a key architectural weakness: because it applies filtered back-projection \emph{after} predicting missing views in the sinogram, small systematic errors in its predicted sinograms get amplified into structured image artifacts under heavy undersampling.

\subsection{Zero-shot denoising}
\label{sec: image_denoising}

\begin{figure*}[tb]
  \centering
  \newcommand{\cw}{0.155\textwidth}
  \begin{minipage}[c]{0.03\textwidth}\centering\rotatebox{90}{\scriptsize\textbf{Gaussian ($45$\,dB})}\end{minipage}\hfill
  \begin{subfigure}[c]{\cw}\centering%
  \begin{tikzpicture}[spy using outlines={circle,orange,line width=0.5pt,magnification=3,size=0.7327cm, connect spies}]
    \node[name=img, inner sep=0pt, line width=0.05mm, draw=white] at (0,0) {\includegraphics[width=\textwidth]{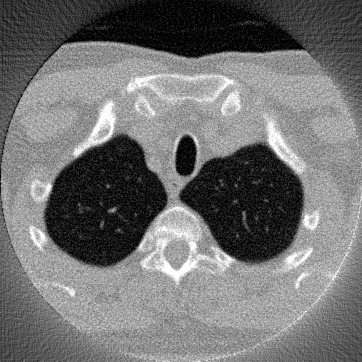}};
    \node[anchor=south west, draw=yellow, line width=0.35pt, fill=black, fill opacity=0.55, text=yellow, text opacity=1, inner sep=1.2pt, font=\tiny] at (img.south west) {28.96\,dB};
    \spy on ($(img.center)+1.181818*(0,0.6)$) in node at ($(img.north west)+(0.39,-0.39)$);
  \end{tikzpicture}\end{subfigure}\hfill
  \begin{subfigure}[c]{\cw}\centering%
  \begin{tikzpicture}[spy using outlines={circle,orange,line width=0.5pt,magnification=3,size=0.7327cm, connect spies}]
    \node[name=img, inner sep=0pt, line width=0.05mm, draw=white] at (0,0) {\includegraphics[width=\textwidth]{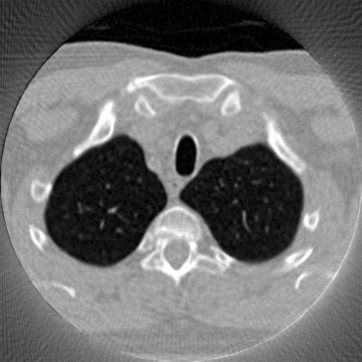}};
    \node[anchor=south west, draw=yellow, line width=0.35pt, fill=black, fill opacity=0.55, text=yellow, text opacity=1, inner sep=1.2pt, font=\tiny] at (img.south west) {35.41\,dB};
    \spy on ($(img.center)+1.181818*(0,0.6)$) in node at ($(img.north west)+(0.39,-0.39)$);
  \end{tikzpicture}\end{subfigure}\hfill
  \begin{subfigure}[c]{\cw}\centering%
  \begin{tikzpicture}[spy using outlines={circle,orange,line width=0.5pt,magnification=3,size=0.7327cm, connect spies}]
    \node[name=img, inner sep=0pt, line width=0.05mm, draw=white] at (0,0) {\includegraphics[width=\textwidth]{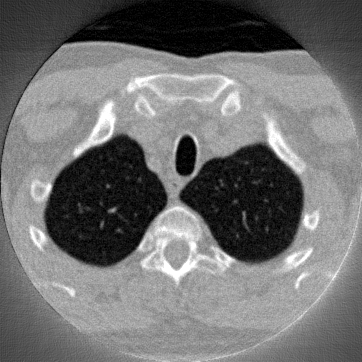}};
    \node[anchor=south west, draw=yellow, line width=0.35pt, fill=black, fill opacity=0.55, text=yellow, text opacity=1, inner sep=1.2pt, font=\tiny] at (img.south west) {34.07\,dB};
    \spy on ($(img.center)+1.181818*(0,0.6)$) in node at ($(img.north west)+(0.39,-0.39)$);
  \end{tikzpicture}\end{subfigure}\hfill
  \begin{subfigure}[c]{\cw}\centering%
  \begin{tikzpicture}[spy using outlines={circle,orange,line width=0.5pt,magnification=3,size=0.7327cm, connect spies}]
    \node[name=img, inner sep=0pt, line width=0.05mm, draw=white] at (0,0) {\includegraphics[width=\textwidth]{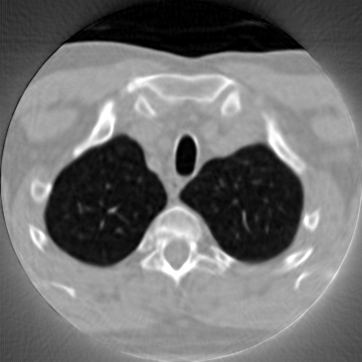}};
    \node[anchor=south west, draw=yellow, line width=0.35pt, fill=black, fill opacity=0.55, text=yellow, text opacity=1, inner sep=1.2pt, font=\tiny] at (img.south west) {36.56\,dB};
    \spy on ($(img.center)+1.181818*(0,0.6)$) in node at ($(img.north west)+(0.39,-0.39)$);
  \end{tikzpicture}\end{subfigure}\hfill
  \begin{subfigure}[c]{\cw}\centering%
  \begin{tikzpicture}[spy using outlines={circle,orange,line width=0.5pt,magnification=3,size=0.7327cm, connect spies}]
    \node[name=img, inner sep=0pt, line width=0.05mm, draw=white] at (0,0) {\includegraphics[width=\textwidth]{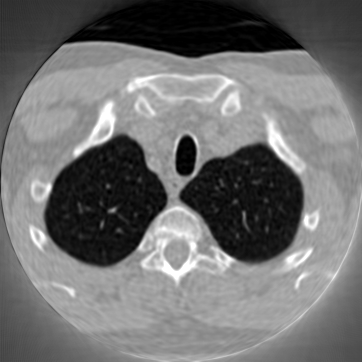}};
    \node[anchor=south west, draw=yellow, line width=0.35pt, fill=black, fill opacity=0.55, text=yellow, text opacity=1, inner sep=1.2pt, font=\tiny] at (img.south west) {\textbf{36.93\,dB}};
    \spy on ($(img.center)+1.181818*(0,0.6)$) in node at ($(img.north west)+(0.39,-0.39)$);
  \end{tikzpicture}\end{subfigure}\hfill
  \begin{subfigure}[c]{\cw}\centering%
  \begin{tikzpicture}[spy using outlines={circle,orange,line width=0.5pt,magnification=3,size=0.7327cm, connect spies}]
    \node[name=img, inner sep=0pt, line width=0.05mm, draw=white] at (0,0) {\includegraphics[width=\textwidth]{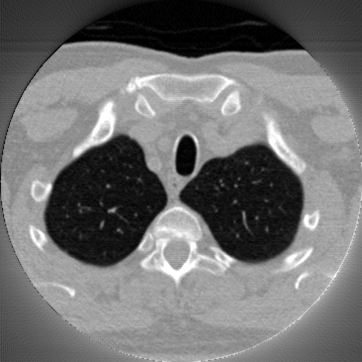}};
    \spy on ($(img.center)+1.181818*(0,0.6)$) in node at ($(img.north west)+(0.39,-0.39)$);
  \end{tikzpicture}\end{subfigure}\par\vspace{2pt}
  \begin{minipage}[c]{0.03\textwidth}\centering\rotatebox{90}{\scriptsize\textbf{Poisson ($I_0\!=\!10^5$})}\end{minipage}\hfill
  \begin{subfigure}[c]{\cw}\centering%
  \begin{tikzpicture}[spy using outlines={circle,orange,line width=0.5pt,magnification=3,size=0.7327cm, connect spies}]
    \node[name=img, inner sep=0pt, line width=0.05mm, draw=white] at (0,0) {\includegraphics[width=\textwidth]{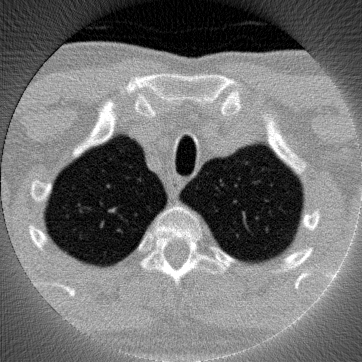}};
    \node[anchor=south west, draw=yellow, line width=0.35pt, fill=black, fill opacity=0.55, text=yellow, text opacity=1, inner sep=1.2pt, font=\tiny] at (img.south west) {30.03\,dB};
    \spy on ($(img.center)+1.181818*(0,0.6)$) in node at ($(img.north west)+(0.39,-0.39)$);
  \end{tikzpicture}\end{subfigure}\hfill
  \begin{subfigure}[c]{\cw}\centering%
  \begin{tikzpicture}[spy using outlines={circle,orange,line width=0.5pt,magnification=3,size=0.7327cm, connect spies}]
    \node[name=img, inner sep=0pt, line width=0.05mm, draw=white] at (0,0) {\includegraphics[width=\textwidth]{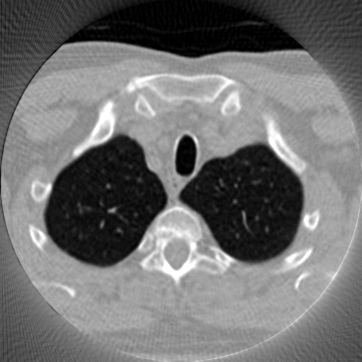}};
    \node[anchor=south west, draw=yellow, line width=0.35pt, fill=black, fill opacity=0.55, text=yellow, text opacity=1, inner sep=1.2pt, font=\tiny] at (img.south west) {35.83\,dB};
    \spy on ($(img.center)+1.181818*(0,0.6)$) in node at ($(img.north west)+(0.39,-0.39)$);
  \end{tikzpicture}\end{subfigure}\hfill
  \begin{subfigure}[c]{\cw}\centering%
  \begin{tikzpicture}[spy using outlines={circle,orange,line width=0.5pt,magnification=3,size=0.7327cm, connect spies}]
    \node[name=img, inner sep=0pt, line width=0.05mm, draw=white] at (0,0) {\includegraphics[width=\textwidth]{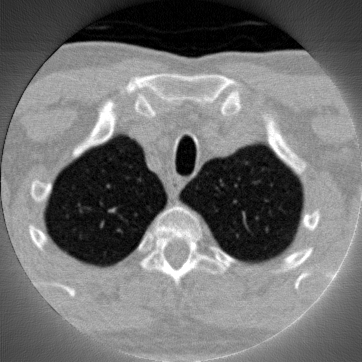}};
    \node[anchor=south west, draw=yellow, line width=0.35pt, fill=black, fill opacity=0.55, text=yellow, text opacity=1, inner sep=1.2pt, font=\tiny] at (img.south west) {35.23\,dB};
    \spy on ($(img.center)+1.181818*(0,0.6)$) in node at ($(img.north west)+(0.39,-0.39)$);
  \end{tikzpicture}\end{subfigure}\hfill
  \begin{subfigure}[c]{\cw}\centering%
  \begin{tikzpicture}[spy using outlines={circle,orange,line width=0.5pt,magnification=3,size=0.7327cm, connect spies}]
    \node[name=img, inner sep=0pt, line width=0.05mm, draw=white] at (0,0) {\includegraphics[width=\textwidth]{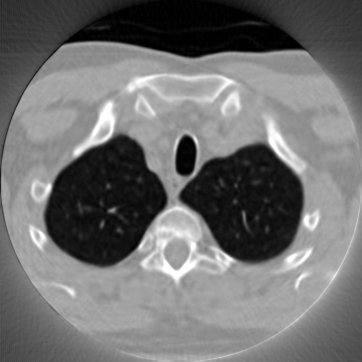}};
    \node[anchor=south west, draw=yellow, line width=0.35pt, fill=black, fill opacity=0.55, text=yellow, text opacity=1, inner sep=1.2pt, font=\tiny] at (img.south west) {37.01\,dB};
    \spy on ($(img.center)+1.181818*(0,0.6)$) in node at ($(img.north west)+(0.39,-0.39)$);
  \end{tikzpicture}\end{subfigure}\hfill
  \begin{subfigure}[c]{\cw}\centering%
  \begin{tikzpicture}[spy using outlines={circle,orange,line width=0.5pt,magnification=3,size=0.7327cm, connect spies}]
    \node[name=img, inner sep=0pt, line width=0.05mm, draw=white] at (0,0) {\includegraphics[width=\textwidth]{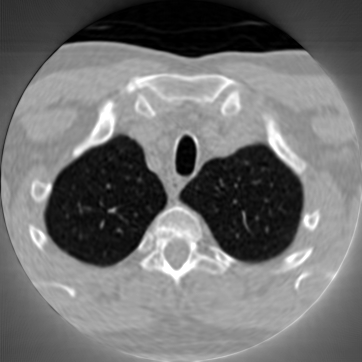}};
    \node[anchor=south west, draw=yellow, line width=0.35pt, fill=black, fill opacity=0.55, text=yellow, text opacity=1, inner sep=1.2pt, font=\tiny] at (img.south west) {\textbf{37.54\,dB}};
    \spy on ($(img.center)+1.181818*(0,0.6)$) in node at ($(img.north west)+(0.39,-0.39)$);
  \end{tikzpicture}\end{subfigure}\hfill
  \begin{subfigure}[c]{\cw}\centering%
  \begin{tikzpicture}[spy using outlines={circle,orange,line width=0.5pt,magnification=3,size=0.7327cm, connect spies}]
    \node[name=img, inner sep=0pt, line width=0.05mm, draw=white] at (0,0) {\includegraphics[width=\textwidth]{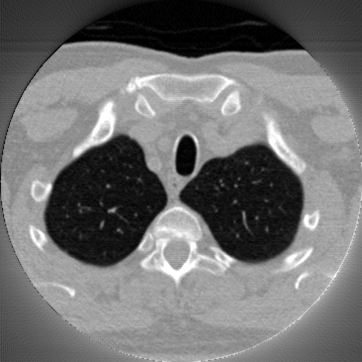}};
    \spy on ($(img.center)+1.181818*(0,0.6)$) in node at ($(img.north west)+(0.39,-0.39)$);
  \end{tikzpicture}\end{subfigure}\par\vspace{2pt}
  \makebox[0.03\textwidth]{}\hfill
  \makebox[\cw]{\scriptsize\shortstack{FBP input\\($128$ noisy views)}}\hfill
  \makebox[\cw]{\scriptsize\shortstack{only predicted \\ views}}\hfill
  \makebox[\cw]{\scriptsize\shortstack{predicted $+$ \\ measured views}}\hfill
  \makebox[\cw]{\scriptsize\shortstack{predicted $+$ \\ blind-spot (BS)}}\hfill
  \makebox[\cw]{\scriptsize\shortstack{predicted $+$ \\ complementary}}\hfill
  \makebox[\cw]{\scriptsize\shortstack{Full-view FBP\\($256$, noise-free)}}
  \caption{\textbf{Zero-shot denoising.} A TomoTransformer pretrained on noiseless data cleans projections corrupted by unseen noise, with no retraining. The blind-spot (BS) and complementary (CR) re-masking passes additionally denoise the measured views. Boxes report PSNR against the noise-free $256$-view FBP (right column).}
  \label{fig: noise_denoising}
\end{figure*}

So far we considered noiseless measurements, with models trained on noiseless sinograms. What happens when the projections are noisy at inference time? This matters in practice, where the noise distribution varies across datasets and acquisition protocols and may be unseen at training time.

\noindent\textbf{A single pass of TomoTransformer is already a denoiser.}
We consider our pre-trained TomoTransformer (varied) of Section \ref{sec: fixed_sparsity} which is trained on noiseless projections and we apply it to noisy sparse-view projections. In Figure~\ref{fig: noise_denoising} we show the results of applying the network to $128$ noisy projections to predict $256$ projections, under additive Gaussian noise at $45$~dB and under a $\mathrm{Poisson}\bigl(I_0 e^{-\A\f}\bigr)$ photon-counting model with $I_0 = 10^{5}$ incident photons per bin and a peak optical depth of $4$, log-transformed back before reconstruction. 

The second column of Figure~\ref{fig: noise_denoising} shows that the network predictions alone, without noisy measured views, not only do not show signs of hallucination under such distribution shift, but also yield reconstructions already much cleaner than the noisy FBP input (first column).
This is a remarkable property of the model, as it was never explicitly trained for denoising, yet it generalizes to this task due to its ability to capture the underlying structure of the data.
When we form the reconstruction from both predicted and measured views as usual, the residual noise in the reconstruction is dominated by the noisy measured views resulting in a lower PSNR, see the third column of Figure~\ref{fig: noise_denoising}.

The flexibility of TomoTransformer allows us to replace the noisy projections as well: \\
\noindent\textbf{Blind-spot re-masking (BS):} we partition the measured angles into $G$ uniformly interleaved subsets and, for each, mask that subset, predicting it from the remaining noisy measurements.
  The fourth column of Figure~\ref{fig: noise_denoising} shows that this approach can further improve the reconstruction quality.\\
\noindent\textbf{Complementary re-masking (CR):} we invert the mask, so that the $R_{\text{total}} - R_{\text{sparse}}$ views predicted in the first pass become the input to predict the $R_{\text{sparse}}$ measured angles. The final reconstruction is then formed by averaging the first-pass predictions and the second-pass replacements, effectively denoising both the measured and unmeasured views. The fifth column of Figure~\ref{fig: noise_denoising} shows that this approach outperforms the blind-spot re-masking.
\rev{Appendix~\ref{sec: denoising_strong} repeats this experiment under stronger noise: the improvement over FBP grows with the noise level, to about $13$\,dB, and blind-spot re-masking becomes the better of the two schemes.}

\subsection{Real experimental data}
\label{sec: real data}

\begin{figure*}[tb]
  \centering
  \begin{minipage}[c]{0.03\textwidth}\centering\rotatebox{90}{\small\textbf{Brain ($z\!=\!103$)}}\end{minipage}\hfill
  \begin{subfigure}[c]{0.188\textwidth}\centering%
  \begin{tikzpicture}[spy using outlines={circle,orange,line width=0.5pt,magnification=2,size=0.6855cm, connect spies}]
    \node[name=img, inner sep=0pt, line width=0.05mm, draw=white] at (0,0) {\includegraphics[width=\textwidth]{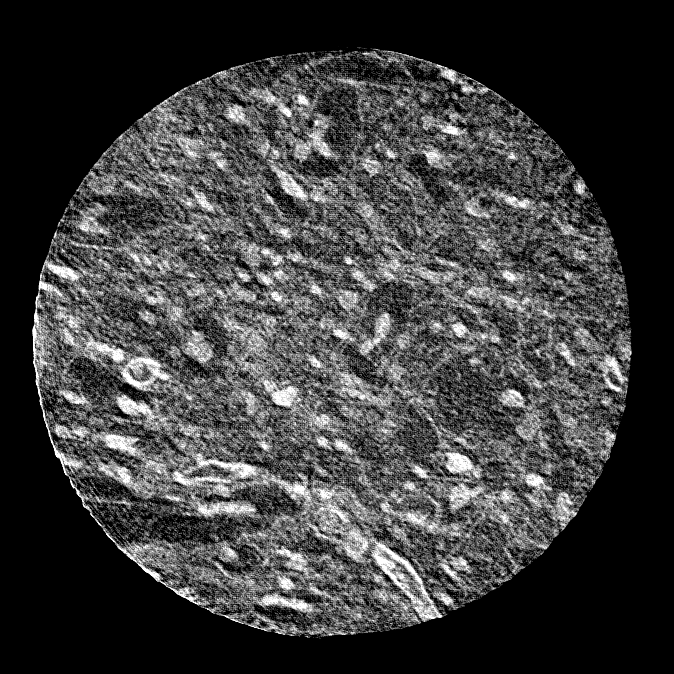}};
    \node[anchor=south west, draw=yellow, line width=0.35pt, fill=black, fill opacity=0.55, text=yellow, text opacity=1, inner sep=1.4pt, font=\scriptsize] at (img.south west) {33.3\,dB};
    \spy on ($(img.center)+1.181818*(-0.08,0.76)$) in node at ($(img.north west)+(0.3545,-0.3545)$);
  \end{tikzpicture}\end{subfigure}\hfill
  \begin{subfigure}[c]{0.188\textwidth}\centering%
  \begin{tikzpicture}[spy using outlines={circle,orange,line width=0.5pt,magnification=2,size=0.6855cm, connect spies}]
    \node[name=img, inner sep=0pt, line width=0.05mm, draw=white] at (0,0) {\includegraphics[width=\textwidth]{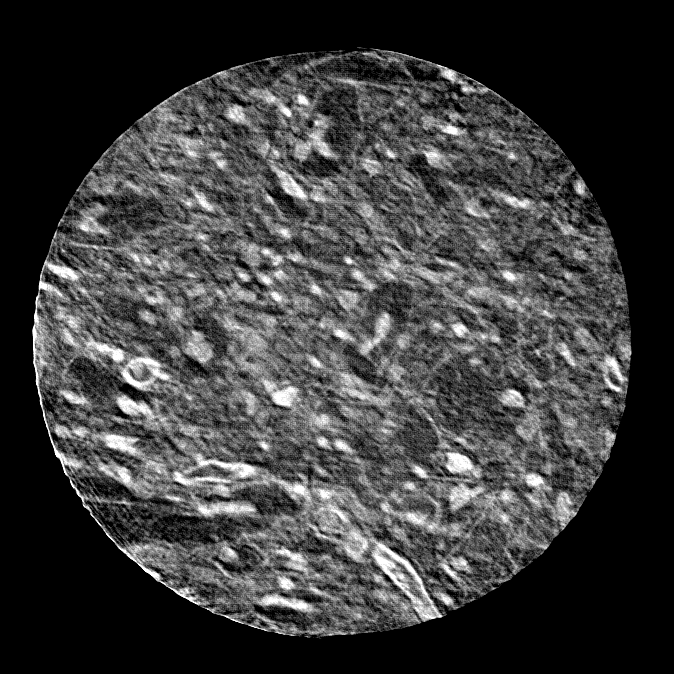}};
    \node[anchor=south west, draw=yellow, line width=0.35pt, fill=black, fill opacity=0.55, text=yellow, text opacity=1, inner sep=1.4pt, font=\scriptsize] at (img.south west) {38.0\,dB};
    \spy on ($(img.center)+1.181818*(-0.08,0.76)$) in node at ($(img.north west)+(0.3545,-0.3545)$);
  \end{tikzpicture}\end{subfigure}\hfill
  \begin{subfigure}[c]{0.188\textwidth}\centering%
  \begin{tikzpicture}[spy using outlines={circle,orange,line width=0.5pt,magnification=2,size=0.6855cm, connect spies}]
    \node[name=img, inner sep=0pt, line width=0.05mm, draw=white] at (0,0) {\includegraphics[width=\textwidth]{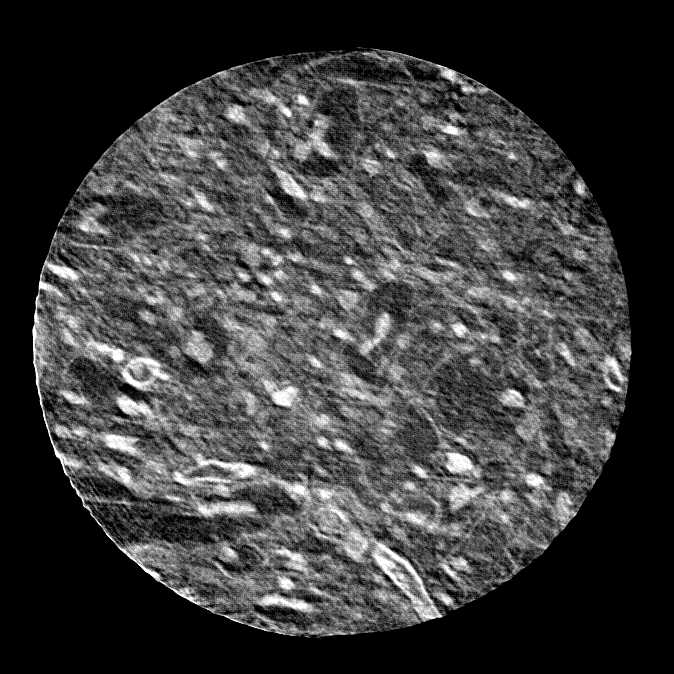}};
    \node[anchor=south west, draw=yellow, line width=0.35pt, fill=black, fill opacity=0.55, text=yellow, text opacity=1, inner sep=1.4pt, font=\scriptsize] at (img.south west) {38.2\,dB};
    \spy on ($(img.center)+1.181818*(-0.08,0.76)$) in node at ($(img.north west)+(0.3545,-0.3545)$);
  \end{tikzpicture}\end{subfigure}\hfill
  \begin{subfigure}[c]{0.188\textwidth}\centering%
  \begin{tikzpicture}[spy using outlines={circle,orange,line width=0.5pt,magnification=2,size=0.6855cm, connect spies}]
    \node[name=img, inner sep=0pt, line width=0.05mm, draw=white] at (0,0) {\includegraphics[width=\textwidth]{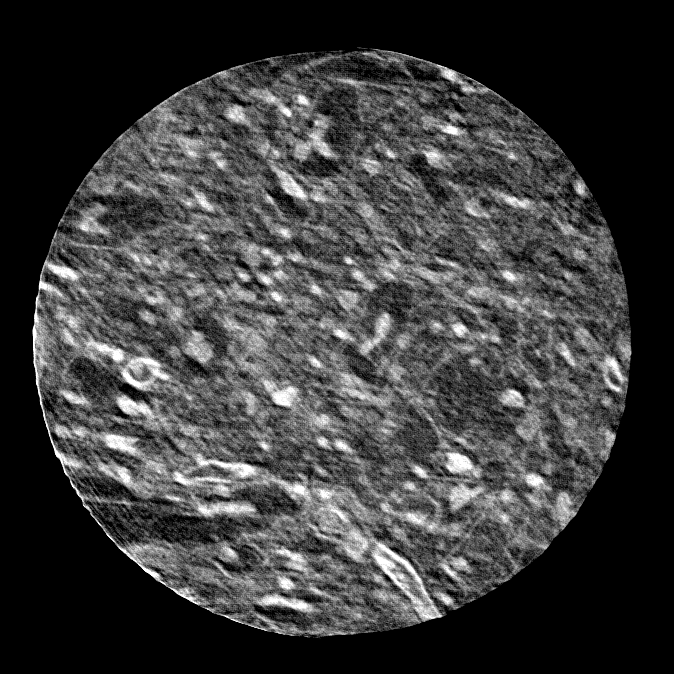}};
    \node[anchor=south west, draw=yellow, line width=0.35pt, fill=black, fill opacity=0.55, text=yellow, text opacity=1, inner sep=1.4pt, font=\scriptsize] at (img.south west) {38.5\,dB};
    \spy on ($(img.center)+1.181818*(-0.08,0.76)$) in node at ($(img.north west)+(0.3545,-0.3545)$);
  \end{tikzpicture}\end{subfigure}\hfill
  \begin{subfigure}[c]{0.188\textwidth}\centering%
  \begin{tikzpicture}[spy using outlines={circle,orange,line width=0.5pt,magnification=2,size=0.6855cm, connect spies}]
    \node[name=img, inner sep=0pt, line width=0.05mm, draw=white] at (0,0) {\includegraphics[width=\textwidth]{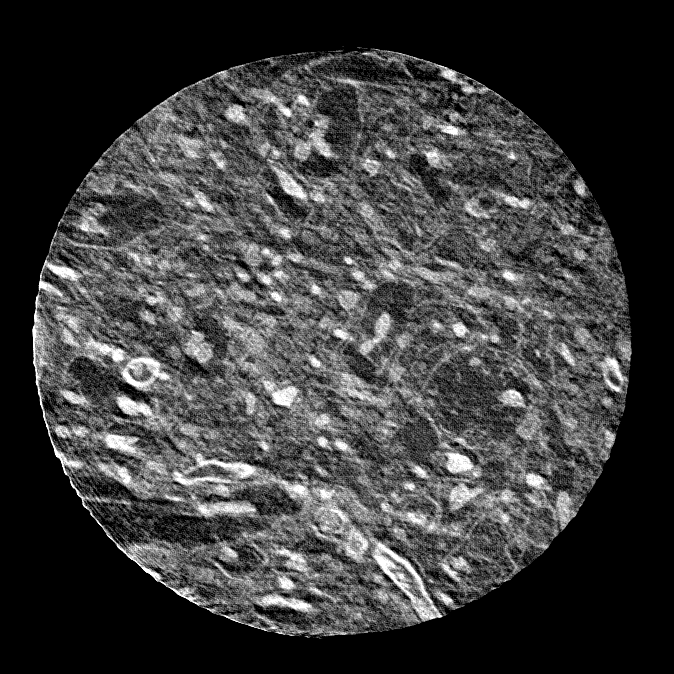}};
    \spy on ($(img.center)+1.181818*(-0.08,0.76)$) in node at ($(img.north west)+(0.3545,-0.3545)$);
  \end{tikzpicture}\end{subfigure}\par\vspace{2pt}
  \begin{minipage}[c]{0.03\textwidth}\centering\rotatebox{90}{\small\textbf{Chip ($z\!=\!435$)}}\end{minipage}\hfill
  \begin{subfigure}[c]{0.188\textwidth}\centering%
  \begin{tikzpicture}[spy using outlines={circle,orange,line width=0.5pt,magnification=3,size=1.0282cm, connect spies}]
    \node[name=img, inner sep=0pt, line width=0.05mm, draw=white] at (0,0) {\includegraphics[width=\textwidth]{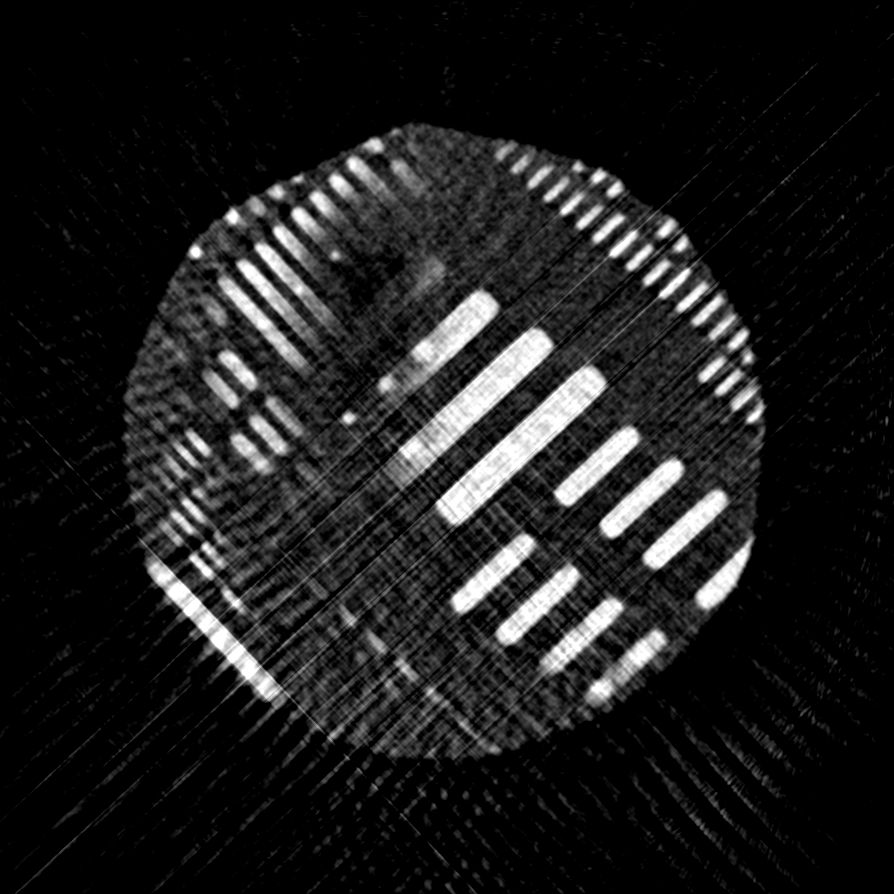}};
    \node[anchor=south west, draw=yellow, line width=0.35pt, fill=black, fill opacity=0.55, text=yellow, text opacity=1, inner sep=1.4pt, font=\scriptsize] at (img.south west) {24.2\,dB};
    \spy on ($(img.center)+1.181818*(0.16,-0.48)$) in node at ($(img.north west)+(0.5318,-0.5318)$);
  \end{tikzpicture}\end{subfigure}\hfill
  \begin{subfigure}[c]{0.188\textwidth}\centering%
  \begin{tikzpicture}[spy using outlines={circle,orange,line width=0.5pt,magnification=3,size=1.0282cm, connect spies}]
    \node[name=img, inner sep=0pt, line width=0.05mm, draw=white] at (0,0) {\includegraphics[width=\textwidth]{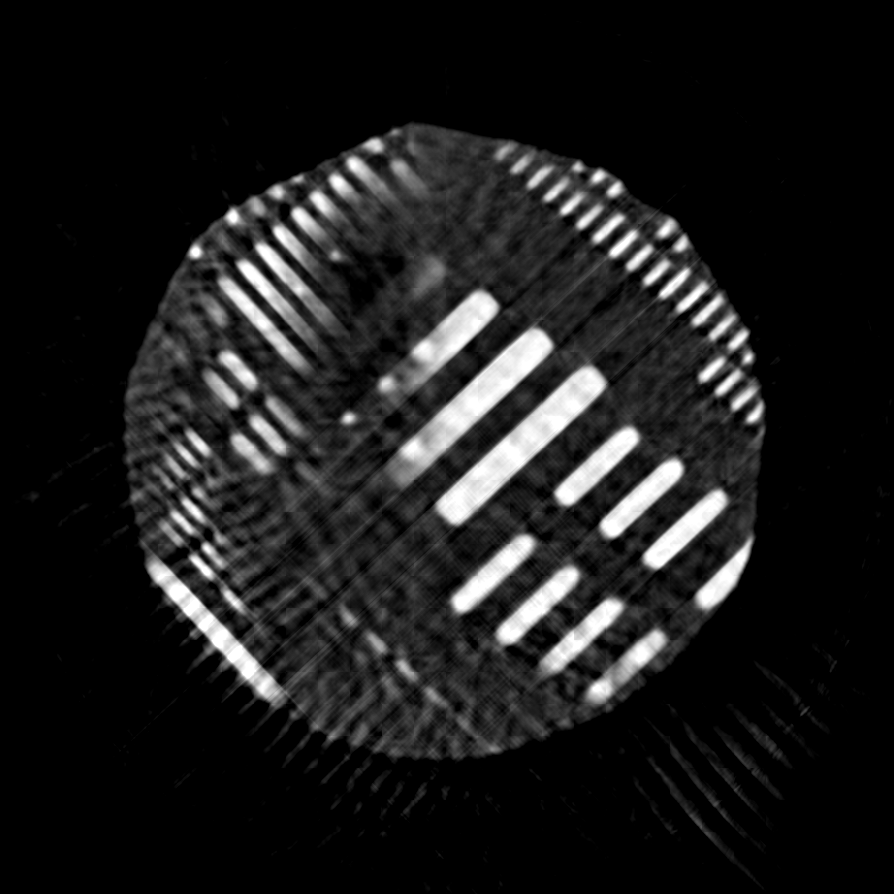}};
    \node[anchor=south west, draw=yellow, line width=0.35pt, fill=black, fill opacity=0.55, text=yellow, text opacity=1, inner sep=1.4pt, font=\scriptsize] at (img.south west) {27.9\,dB};
    \spy on ($(img.center)+1.181818*(0.16,-0.48)$) in node at ($(img.north west)+(0.5318,-0.5318)$);
  \end{tikzpicture}\end{subfigure}\hfill
  \begin{subfigure}[c]{0.188\textwidth}\centering%
  \begin{tikzpicture}[spy using outlines={circle,orange,line width=0.5pt,magnification=3,size=1.0282cm, connect spies}]
    \node[name=img, inner sep=0pt, line width=0.05mm, draw=white] at (0,0) {\includegraphics[width=\textwidth]{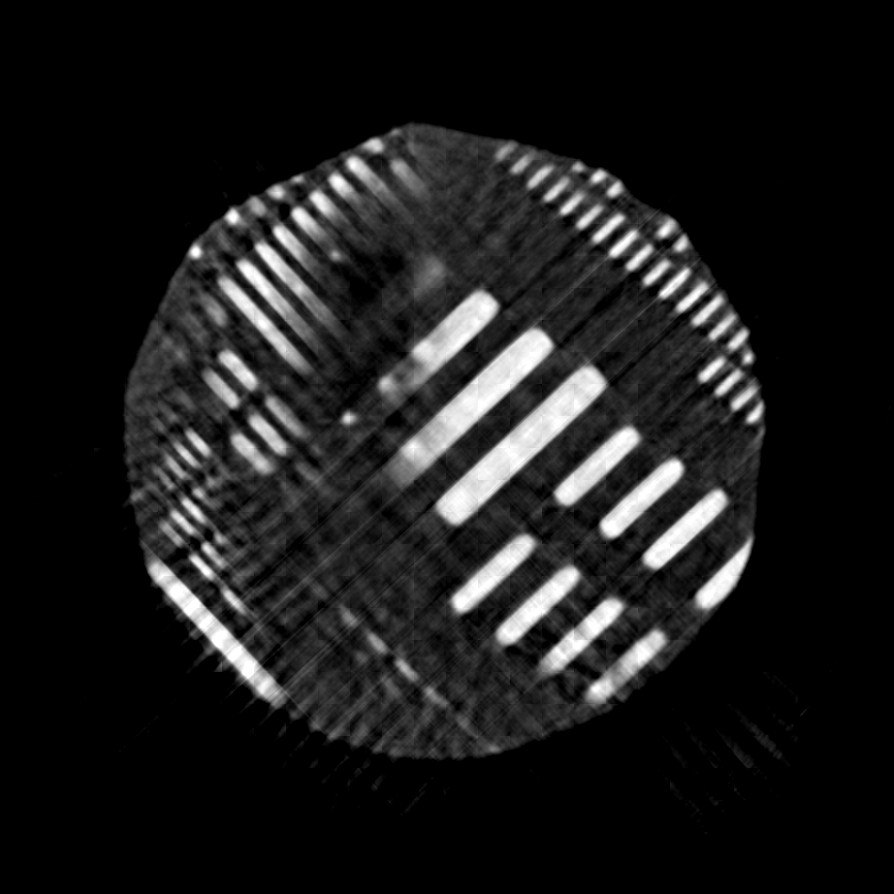}};
    \node[anchor=south west, draw=yellow, line width=0.35pt, fill=black, fill opacity=0.55, text=yellow, text opacity=1, inner sep=1.4pt, font=\scriptsize] at (img.south west) {29.5\,dB};
    \spy on ($(img.center)+1.181818*(0.16,-0.48)$) in node at ($(img.north west)+(0.5318,-0.5318)$);
  \end{tikzpicture}\end{subfigure}\hfill
  \begin{subfigure}[c]{0.188\textwidth}\centering%
  \begin{tikzpicture}[spy using outlines={circle,orange,line width=0.5pt,magnification=3,size=1.0282cm, connect spies}]
    \node[name=img, inner sep=0pt, line width=0.05mm, draw=white] at (0,0) {\includegraphics[width=\textwidth]{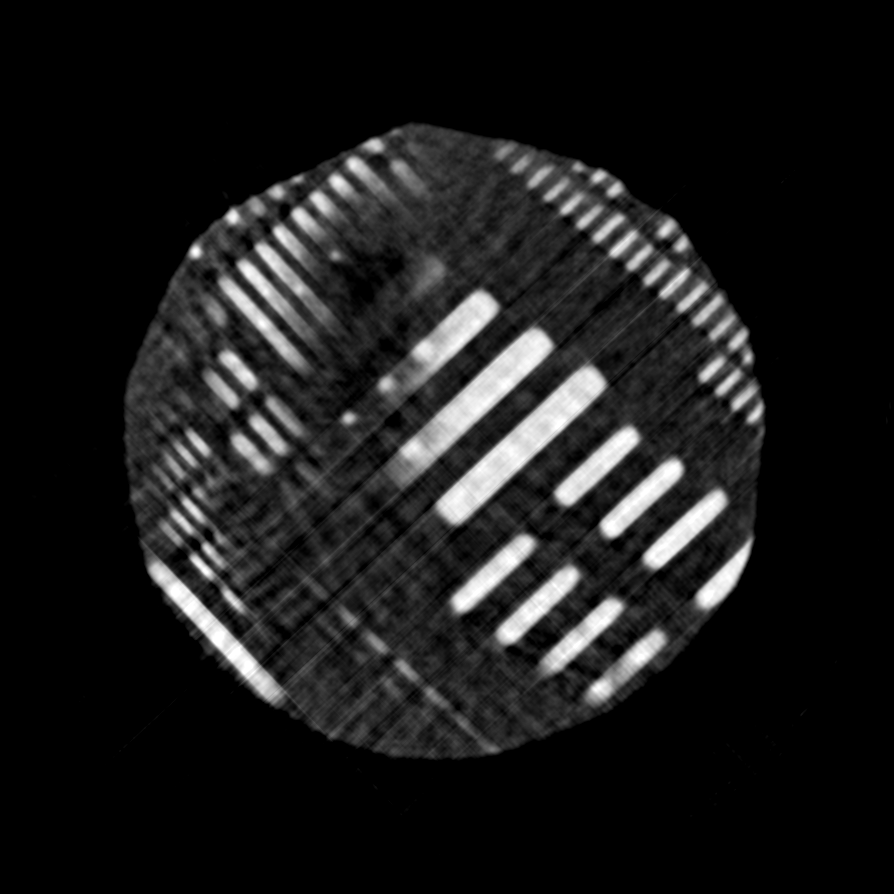}};
    \node[anchor=south west, draw=yellow, line width=0.35pt, fill=black, fill opacity=0.55, text=yellow, text opacity=1, inner sep=1.4pt, font=\scriptsize] at (img.south west) {30.3\,dB};
    \spy on ($(img.center)+1.181818*(0.16,-0.48)$) in node at ($(img.north west)+(0.5318,-0.5318)$);
  \end{tikzpicture}\end{subfigure}\hfill
  \begin{subfigure}[c]{0.188\textwidth}\centering%
  \begin{tikzpicture}[spy using outlines={circle,orange,line width=0.5pt,magnification=3,size=1.0282cm, connect spies}]
    \node[name=img, inner sep=0pt, line width=0.05mm, draw=white] at (0,0) {\includegraphics[width=\textwidth]{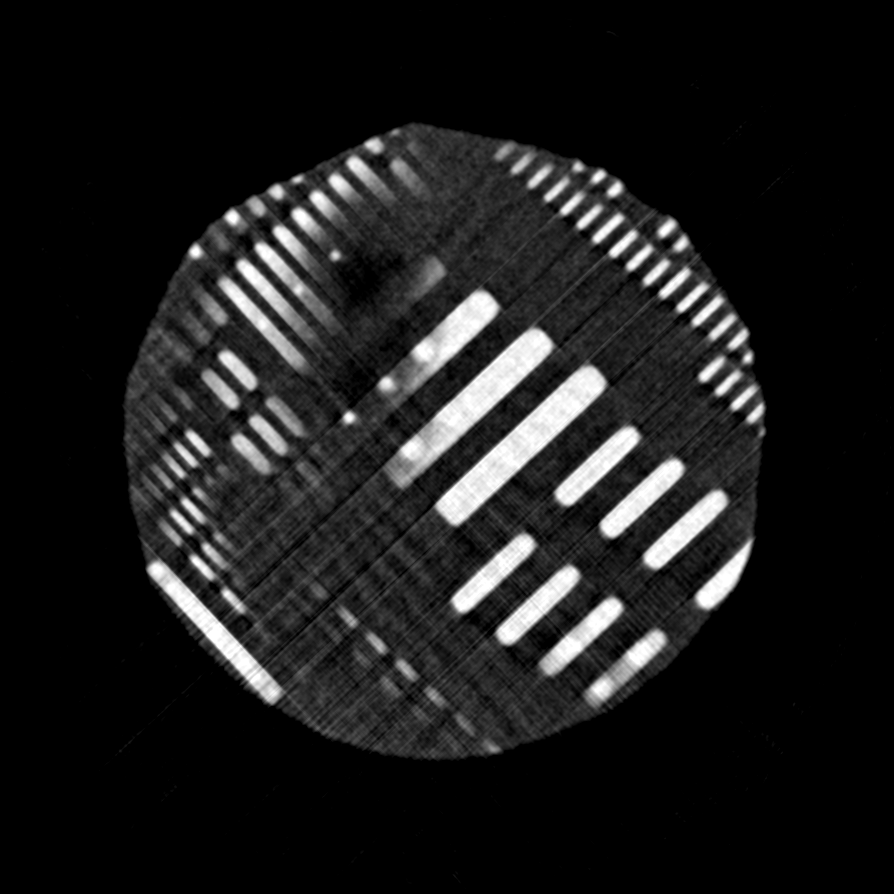}};
    \spy on ($(img.center)+1.181818*(0.16,-0.48)$) in node at ($(img.north west)+(0.5318,-0.5318)$);
  \end{tikzpicture}\end{subfigure}\par\vspace{2pt}
  \makebox[0.03\textwidth]{}\hfill
  \makebox[0.188\textwidth]{\small\shortstack{FBP input\\($128$ / $86$ views)}}\hfill
  \makebox[0.188\textwidth]{\small\shortstack{Zero-shot\\(${\rightarrow}256$)}}\hfill
  \makebox[0.188\textwidth]{\small\shortstack{Self-supervised\\(${\rightarrow}256$)}}\hfill
  \makebox[0.188\textwidth]{\small\shortstack{Self-supervised\\+ RCS (${\rightarrow}256$)}}\hfill
  \makebox[0.188\textwidth]{\small\shortstack{Full-view FBP\\($256$)}}
  \caption{\rev{Performance of TomoTransformer on two real experimental PXCT scans: a nanoscale brain (top, $128$ measured views) and a  chip (bottom, $86$ measured views). The $256$-view image is the reference. Boxes report per-slice PSNR (dB) against it.}}
  \label{fig: ssl_real}
\end{figure*}

We trained TomoTransformer (varied) on the heterogeneous dataset of CT and natural images at their original resolutions and varying view counts (the eight datasets of Section~\ref{sec: exp_setup} with the sampling scheme of Section~\ref{sec: variable_sparsity}). We already reported its performance in Section \ref{sec: fixed_sparsity} (see Tables \ref{tab: fixed_sparsity_psnr} and \ref{tab: fixed_sparsity_ssim}). To test zero-shot performance, now we apply it to \rev{two real experimental datasets from} a completely different domain: \rev{nanoscale ptychographic X-ray computed tomography (PXCT)}, which differs radically from medical CT in scale, contrast, and resolution. Even though this modality measures the refractive index instead of standard X-ray absorption, the physics still relies on the Radon transform. \rev{The first sample is} a mouse-brain pillar~\citep{bosch2025nondestructive} ($617$ views over $[0^\circ,180^\circ)$, $37.6\,\text{nm}$ pixels, $768$-pixel detector \rev{, $448$ axial slices}). \rev{The second is a commercial processor chip from the publicly released $4$-nm-resolution dataset of~\citet{aidukas2024high} ($1594$ views over $180^\circ$, $1792$-pixel detector, $4.21\,\text{nm}$ pixels, $736$ axial slices). Its high-contrast metal interconnects with sharp edges makes it completely different from the training set and streak far more severely under sparse sampling than brain tissue. We bin its detector by two ($896$ pixels, $8.4\,\text{nm}$) and evaluate the $24$ slices of its structured interconnect layers. Both datasets were collected at the cSAXS beamline (SLS/PSI). For both samples, the sparse FBP of the measured views is the input ($128$ views for the brain; $86$ for the chip, every third view of the $256$-view grid, to test a sparser setting), which TomoTransformer completes to a $256$-view grid,} and the dense FBP of the same measurements is the reference.

Figure~\ref{fig: ssl_real} \rev{(second column)} and Table~\ref{tab: ssl_real} (Appendix) show that TomoTransformer generalizes seamlessly to this out-of-distribution data with no fine-tuning. \rev{We now use self-supervised fine-tuning to adapt the pre-trained TomoTransformer to each scan as explained in Section~\ref{sec: ssl}, using only its measured views. Figure~\ref{fig: ssl_real} (third column) and Table~\ref{tab: ssl_real} (Appendix) show that the self-supervised fine-tuning improves the reconstructions, and much more so on the chip, which is further from the training distribution. In addition, RCS inference could effectively kill the blocking artifacts in the chip reconstruction, look at the zoomed-in area of Figure~\ref{fig: ssl_real} (bottom row, third and fourth column). 
Please refer to Appendix~\ref{sec: more_experiments_real} for the fine-tuning details and additional results.}

\section{Conclusion}
In this paper, we introduced TomoTransformer, a novel transformer-based architecture for sparse-view CT reconstruction. By disentangling the representation of projection angles and local back-projection patches, our model achieves robust performance across a variety of datasets and sparsity levels. Notably, TomoTransformer demonstrates remarkable generalization capabilities, effectively handling out-of-distribution data without fine-tuning. Future work will explore designing this architecture for other tomographic modalities including cone-beam CT.

\newpage

\bibliographystyle{plainnat}
\bibliography{references}

\newpage

\appendix

\section{TomoTransformer architecture}
\label{sec: tomo_arch_details}

\begin{table*}[t]
\caption{Fixed sparse-view reconstruction ($64\!\rightarrow\!256$): \textbf{SSIM} per dataset (against the dense $256$-view reference) for TomoTransformer and the baselines. \rev{RCS and + supervised fine-tuning as in Table~\ref{tab: fixed_sparsity_psnr}.} Best per column in \textbf{bold}.}
\centering
\renewcommand\arraystretch{1.25}
\resizebox{\textwidth}{!}{%
\begin{tabular}{c|cccccccc|c}
\hline
Method & LoDoPaB & COVID-19 & KiTS23 & MSD-T10 & COLONOG & HNSCC & FFHQ & ImageNet & Average \\
\hline
FBP~\citep{kak1988principles} & 0.635 & 0.654 & 0.662 & 0.554 & 0.610 & 0.395 & 0.631 & 0.598 & 0.592 \\
U-Net~\citep{ronneberger2015unet} & 0.895 & 0.919 & 0.948 & 0.919 & 0.889 & 0.895 & 0.857 & 0.836 & 0.895 \\
DRUNet~\citep{zhang2021plug} & 0.910 & 0.937 & 0.965 & 0.939 & 0.907 & 0.945 & 0.892 & 0.871 & 0.921 \\
NAFNet~\citep{chen2022simple} & 0.914 & 0.941 & 0.968 & 0.943 & 0.911 & 0.954 & 0.900 & 0.877 & 0.926 \\
Restormer~\citep{zamir2022restormer} & 0.915 & 0.943 & 0.969 & 0.944 & 0.912 & 0.955 & 0.902 & 0.879 & 0.927 \\
CTformer~\citep{wang2023ctformer} & 0.868 & 0.908 & 0.908 & 0.871 & 0.855 & 0.815 & 0.856 & 0.809 & 0.861 \\
\rev{CIN~\citep{guo2023efficient}} & \rev{0.903} & \rev{0.936} & \rev{0.965} & \rev{0.939} & \rev{0.906} & \rev{0.944} & \rev{0.893} & \rev{0.868} & \rev{0.919} \\
\hline
TomoTransformer (fixed) & 0.913 & 0.946 & 0.971 & 0.948 & 0.914 & 0.959 & 0.918 & 0.878 & 0.931 \\
\rev{TomoTransformer (fixed, RCS)} & \rev{\textbf{0.918}} & \rev{\textbf{0.951}} & \rev{\textbf{0.974}} & \rev{\textbf{0.953}} & \rev{\textbf{0.919}} & \rev{\textbf{0.972}} & \rev{\textbf{0.927}} & \rev{\textbf{0.885}} & \rev{\textbf{0.937}} \\
\hline
TomoTransformer (varied) & 0.907 & 0.941 & 0.964 & 0.937 & 0.908 & 0.920 & 0.908 & 0.872 & 0.920 \\
\rev{TomoTransformer (varied, RCS)} & \rev{0.913} & \rev{0.948} & \rev{0.970} & \rev{0.946} & \rev{0.914} & \rev{0.944} & \rev{0.919} & \rev{0.881} & \rev{0.929} \\
\rev{+ supervised fine-tuning} & \rev{0.915} & \rev{0.949} & \rev{0.973} & \rev{0.950} & \rev{0.917} & \rev{0.961} & \rev{0.921} & \rev{0.884} & \rev{0.934} \\
\hline
\end{tabular}}
\label{tab: fixed_sparsity_ssim}
\end{table*}

\begin{table*}[t]
\caption{\rev{The checkpoints of Tables~\ref{tab: fixed_sparsity_psnr} and~\ref{tab: fixed_sparsity_ssim}, evaluated \emph{without retraining} at $128\!\rightarrow\!256$ ($128$ measured views of the same $256$-view grid), on the same test volumes and against the same dense $256$-view reference: \textbf{PSNR} (dB, top) and \textbf{SSIM} (bottom) per dataset. Best per column in \textbf{bold}.}}
\centering
\renewcommand\arraystretch{1.25}
\resizebox{\textwidth}{!}{%
\begin{tabular}{c|cccccccc|c}
\hline
Method & LoDoPaB & COVID-19 & KiTS23 & MSD-T10 & COLONOG & HNSCC & FFHQ & ImageNet & Average \\
\hline
\multicolumn{10}{c}{PSNR (dB)} \\
\hline
FBP~\citep{kak1988principles} & 36.68 & 34.95 & 37.87 & 35.51 & 35.46 & 35.80 & 34.89 & 34.08 & 35.65 \\
U-Net~\citep{ronneberger2015unet} & 38.56 & 39.24 & 42.62 & 40.71 & 38.65 & 40.45 & 37.60 & 34.37 & 39.02 \\
DRUNet~\citep{zhang2021plug} & 40.07 & 40.46 & 44.62 & 41.94 & 39.52 & 41.80 & 38.61 & 36.36 & 40.42 \\
NAFNet~\citep{chen2022simple} & 40.39 & 40.65 & 44.96 & 42.15 & 39.66 & 42.03 & 38.90 & 37.15 & 40.74 \\
Restormer~\citep{zamir2022restormer} & 40.49 & 40.69 & 45.02 & 42.22 & 39.73 & 42.05 & 39.00 & 37.32 & 40.82 \\
CTformer~\citep{wang2023ctformer} & 35.50 & 34.57 & 35.88 & 35.35 & 35.01 & 32.19 & 35.53 & 34.10 & 34.77 \\
\hline
TomoTransformer (fixed) & 40.66 & 40.77 & 44.06 & 41.67 & 40.21 & 43.02 & 39.02 & 37.38 & 40.85 \\
TomoTransformer (fixed, RCS) & 39.81 & 40.34 & 43.23 & 40.95 & 39.74 & 42.27 & 38.42 & 36.54 & 40.16 \\
\hline
TomoTransformer (varied) & 42.16 & 42.93 & 47.20 & 44.58 & 41.52 & 44.08 & 41.36 & 39.07 & 42.86 \\
TomoTransformer (varied, RCS) & \textbf{42.57} & \textbf{43.59} & \textbf{47.93} & \textbf{45.19} & \textbf{41.89} & \textbf{44.65} & \textbf{41.97} & \textbf{39.63} & \textbf{43.43} \\
\hline
\multicolumn{10}{c}{SSIM} \\
\hline
FBP~\citep{kak1988principles} & 0.873 & 0.857 & 0.882 & 0.822 & 0.845 & 0.711 & 0.850 & 0.847 & 0.836 \\
U-Net~\citep{ronneberger2015unet} & 0.918 & 0.937 & 0.963 & 0.941 & 0.912 & 0.911 & 0.918 & 0.894 & 0.924 \\
DRUNet~\citep{zhang2021plug} & 0.930 & 0.945 & 0.971 & 0.949 & 0.919 & 0.933 & 0.928 & 0.919 & 0.937 \\
NAFNet~\citep{chen2022simple} & 0.934 & 0.946 & 0.973 & 0.951 & 0.921 & 0.937 & 0.931 & 0.928 & 0.940 \\
Restormer~\citep{zamir2022restormer} & 0.935 & 0.947 & 0.973 & 0.951 & 0.922 & 0.937 & 0.932 & 0.930 & 0.941 \\
CTformer~\citep{wang2023ctformer} & 0.826 & 0.810 & 0.835 & 0.822 & 0.816 & 0.559 & 0.848 & 0.852 & 0.796 \\
\hline
TomoTransformer (fixed) & 0.941 & 0.949 & 0.962 & 0.937 & 0.931 & 0.915 & 0.935 & 0.926 & 0.937 \\
TomoTransformer (fixed, RCS) & 0.945 & 0.956 & 0.970 & 0.946 & 0.938 & 0.936 & 0.944 & 0.932 & 0.946 \\
\hline
TomoTransformer (varied) & 0.957 & 0.968 & 0.984 & 0.971 & 0.950 & 0.967 & 0.961 & 0.951 & 0.964 \\
TomoTransformer (varied, RCS) & \textbf{0.960} & \textbf{0.973} & \textbf{0.987} & \textbf{0.975} & \textbf{0.953} & \textbf{0.978} & \textbf{0.967} & \textbf{0.958} & \textbf{0.969} \\
\hline
\end{tabular}}
\label{tab: fixed_sparsity_128}
\end{table*}

\begin{table}[t]
\caption{Varied sparse-view reconstruction on LoDoPaB-CT ($\rightarrow\!256$ views), against the dense $256$-view FBP reference (\textbf{PSNR} in dB\,/\,\textbf{SSIM}, averaged over the $32$ test slices). The last row is the very-sparse regime, with only $64$ measured views---the extreme sparse end of the training distribution. \rev{RCS: the same checkpoint assembled with ray-consistent stitching (Section~\ref{sec: sino_aggregation}).} Best per column in \textbf{bold}.}
\centering
\renewcommand\arraystretch{1.45}
\begin{tabular}{c|cc|cc|cc|cc}
\hline
 & \multicolumn{2}{c|}{FBP} & \multicolumn{2}{c|}{ViewTrans} & \multicolumn{2}{c|}{TomoTransformer} & \multicolumn{2}{c}{\rev{TomoTransformer (RCS)}} \\
Target & PSNR & SSIM & PSNR & SSIM & PSNR & SSIM & \rev{PSNR} & \rev{SSIM} \\
\hline
$128\!\rightarrow\!256$ & 36.68 & 0.873 & 37.75 & 0.914 & 42.40 & 0.959 & \rev{\textbf{42.84}} & \rev{\textbf{0.963}} \\
\hline
$64\!\rightarrow\!256$ & 29.24 & 0.635 & 33.07 & 0.832 & 37.68 & 0.910 & \rev{\textbf{37.99}} & \rev{\textbf{0.916}} \\
\hline
\end{tabular}
\label{tab: variable_sparsity}
\end{table}

Each token is a local $P \times P$ patch ($P = 32$ pixels) cropped from a single-view back-projection at its native resolution. Our geometric tokenizer (Section~\ref{sec: Geometric tokenizer}) samples each patch along its maximum-$\tau$ diagonal chord at $K = \lceil P\sqrt{2}\rceil = 46$ equispaced points using bilinear interpolation. This reduces the $P^2 = 1024$ pixels of the patch into a $46$-dimensional feature vector. A linear layer $\mathbf{E} \in \mathbb{R}^{D \times K}$ then projects this feature to the transformer embedding dimension $D = 512$. The detokenizer uses a linear layer $\mathbf{E}' \in \mathbb{R}^{K \times D}$ followed by the reconstruction sweep described in Section~\ref{sec: Geometric tokenizer}. A single tokenizer-detokenizer pair is shared across all patch locations and projection views.

To represent geometry, each angle $\theta_r$ is mapped to Fourier features $\bigl[\sin(k\theta_r),\, \cos(k\theta_r)\bigr]_{k=1}^{F}$ using $F = 32$ frequencies, followed by a linear projection $\mathbb{R}^{2F} \to \mathbb{R}^{D}$. This positional embedding is added at two points: (1) to the encoder input and (2) to the complete decoder input. Each addition is scaled by a learnable scalar gate ($\alpha_1$, initialized to $0$, and $\alpha_2$, initialized to $0.5$), allowing the model to adaptively weigh the importance of absolute angles at each stage.

Our backbone follows a Masked Autoencoder (MAE) design~\citep{he2022masked}. The encoder processes \emph{only} the observed measured tokens. The decoder then receives the complete sequence, where measured slots contain the encoder outputs and target query positions are filled with a learnable mask token (initialized from $\mathcal{N}(0, 0.02^2)$). Both encoder and decoder consist of $4$ transformer blocks with width $D = 512$, $H = 8$ attention heads ($64$ dimensions per head), GELU activation and an MLP ratio of $1.0$. All linear layers use Xavier-uniform weight initialization with zero biases.

Because the geometric relationship between any pair of tokens is known, we supply this geometric prior directly to the attention mechanism rather than relying on the model to infer it from content alone. For each pair of tokens $(r,r')$ we form the angular difference $\Delta_{rr'} = \theta_r - \theta_{r'}$, wrapped to $(-\pi,\pi]$, and add a learned function of it to the pre-softmax logits:
\begin{equation}
    A^{(h)}_{rr'} \;=\; \frac{\bigl\langle \mathbf{q}^{(h)}_r,\, \mathbf{k}^{(h)}_{r'} \bigr\rangle}{\sqrt{D_h}} \;+\; \Bigl[\, g_{\phi}\bigl(\sin \Delta_{rr'},\, \cos \Delta_{rr'}\bigr) \Bigr]_h ,
    \qquad
    \mathbf{a}^{(h)}_r = \operatorname{softmax}_{r'}\bigl(A^{(h)}_{rr'}\bigr),
    \label{eq: geometric_attention}
\end{equation}
where $\mathbf{q}^{(h)}_r$ and $\mathbf{k}^{(h)}_{r'}$ denote the query and key projections of head $h$, $D_h = D/H = 64$ is the per-head dimension, $\mathbf{a}^{(h)}_r$ collects the resulting attention weights of token $r$, and $g_{\phi} : \R^{2} \to \R^{H}$ is a lightweight two-layer GELU MLP ($\mathbb{R}^{2} \to \mathbb{R}^{32} \to \mathbb{R}^{H}$) shared across all token pairs that outputs a scalar bias for each of the $H$ attention heads. We initialize the final layer of $g_{\phi}$ to zero so that training begins from standard dot-product attention. We call this the geometric attention mechanism (GAM), and our results show that this physics-informed attention mechanism slightly enhances reconstruction quality and convergence speed, see Tables \ref{tab: GAM psnr} and \ref{tab: GAM ssim}.

\begin{table*}[t]
\caption{Ablation studies on varied sparse-view reconstruction ($64\!\rightarrow\!256$): \textbf{PSNR} (dB) per dataset (against the dense $256$-view reference). The two TomoTransformer variants are identical except for the geometric attention bias of Eq.~\eqref{eq: geometric_attention}. Best per column in \textbf{bold}.}
\centering
\renewcommand\arraystretch{1.25}
\resizebox{\textwidth}{!}{%
\begin{tabular}{c|cccccccc|c}
\hline
Method & LoDoPaB & COVID-19 & KiTS23 & MSD-T10 & COLONOG & HNSCC & FFHQ & ImageNet & Average \\
\hline
FBP~\citep{kak1988principles} & 29.24 & 28.48 & 30.27 & 28.13 & 28.59 & 28.93 & 27.97 & 26.46 & 28.51 \\
TomoTransformer (standard) & 37.41 & 38.92 & 42.38 & 39.77 & 38.08 & 39.34 & 36.06 & 33.37 & 38.17 \\
TomoTransformer (GAM) & \textbf{37.47} & \textbf{39.05} & \textbf{42.68} & \textbf{40.02} & \textbf{38.18} & \textbf{39.61} & \textbf{36.11} & \textbf{33.42} & \textbf{38.32} \\
\hline
\end{tabular}}
\label{tab: GAM psnr}
\end{table*}

\begin{table*}[t]
\caption{Ablation studies on varied sparse-view reconstruction ($64\!\rightarrow\!256$): \textbf{SSIM} per dataset (against the dense $256$-view reference). Best per column in \textbf{bold}.}
\centering
\renewcommand\arraystretch{1.25}
\resizebox{\textwidth}{!}{%
\begin{tabular}{c|cccccccc|c}
\hline
Method & LoDoPaB & COVID-19 & KiTS23 & MSD-T10 & COLONOG & HNSCC & FFHQ & ImageNet & Average \\
\hline
FBP~\citep{kak1988principles} & 0.635 & 0.654 & 0.662 & 0.554 & 0.610 & 0.395 & 0.631 & 0.598 & 0.592 \\
TomoTransformer (standard) & 0.906 & 0.940 & 0.963 & 0.935 & 0.907 & 0.917 & 0.907 & 0.871 & 0.918 \\
TomoTransformer (GAM) & \textbf{0.907} & \textbf{0.941} & \textbf{0.964} & \textbf{0.937} & \textbf{0.908} & \textbf{0.920} & \textbf{0.908} & \textbf{0.872} & \textbf{0.920} \\
\hline
\end{tabular}}
\label{tab: GAM ssim}
\end{table*}

\section{Data simulation and training}
\label{sec: training_details}

In each training iteration, we sample a batch of $32$ axial slices at their native resolutions from the dataset mixture described in Section~\ref{sec: exp_setup}. Each slice is normalized to a common intensity range and masked with a circular field-of-view mask. We compute parallel-beam 2D projections using the Operator Discretization Library (ODL) \citep{adler2017odl}, setting the detector width to match the image dimension. The resulting sinogram is ramp-filtered and back-projected view by view to create the disentangled tensor stack $\{\bb_r\}$, from which input patches are cropped.

We evaluate two sparse-view sampling protocols. Under the \emph{fixed} protocol (Section~\ref{sec: fixed_sparsity}), full-view projections are placed on an equispaced grid of $R_{\text{total}} = 256$ angles over $[0^\circ, 180^\circ)$. Taking every fourth view yields $R_{\text{sparse}} = 64$ measured angles for the encoder, leaving the remaining $192$ views as prediction targets. Under the \emph{varied} protocol (Sections~\ref{sec: variable_sparsity} and~\ref{sec: real data}), we resample the full-view and sparse-view at every iteration: full-view projections are drawn uniformly from $[128, 512]$, each angle is perturbed by i.i.d.\ Gaussian jitter of standard deviation $\sigma\delta$, where $\delta = \pi/R_{\text{total}}$ is the nominal angular spacing and $\sigma = 0.05$, and the sparse projections are sampled almost uniformly with sparsity ratio ranging from $[20\%, 80\%]$.

We extract $16$ random patches per training step. The loss function is the $\ell_1$ distance between predicted and ground-truth tokens, evaluated on unobserved target views:
\begin{equation}
    \mathcal{L} = \frac{1}{|\mathcal{Q}|} \sum_{q \in \mathcal{Q}} \bigl\| \hat{\tt}_q - \tt_q \bigr\|_1 ,
\end{equation}
where $\mathcal{Q}$ denotes the set of target angles and $\tt_q$ is the ground-truth back-projection token generated from the dense sinogram. Models are trained using AdamW ($\beta_1 = 0.9$, $\beta_2 = 0.95$, weight decay $0.05$) with a base learning rate of $3 \times 10^{-4}$ decayed to zero via a cosine schedule, and gradients clipped at a norm of $1.0$. Training runs for $10^6$ iterations on a single NVIDIA A100 GPU.

\begin{revblock}
\section{\texorpdfstring{\rev{Ray-consistent stitching: details}}{Ray-consistent stitching: details}}
\label{sec: sino_aggregation_appendix}
Section~\ref{sec: sino_aggregation} describes RCS as averaging the patch predictions along each ray. The rays are not aligned with the pixel grid, so a ray does not pass through pixel centers: each pixel of a token reads the filtered projection between two neighboring detector bins, by linear interpolation \eqref{eq:single_view_bp}. We therefore carry out the averaging as a least-squares fit in the sinogram domain. For each missing angle $\theta_q$, we find the single filtered projection $\mathbf{g}_q \in \R^{M}$ whose back-projection best matches the predictions of all patches:
\begin{equation}
    \hat{\mathbf{g}}_q = \argmin_{\mathbf{g} \in \R^{M}} \sum_{p} \bigl\| \mathbf{w} \odot \bigl( \mathcal{B}_{q,p}\, \mathbf{g} - \hat{\tt}_{q,p} \bigr) \bigr\|_2^2 ,
    \label{eq: sino_ls}
\end{equation}
where $\hat{\tt}_{q,p}$ is the prediction for patch $p$, $\mathcal{B}_{q,p}\,\mathbf{g}$ is the back-projection \eqref{eq:single_view_bp} of $\mathbf{g}$ at angle $\theta_q$ restricted to patch $p$, and the mask $\mathbf{w}$ keeps only the disc inscribed in the patch, discarding the corners, where predictions are least reliable. If every pixel read a single detector bin, the solution of \eqref{eq: sino_ls} would be exactly the mean of the predictions of all pixels on that bin, i.e.\ the plain average along the ray. Linear interpolation couples neighboring bins, so the normal equations are tridiagonal rather than diagonal, and they are solved exactly in $\mathcal{O}(M)$ operations per view. We then insert the fitted projections at the missing angles, keep the measured ones unchanged, and reconstruct the image by FBP. This requires no retraining and adds negligible cost.
\end{revblock}

\begin{revblock}
\section{\texorpdfstring{\rev{Self-supervised test-time fine-tuning: details}}{Self-supervised test-time fine-tuning: details}}
\label{sec: ssl_appendix}
Let's assume $f_\phi$ is a pretrained TomoTransformer and $\mathcal{M}$ be the $R_{\text{sparse}}$ measured projections of a single scan. At every fine-tuning step, we mask a subset $\mathcal{H} \subset \mathcal{M}$, feed the remaining projections $\mathcal{M} \setminus \mathcal{H}$ and query the masked angles in $\mathcal{H}$; we can then minimize the self-supervised loss between the predicted and masked measured views:
\begin{equation}
    \mathcal{L}_{\text{SSL}} = \frac{1}{|\mathcal{H}|} \sum_{q \in \mathcal{H}} \bigl\| \hat{\tt}_q - \tt_q \bigr\|_1 ,
    \qquad
    \hat{\tt}_q = f_{\phi}\bigl(\{(\tt_r, \theta_r)\}_{r \in \mathcal{M} \setminus \mathcal{H}};\, \theta_q\bigr),
    \label{eq: ssl_loss}
\end{equation}
where the targets $\tt_q$ are tokens of \emph{measured} views rather than of a simulated dense sinogram, so no ground truth image is required as opposed to the standard image-domain methods for CT reconstruction. The held-out fraction $|\mathcal{H}|/|\mathcal{M}|$ is drawn uniformly from $[20\%, 80\%]$, the sparsity range of pre-training, and the held-out views are spread evenly over $\mathcal{M}$ with a random offset. All weights are initialized from the pre-trained model and fine-tuned on random patches of the scan being reconstructed. At test time the fine-tuned model is used exactly like the pre-trained one.

\end{revblock}

\section{Baselines}
\label{sec: baseline_details}

\subsection{Image-domain post-processing baselines}
We compare against several established image-domain post-processing networks: U-Net~\citep{jin2017deep}, DRUNet~\citep{zhang2021plug}, NAFNet~\citep{chen2022simple}, Restormer~\citep{zamir2022restormer}, and CTformer~\citep{wang2023ctformer}. These models are trained to map a sparse-view FBP patch to its dense-view counterpart. They share the exact same training data, ramp filter, and back-projection operator as TomoTransformer, differing only in the neural architecture used for enhancement. At test time, all baselines are applied patch-wise using the identical tiling, stride, and overlap-averaging scheme as the patch-wise inference of Section~\ref{sec: sino_aggregation}, making them resolution-independent.

Each baseline retains its original published design, with channel widths and block depths adjusted slightly to match $\sim 12.71\,\text{M}$ parameters:
\begin{itemize}
    \item \textbf{U-Net}~\citep{ronneberger2015unet,jin2017deep}: A standard four-scale encoder--decoder using double $3\times3$ convolutions (BatchNorm, ReLU), max-pooling, and transposed-convolution upsampling. We adjust the base width to $40$ channels (channel progression: $40/80/160/320/640$), yielding $12.13\,\text{M}$ parameters.
    \item \textbf{DRUNet}~\citep{zhang2021plug}: The residual U-Net from DPIR featuring $4$ scales, $nb = 4$ residual blocks per scale, strided-convolution downsampling, and transposed-convolution upsampling. Scaling channel widths to $[40, 80, 160, 320]$ results in $12.76\,\text{M}$ parameters.
    \item \textbf{NAFNet}~\citep{chen2022simple}: A three-scale U-shaped architecture built with NAFBlocks (SimpleGate and simplified channel attention, without activation functions) distributed as $[2, 2, 2]$ encoder blocks, $2$ middle blocks, and $[2, 2, 2]$ decoder blocks. Using a base width of $80$ with a global residual connection yields $11.77\,\text{M}$ parameters.
    \item \textbf{Restormer}~\citep{zamir2022restormer}: A four-scale encoder--decoder using MDTA/GDFN transformer blocks ($[2, 2, 2, 2]$ blocks per scale, $2$ refinement blocks, and $[1, 2, 4, 8]$ heads). Setting the embedding dimension to $52$ and the FFN expansion factor to $2.66$ with a global residual connection gives $11.98\,\text{M}$ parameters.
    \item \textbf{CTformer}~\citep{wang2023ctformer}: Utilizes a token-to-token (T2T) dilated performer tokenizer/detokenizer with $2$ intermediate transformer blocks, $8$ heads, and an MLP ratio of $2.0$. To match capacity, we expand the original $1.6\,\text{M}$-parameter configuration (embedding dimension $64$, depth $1$) to an embedding dimension of $320$ and token dimension of $160$, yielding $10.72\,\text{M}$ parameters. Following the original design, the network predicts a residual subtracted from the input.
\end{itemize}

All baselines are trained on $P \times P$ patches ($P = 32$), matching TomoTransformer's token size, to ensure identical spatial context. The only exception is CTformer, whose public implementation hard-codes a $64 \times 64$ unfold/fold tokenizer grid. All baselines share TomoTransformer's optimization recipe described in Section~\ref{sec: training_details}: optimizer, learning rate, schedule, gradient clipping and iteration count. The objective necessarily differs: a post-processing network maps the sparse-view FBP patch to the dense-view FBP patch and is trained with an $\ell_1$ loss in the image domain, whereas TomoTransformer is supervised per projection on the unobserved views only.

\subsection{ViewTrans baseline}
\label{sec: viewtrans_details}
ViewTrans~\citep{chen2026viewtrans} is a sinogram-domain baseline that treats each projection view (a full row of the sinogram across all detector bins) as an individual token. We reimplement the original architecture using its view-as-token sequence representation, a cascade of $5$ pre-LayerNorm encoders starting with a geometry-aware self-attention (G-MSA) block, a pseudo-full-view input generated by linear view interpolation, a differentiable FBP head, and an image-domain $\ell_2$ loss. To adapt the baseline to our setup fairly, we make the following modifications:
\begin{itemize}
    \item \textbf{Parallel-beam geometry encoding.} The G-MSA block constructs queries and keys by concatenating layer-normalized sinogram features with three geometry maps. Because parallel-beam geometry lacks a fan angle, we replace the original $\tan(\gamma_i)$ detector encoding with the normalized detector coordinate $t_i \in [-1, 1]$ matching our projection operator. The angular map $\cos(\beta_j)$ and the binary observed/interpolated indicator remain unchanged.
    \item \textbf{Decoupling attention width from $M$.} ViewTrans sets its token dimension equal to the detector width $M$, tying the architecture to a single resolution. We benchmark at LoDoPaB-CT's native resolution ($M = 362$), which is not divisible by the attention head count. To resolve this, we decouple the inner attention dimension $H \cdot d$ from $M$ and project features back to $\mathbb{R}^{M}$ for the residual skip connection. Using the original head counts ($4$ for G-MSA, $8$ for standard blocks), a head dimension of $128$, and an MLP ratio of $4.0$, the model contains $13.06\,\text{M}$ parameters, closely matching TomoTransformer's $12.71\,\text{M}$.
    \item \textbf{Shared reconstruction operator.} We build the differentiable FBP head using the exact same ramp filter and back-projection operator as TomoTransformer, supervising predictions against dense-view FBPs generated by this operator. This ensures both models operate within an identical reconstruction domain.
    \item \textbf{Training schedule.} Because our training runs are longer than those in the original paper, we replace the StepLR schedule with a cosine decay to zero. We train using Adam at a learning rate of $5 \times 10^{-4}$, gradient clipping at $1.0$, and $8$ slices per gradient step for $10^{6}$ iterations, matching TomoTransformer's training length.
\end{itemize}
ViewTrans is trained on LoDoPaB-CT under the same dynamic sparse-view schedule as TomoTransformer, ensuring an identical experimental comparison.

\begin{revblock}
\subsection{CIN baseline}
\label{sec: sino_baseline_details}
The convolutional interpolation network (CIN) of \citet{guo2023efficient} restores the sparse sinogram before FBP with a fully convolutional network that predicts the $c-1$ missing views between each pair of measured neighbours ($c = R_{\text{total}}/R_{\text{sparse}} = 4$ here) as a residual on their linear interpolation, and passes the measured views through unchanged. We port a public TensorFlow implementation of the network to PyTorch: the sparse sinogram is extended by $6$ views at each end of the angle axis and zero-padded by $6$ bins on the detector axis, followed by three valid $5\times5$ convolutions with $32$ channels and $\tanh$ activations, and a $1\times1$ output convolution with $c-1$ channels, trained with an $\ell_2$ loss on the predicted views. This network has $52$\,k parameters and a receptive field of $\pm6$ views. To compare at the capacity of the other baselines, we keep everything around the network (input, interpolation residual, interleaving, loss) and replace the four-layer stack by the standard way of scaling such super-resolution networks, an EDSR-style residual body~\citep{lim2017enhanced}: a $3\times3$ head, $18$ residual blocks (conv--ReLU--conv, width $192$, residual scaling $0.1$) and a zero-initialized $3\times3$ tail, so that training starts from linear interpolation. The angle axis is extended by the receptive-field radius ($\pm38$ views) before the convolutions and cropped back afterwards. This model, CIN in Tables~\ref{tab: fixed_sparsity_psnr} and~\ref{tab: fixed_sparsity_ssim}, has $11.96\,\text{M}$ parameters.

Two adaptations are needed for our setting. First, the original wraps the angle axis circularly, which is correct for their $360^\circ$ fan beam; for parallel beam over $[0^\circ, 180^\circ)$ the view after the last one is the first view seen from the opposite side, so we wrap with the detector axis reversed. Second, sinograms of different detector widths $M$ are brought to a common scale by working with raw line integrals, so that, being fully convolutional, a single model is trained on all eight datasets at their native resolution, exactly like the other baselines. Training follows the fixed $64\!\rightarrow\!256$ protocol of Section~\ref{sec: fixed_sparsity} with the same data, projector and test volumes: Adam with cosine decay ($10^{-4}$, $4$ slices of $32$ per step), $2\times10^{4}$ epochs. The restored sinogram is reconstructed with the same FBP operator and scored against the same dense $256$-view reference. Since the upsampling factor is fixed by the number of output channels, the model is not evaluated at $128\!\rightarrow\!256$ (Table~\ref{tab: fixed_sparsity_128}).
\end{revblock}

\begin{revblock}
\section{\texorpdfstring{\rev{Additional experiments}}{Additional experiments}}
\label{sec: additional_experiments}

\subsection{Computational cost}
\label{sec: compute_cost}
Table~\ref{tab: compute_cost} reports the inference cost of the models of Table~\ref{tab: fixed_sparsity_psnr} in the setting of that table: one LoDoPaB-CT slice ($362\times362$, $M = 362$) with $R_{\text{sparse}} = 64$ of $R_{\text{total}} = 256$ views. Table~\ref{tab: compute_cost} shows that CIN requires the most compute, about twice as much as TomoTransformer. However, TomoTransformer is the slowest model ($67$\,ms per slice against $10$--$58$\,ms) and needs about as much memory as Restormer ($1.5$\,GiB against $1.4$\,GiB). Most of this cost is attention: each patch carries one token per view of the dense grid, so the decoder attends over $R_{\text{total}} = 256$ tokens for every patch. Nevertheless, TomoTransformer remains scalable because it processes small patches: the network only ever sees $P\times P$ patches, so its peak memory is set by the number of patches processed in one forward pass, not by the image size, and its compute grows linearly with the number of pixels. The image-domain baselines are applied patch-wise in the same way (Appendix~\ref{sec: baseline_details}), whereas sinogram-domain models, in particular those that include the Radon transform and its adjoint in the network, can become computationally expensive for large images.

\begin{table}[h]\normalcolor %
\caption{\rev{Inference cost per $362\times362$ slice at $64\!\rightarrow\!256$ on one NVIDIA A100 (40\,GB), PyTorch~2.8, median of $20$ runs. GFLOPs is computed only for the network only. The peak memory of the allocated GPU is reported during one slice reconstruction.}}
\centering
\renewcommand\arraystretch{1.2}
\begin{tabular}{l|c|ccc}
\hline
Method & Params (M) & GFLOPs & Time (ms) & Memory (GiB) \\
\hline
FBP~\citep{kak1988principles} & -- & -- & 1.2 & 0.11 \\
U-Net~\citep{ronneberger2015unet} & 12.13 & 85 & 9.7 & 0.27 \\
DRUNet~\citep{zhang2021plug} & 12.76 & 244 & 19.8 & 0.28 \\
NAFNet~\citep{chen2022simple} & 11.77 & 185 & 34.2 & 0.47 \\
Restormer~\citep{zamir2022restormer} & 11.98 & 374 & 57.6 & 1.44 \\
CTformer~\citep{wang2023ctformer} & 10.72 & 669 & 57.2 & 1.16 \\
CIN~\citep{guo2023efficient} & 11.96 & 1211 & 45.8 & 0.49 \\
ViewTrans~\citep{chen2026viewtrans} & 13.06 & 7.9 & 5.6 & 0.50 \\
TomoTransformer & 12.71 & 666 & 67.1 & 1.51 \\
\hline
\end{tabular}
\label{tab: compute_cost}
\end{table}

\subsection{Effect of the patch size}
\label{sec: patch_size}
In this section we study the effect of the patch size $P$ on reconstruction quality. Our experiments in this paper use $P = 32$ throughout. Here we also evaluate $P \in \{16, 64, 128\}$ to understand the trade-offs, both for TomoTransformer and for the image-domain baseline DRUNet (Section~\ref{sec: fixed_sparsity}). We chose DRUNet because it is one of the best-performing image-domain baseline in Table~\ref{tab: fixed_sparsity_psnr} and it is computationally efficient in particular for large patches which  makes it a good candidate to run this patch-based evaluation. Both models are trained under the fixed protocol (Appendix~\ref{sec: training_details}), for $R_{\text{sparse}} = 128$ of $R_{\text{total}} = 256$, with $P \in \{16, 32, 64, 128\}$ and nothing else changed. For TomoTransformer, the tokenizer samples $K = \lceil P\sqrt{2}\,\rceil = 23$, $46$, $91$ and $182$ points per patch, so only the linear maps $\mathbf{E}$ and $\mathbf{E}'$ change size and all four models have $12.7$--$12.9$\,M parameters. DRUNet is fully convolutional, so its number of parameters does not depend on $P$. To keep the cost of the eight trainings manageable, each runs for $2\times10^5$ iterations instead of $10^6$.

Table~\ref{tab: patch_size} shows that the two models respond to the patch size in opposite ways. TomoTransformer is best with small patches, and larger patches are worse on all eight datasets. This is expected from the angular sampling of the sinogram: the number of views needed to sample an object without aliasing grows with its size. A larger patch may need more projections for the same reconstruction quality, and with the same $128$ measured views its missing views are harder to predict.

DRUNet, in contrast, improves steadily with the patch size. since it removes the streak artifacts of the sparse-view FBP in the image domain, a larger patch provides more spatial context which helps improve performance. DRUNet with patch size $P = 128$ shows comparable performance to TomoTransformer with $P = 16$ in average PSNR and SSIM; spatial context lead to better performance on HNSCC, ImageNet and LoDoPaB but TomoTransformer remains better on the other five datasets.

\begin{table*}[t]\normalcolor %
\caption{\rev{Effect of the patch size $P$ (Appendix~\ref{sec: patch_size}): TomoTransformer and DRUNet trained under the fixed protocol at $128\!\rightarrow\!256$ with $P\times P$ patches. \textbf{PSNR} (dB, top) and \textbf{SSIM} (bottom) per dataset, mean over test slices. $P = 32$ is the setting used in the paper. Best per column in \textbf{bold}.}}
\centering
\renewcommand\arraystretch{1.25}
\resizebox{\textwidth}{!}{%
\begin{tabular}{c|c|cccccccc|c}
\hline
Model & Patch size & LoDoPaB & COVID-19 & KiTS23 & MSD-T10 & COLONOG & HNSCC & FFHQ & ImageNet & Average \\
\hline
\multicolumn{11}{c}{PSNR (dB)} \\
\hline
\multirow{4}{*}{\shortstack{TomoTransformer\\(fixed, RCS)}} & $P = 16$ & 42.83 & \textbf{44.03} & \textbf{48.33} & 45.50 & \textbf{42.12} & 48.87 & 42.10 & 39.79 & 44.20 \\
 & $P = 32$ & 42.83 & 43.99 & 48.31 & \textbf{45.51} & 42.11 & 48.72 & \textbf{42.21} & 39.70 & 44.17 \\
 & $P = 64$ & 42.55 & 43.61 & 47.96 & 45.28 & 41.93 & 48.16 & 41.87 & 39.33 & 43.84 \\
 & $P = 128$ & 41.92 & 42.68 & 47.21 & 44.70 & 41.47 & 46.18 & 40.89 & 38.36 & 42.93 \\
\hline
\multirow{4}{*}{DRUNet} & $P = 16$ & 41.49 & 41.63 & 45.94 & 43.20 & 40.68 & 45.27 & 39.22 & 38.35 & 41.97 \\
 & $P = 32$ & 42.20 & 42.61 & 46.94 & 44.09 & 41.30 & 47.49 & 40.30 & 39.21 & 43.02 \\
 & $P = 64$ & 42.73 & 43.24 & 47.53 & 44.61 & 41.69 & 48.85 & 41.11 & 39.92 & 43.71 \\
 & $P = 128$ & \textbf{43.26} & 43.89 & 48.10 & 45.02 & 42.03 & \textbf{50.01} & 41.84 & \textbf{40.57} & \textbf{44.34} \\
\hline
\multicolumn{11}{c}{SSIM} \\
\hline
\multirow{4}{*}{\shortstack{TomoTransformer\\(fixed, RCS)}} & $P = 16$ & 0.960 & \textbf{0.973} & \textbf{0.987} & \textbf{0.976} & \textbf{0.954} & \textbf{0.988} & \textbf{0.967} & 0.957 & \textbf{0.970} \\
 & $P = 32$ & 0.960 & \textbf{0.973} & \textbf{0.987} & \textbf{0.976} & \textbf{0.954} & 0.987 & \textbf{0.967} & 0.956 & \textbf{0.970} \\
 & $P = 64$ & 0.959 & 0.971 & 0.986 & 0.975 & 0.952 & 0.983 & 0.965 & 0.953 & 0.968 \\
 & $P = 128$ & 0.955 & 0.966 & 0.984 & 0.972 & 0.949 & 0.973 & 0.959 & 0.946 & 0.963 \\
\hline
\multirow{4}{*}{DRUNet} & $P = 16$ & 0.948 & 0.958 & 0.978 & 0.961 & 0.939 & 0.956 & 0.938 & 0.942 & 0.953 \\
 & $P = 32$ & 0.954 & 0.964 & 0.982 & 0.967 & 0.945 & 0.972 & 0.950 & 0.951 & 0.960 \\
 & $P = 64$ & 0.958 & 0.967 & 0.984 & 0.970 & 0.948 & 0.979 & 0.957 & 0.956 & 0.965 \\
 & $P = 128$ & \textbf{0.961} & 0.970 & 0.986 & 0.972 & 0.951 & 0.984 & 0.963 & \textbf{0.960} & 0.968 \\
\hline
\end{tabular}}
\label{tab: patch_size}
\end{table*}

\subsection{Effect of the geometric tokenizer}
\label{sec: tokenizer_ablation}
To isolate the contribution of the geometric tokenizer and decoder (Section~\ref{sec: Geometric tokenizer}), we replace them by plain linear maps of the full patch: the encoder $\mathbf{E}: \R^{P^2} \to \R^{D}$ embeds the vectorized $P \times P$ back-projection patch $\tt_r$, and the decoder $\mathbf{E}': \R^{D} \to \R^{P^2}$ predicts the output patch $\hat{\tt}_q$ directly. Everything else is unchanged. Both models are trained under the varied protocol for $2\times10^5$ iterations instead of $10^6$, and are evaluated at $128\!\rightarrow\!256$ on the test volumes of Table~\ref{tab: fixed_sparsity_psnr}, against the same dense $256$-view reference.

Table~\ref{tab: tokenizer_ablation} shows that the geometric tokenizer is better on all eight datasets in both PSNR and SSIM, despite having $1$\,M fewer parameters. The gap is the same at $64\!\rightarrow\!256$, so it does not depend on the sparsity level. A single-view back-projection patch is fully determined by its $1$-D profile across the stripes. The geometric tokenizer builds this structure into the model, so every decoded patch is a valid single-view back-projection. The full-patch model has to learn this structure from data, and its decoder can output patches that no single view could produce.

\begin{table*}[t]\normalcolor %
\caption{\rev{Effect of the geometric tokenizer (Appendix~\ref{sec: tokenizer_ablation}): TomoTransformer with the geometric tokenizer and decoder of Section~\ref{sec: Geometric tokenizer} vs.\ linear maps of the full $32\times32$ patch. \textbf{PSNR} (dB, top) and \textbf{SSIM} (bottom) per dataset, mean over test slices. Best per column in \textbf{bold}.}}
\centering
\renewcommand\arraystretch{1.25}
\resizebox{\textwidth}{!}{%
\begin{tabular}{c|c|cccccccc|c}
\hline
Tokenizer & Params & LoDoPaB & COVID-19 & KiTS23 & MSD-T10 & COLONOG & HNSCC & FFHQ & ImageNet & Average \\
\hline
\multicolumn{11}{c}{PSNR (dB)} \\
\hline
Full patch & 13.71\,M & 41.45 & 41.88 & 46.60 & 44.31 & 40.99 & 42.96 & 40.28 & 37.84 & 42.04 \\
Geometric & 12.71\,M & \textbf{42.11} & \textbf{43.05} & \textbf{47.43} & \textbf{44.83} & \textbf{41.58} & \textbf{43.27} & \textbf{41.32} & \textbf{39.00} & \textbf{42.82} \\
\hline
\multicolumn{11}{c}{SSIM} \\
\hline
Full patch & 13.71\,M & 0.954 & 0.964 & 0.983 & 0.970 & 0.947 & 0.964 & 0.955 & 0.945 & 0.960 \\
Geometric & 12.71\,M & \textbf{0.958} & \textbf{0.970} & \textbf{0.985} & \textbf{0.973} & \textbf{0.951} & \textbf{0.972} & \textbf{0.962} & \textbf{0.953} & \textbf{0.965} \\
\hline
\end{tabular}}
\label{tab: tokenizer_ablation}
\end{table*}

\subsection{Effect of the transformer backbone}
\label{sec: backbone_ablation}
TomoTransformer uses an encoder--decoder backbone in the style of MAE \citep{he2022masked} (Section~\ref{sec: tomotransformers} and Appendix~\ref{sec: tomo_arch_details}). Now we replace this encoder--decoder backbone  with a BERT-style backbone \citep{devlin2019bert}: a single stack of $8$ transformer blocks, as many as the encoder and decoder together, that processes the full sequence of $R_{\text{total}}$ tokens. Everything else is unchanged. Both models have $12.71$\,M parameters.

Table~\ref{tab: backbone_ablation} shows that the two backbones perform almost identically. TomoTransformer's performance does not depend on the encoder--decoder split, unlike the choice of tokenizer. We keep the MAE-style backbone because it is cheaper: its encoder attends over the $R_{\text{sparse}}$ measured tokens only, while every block of the BERT-style model attends over all $R_{\text{total}}$ tokens.

\begin{table*}[t]\normalcolor %
\caption{\rev{Effect of the transformer backbone (Appendix~\ref{sec: backbone_ablation}): TomoTransformer with the MAE-style encoder--decoder of Section~\ref{sec: tomotransformers} vs.\ a BERT-style stack of the same $8$ blocks over all tokens. \textbf{PSNR} (dB, top) and \textbf{SSIM} (bottom) per dataset, mean over test slices. The MAE-style model is the geometric model of Table~\ref{tab: tokenizer_ablation}. Best per column in \textbf{bold}.}}
\centering
\renewcommand\arraystretch{1.25}
\resizebox{\textwidth}{!}{%
\begin{tabular}{c|c|cccccccc|c}
\hline
Backbone & Params & LoDoPaB & COVID-19 & KiTS23 & MSD-T10 & COLONOG & HNSCC & FFHQ & ImageNet & Average \\
\hline
\multicolumn{11}{c}{PSNR (dB)} \\
\hline
BERT-style & 12.71\,M & \textbf{42.15} & \textbf{43.15} & \textbf{47.54} & \textbf{44.90} & \textbf{41.64} & \textbf{43.46} & \textbf{41.39} & \textbf{39.11} & \textbf{42.92} \\
MAE-style & 12.71\,M & 42.11 & 43.05 & 47.43 & 44.83 & 41.58 & 43.27 & 41.32 & 39.00 & 42.82 \\
\hline
\multicolumn{11}{c}{SSIM} \\
\hline
BERT-style & 12.71\,M & \textbf{0.958} & \textbf{0.970} & \textbf{0.985} & \textbf{0.974} & \textbf{0.951} & \textbf{0.974} & \textbf{0.963} & \textbf{0.954} & \textbf{0.966} \\
MAE-style & 12.71\,M & \textbf{0.958} & \textbf{0.970} & \textbf{0.985} & 0.973 & \textbf{0.951} & 0.972 & 0.962 & 0.953 & 0.965 \\
\hline
\end{tabular}}
\label{tab: backbone_ablation}
\end{table*}

\subsection{Training on a single dataset}
\label{sec: single_dataset}
All models of Table~\ref{tab: fixed_sparsity_psnr} are trained on the mixture of eight datasets. To see whether the comparison changes when training data is limited to a single source, we train Restormer, the strongest image-domain baseline of Table~\ref{tab: fixed_sparsity_psnr}, and TomoTransformer (fixed) on a single dataset, either LoDoPaB-CT or KiTS23. The protocol ($64\!\rightarrow\!256$), architectures and all other settings are those of Table~\ref{tab: fixed_sparsity_psnr}, except that both models are trained for $2\times10^5$ iterations instead of $10^6$. Both are evaluated on the test volumes of all eight datasets, against the same dense $256$-view reference, so the training dataset is in-domain and the other seven datasets are not seen during training. Table~\ref{tab: single_dataset} reports the results.

Trained on LoDoPaB-CT, Restormer and TomoTransformer show comparable performance on in-distribution data. On the seven out-of-distribution datasets, TomoTransformer is remarkably better on every dataset and its average over all eight datasets is $1.24$\,dB higher ($38.85$ against $37.61$\,dB). We see a similar pattern when training on KiTS23. This experiment shows that, when trained on a single dataset, TomoTransformer matches the strongest image-domain baseline in-domain and generalizes better to out-of-distribution data.

\begin{table*}[t]\normalcolor %
\caption{\rev{Training on a single dataset (Appendix~\ref{sec: single_dataset}), fixed sparse-view reconstruction ($64\!\rightarrow\!256$): (a) \textbf{PSNR} (dB) and (b) \textbf{SSIM} per dataset, mean over the $32$ test slices of each dataset, against the dense $256$-view FBP. Best per column within each group in \textbf{bold}.}}
\centering
\renewcommand\arraystretch{1.25}
\begin{subtable}{\textwidth}
\centering
\caption{PSNR (dB)}
\resizebox{\textwidth}{!}{%
\begin{tabular}{c|cccccccc|c}
\hline
Method & LoDoPaB & COVID-19 & KiTS23 & MSD-T10 & COLONOG & HNSCC & FFHQ & ImageNet & Average \\
\hline
FBP~\citep{kak1988principles} & 29.24 & 28.48 & 30.27 & 28.13 & 28.59 & 28.93 & 27.97 & 26.46 & 28.51 \\
\hline
\multicolumn{10}{l}{\textit{Trained on LoDoPaB-CT}} \\
Restormer~\citep{zamir2022restormer} & \textbf{38.48} & 37.48 & 41.73 & 38.79 & 37.51 & 40.38 & 33.51 & 33.04 & 37.61 \\
TomoTransformer (fixed, RCS) & 38.39 & \textbf{39.38} & \textbf{43.61} & \textbf{40.26} & \textbf{38.68} & \textbf{41.02} & \textbf{35.95} & \textbf{33.51} & \textbf{38.85} \\
\hline
\multicolumn{10}{l}{\textit{Trained on KiTS23}} \\
Restormer~\citep{zamir2022restormer} & \textbf{37.79} & 39.15 & 43.60 & 40.51 & 38.50 & 40.97 & 34.15 & 32.60 & 38.41 \\
TomoTransformer (fixed, RCS) & 37.75 & \textbf{39.68} & \textbf{43.73} & \textbf{40.75} & \textbf{38.66} & \textbf{41.45} & \textbf{36.04} & \textbf{33.16} & \textbf{38.90} \\
\hline
\end{tabular}}
\label{tab: single_dataset_psnr}
\end{subtable}

\vspace{0.8em}
\begin{subtable}{\textwidth}
\centering
\caption{SSIM}
\resizebox{\textwidth}{!}{%
\begin{tabular}{c|cccccccc|c}
\hline
Method & LoDoPaB & COVID-19 & KiTS23 & MSD-T10 & COLONOG & HNSCC & FFHQ & ImageNet & Average \\
\hline
FBP~\citep{kak1988principles} & 0.635 & 0.654 & 0.662 & 0.554 & 0.610 & 0.395 & 0.631 & 0.598 & 0.592 \\
\hline
\multicolumn{10}{l}{\textit{Trained on LoDoPaB-CT}} \\
Restormer~\citep{zamir2022restormer} & \textbf{0.918} & 0.926 & 0.952 & 0.920 & 0.897 & 0.903 & 0.859 & 0.859 & 0.904 \\
TomoTransformer (fixed, RCS) & \textbf{0.918} & \textbf{0.945} & \textbf{0.971} & \textbf{0.946} & \textbf{0.915} & \textbf{0.946} & \textbf{0.909} & \textbf{0.873} & \textbf{0.928} \\
\hline
\multicolumn{10}{l}{\textit{Trained on KiTS23}} \\
Restormer~\citep{zamir2022restormer} & 0.907 & 0.938 & 0.969 & 0.942 & 0.908 & 0.921 & 0.873 & 0.850 & 0.914 \\
TomoTransformer (fixed, RCS) & \textbf{0.911} & \textbf{0.945} & \textbf{0.972} & \textbf{0.948} & \textbf{0.914} & \textbf{0.945} & \textbf{0.910} & \textbf{0.866} & \textbf{0.926} \\
\hline
\end{tabular}}
\label{tab: single_dataset_ssim}
\end{subtable}
\label{tab: single_dataset}
\end{table*}

\subsection{Statistical significance of the fixed sparse-view comparison}
\label{sec: significance}
The values of Table~\ref{tab: fixed_sparsity_psnr} are means over the $32$ test slices of each dataset. Here we provide some statistical analysis of the differences between TomoTransformer and the strongest baseline, Restormer, on the same test slices. For every test slice $z$ we compute $\Delta_z = \text{PSNR}_{\text{ours}}(z) - \text{PSNR}_{\text{Restormer}}(z)$. Table~\ref{tab: significance} reports the mean of $\Delta_z$, its $95\%$ confidence interval, and the number of slices on which TomoTransformer is better. Slices from the same CT volume are correlated, so the interval is obtained by a bootstrap over test volumes: in each of $10^4$ resamples, the test volumes of a dataset are drawn with replacement together with all their slices, and the interval is given by the $2.5$th and $97.5$th percentiles of the resampled means. For LoDoPaB-CT, FFHQ and ImageNet, whose test sets are collections of single slices, the slices are resampled individually. For the average, each dataset is resampled separately and the eight dataset means are averaged. 

TomoTransformer (fixed, RCS) improves on Restormer by $0.60$\,dB on average ($95\%$ CI $[0.52, 0.68]$) and is better on $217$ of the $256$ test slices. The improvement is significant on seven of the eight datasets. On HNSCC, whose test set contains only six volumes, the interval includes zero ($+0.24$\,dB, $[-0.15, 0.66]$), so the two models are on par there. Without ray-consistent stitching, TomoTransformer (fixed) is still significantly better on average ($+0.19$\,dB, $[0.13, 0.25]$), but significantly worse on LoDoPaB-CT and HNSCC. TomoTransformer (varied), which is not trained for this protocol, is below Restormer on every dataset except FFHQ, and most of its average gap comes from HNSCC (see Appendix~\ref{sec: hnscc}).

\begin{table*}[t]\normalcolor %
\caption{\rev{Paired comparison with the strongest baseline in Table~\ref{tab: fixed_sparsity_psnr}, fixed sparse-view reconstruction ($64\!\rightarrow\!256$): PSNR difference $\Delta$ (dB) between each TomoTransformer variant and Restormer on the same test slices. Mean $\Delta$ with its $95\%$ confidence interval from a bootstrap over test volumes, and the number of test slices on which TomoTransformer is better (wins). Positive values favor TomoTransformer.}}
\centering
\renewcommand\arraystretch{1.25}
\resizebox{\textwidth}{!}{%
\begin{tabular}{l|rc|rc|rc|rc}
\hline
& \multicolumn{2}{c|}{TomoTransformer (fixed)} & \multicolumn{2}{c|}{TomoTransformer (fixed, RCS)} & \multicolumn{2}{c|}{TomoTransformer (varied)} & \multicolumn{2}{c}{TomoTransformer (varied, RCS)} \\
Dataset & $\Delta$ [$95\%$ CI] & wins & $\Delta$ [$95\%$ CI] & wins & $\Delta$ [$95\%$ CI] & wins & $\Delta$ [$95\%$ CI] & wins \\
\hline
LoDoPaB & $-0.13$ $[-0.25, -0.01]$ & 10/32 & $+0.14$ $[0.02, 0.27]$ & 21/32 & $-0.85$ $[-1.04, -0.68]$ & 1/32 & $-0.59$ $[-0.79, -0.42]$ & 2/32 \\
COVID-19 & $+0.27$ $[0.16, 0.35]$ & 29/32 & $+0.69$ $[0.54, 0.81]$ & 32/32 & $-0.91$ $[-1.10, -0.74]$ & 0/32 & $-0.48$ $[-0.67, -0.31]$ & 2/32 \\
KiTS23 & $+0.36$ $[0.14, 0.56]$ & 24/32 & $+0.77$ $[0.53, 1.00]$ & 32/32 & $-0.91$ $[-1.11, -0.74]$ & 0/32 & $-0.40$ $[-0.60, -0.23]$ & 3/32 \\
MSD-T10 & $+0.42$ $[0.26, 0.60]$ & 28/32 & $+0.77$ $[0.55, 1.03]$ & 28/32 & $-0.71$ $[-0.82, -0.59]$ & 1/32 & $-0.30$ $[-0.46, -0.14]$ & 11/32 \\
COLONOG & $+0.11$ $[-0.02, 0.25]$ & 19/32 & $+0.34$ $[0.22, 0.48]$ & 29/32 & $-0.56$ $[-0.73, -0.40]$ & 2/32 & $-0.29$ $[-0.44, -0.13]$ & 8/32 \\
HNSCC & $-0.65$ $[-0.87, -0.37]$ & 5/32 & $+0.24$ $[-0.15, 0.66]$ & 17/32 & $-5.12$ $[-5.39, -4.92]$ & 0/32 & $-4.87$ $[-5.23, -4.53]$ & 0/32 \\
FFHQ & $+1.16$ $[1.00, 1.32]$ & 32/32 & $+1.56$ $[1.39, 1.74]$ & 32/32 & $+0.33$ $[0.18, 0.49]$ & 25/32 & $+0.68$ $[0.51, 0.85]$ & 29/32 \\
ImageNet & $+0.01$ $[-0.18, 0.17]$ & 18/32 & $+0.31$ $[0.12, 0.47]$ & 26/32 & $-0.47$ $[-0.69, -0.27]$ & 7/32 & $-0.18$ $[-0.39, 0.00]$ & 10/32 \\
\hline
Average & $+0.19$ $[0.13, 0.25]$ & 165/256 & $+0.60$ $[0.52, 0.68]$ & 217/256 & $-1.15$ $[-1.22, -1.09]$ & 36/256 & $-0.80$ $[-0.88, -0.73]$ & 65/256 \\
\hline
\end{tabular}}
\label{tab: significance}
\end{table*}

\subsection{\texorpdfstring{\rev{Zero-shot denoising under stronger noise}}{Zero-shot denoising under stronger noise}}
\label{sec: denoising_strong}
\rev{Figure~\ref{fig: noise_denoising} considers Gaussian noise at $45$\,dB and Poisson noise with $I_0 = 10^5$ photons per bin. Here we repeat the experiment of Section~\ref{sec: image_denoising} under stronger noise, with the same model (TomoTransformer (varied). We lower the SNR of the Gaussian noise to $40$ and $35$\,dB and the dose of the Poisson noise to $I_0 = 3\times10^4$ and $10^4$, which correspond to an SNR of $41$ and $36$\,dB on the projections. Table~\ref{tab: denoising_strong} reports the results and Figure~\ref{fig: noise_denoising_strong} shows the reconstructions at the strongest noise level of each model.}

\rev{Table~\ref{tab: denoising_strong} shows that TomoTransformer trained on noiseless projections still generalizes to these stronger noise levels, in particular, the improvement over FBP grows with the noise levels, from $8.0$ to $13.1$\,dB for Gaussian noise and from $7.5$ to $13.1$\,dB for Poisson noise. Keeping the noisy measured views becomes increasingly harmful as expected. Both re-masking schemes improve on the predicted views alone at every noise level, by up to $2.2$\,dB. Complementary re-masking is the better scheme at the mild noise levels of Figure~\ref{fig: noise_denoising}, and blind-spot re-masking at stronger levels.}

\begin{table}[t]\normalcolor %
\caption{\rev{Zero-shot denoising under increasing noise (Appendix~\ref{sec: denoising_strong}): TomoTransformer (varied), trained on noiseless projections, applied to $128$ noisy views of LoDoPaB-CT to predict $256$. \textbf{PSNR} (dB) / \textbf{SSIM} against the noise-free $256$-view FBP, on the test slices of Figure~\ref{fig: noise_denoising}. Columns as in Figure~\ref{fig: noise_denoising}. Best per row in \textbf{bold}. $^\dagger$Noise levels of Figure~\ref{fig: noise_denoising}; $^\ddagger$noise levels of Figure~\ref{fig: noise_denoising_strong}.}}
\centering
\renewcommand\arraystretch{1.25}
\resizebox{\textwidth}{!}{%
\begin{tabular}{lc|ccccc}
\hline
Noise & SNR (dB) & FBP input & Predicted only & Predicted $+$ measured & BS & CR \\
\hline
Gaussian$^\dagger$ & $45$ & 28.96 / 0.559 & 35.41 / 0.836 & 34.07 / 0.784 & 36.56 / 0.881 & \textbf{36.93} / \textbf{0.887} \\
Gaussian & $40$ & 24.45 / 0.339 & 33.62 / 0.786 & 29.66 / 0.589 & \textbf{35.08} / \textbf{0.852} & 34.85 / 0.840 \\
Gaussian$^\ddagger$ & $35$ & 19.62 / 0.164 & 30.56 / 0.668 & 24.88 / 0.354 & \textbf{32.71} / \textbf{0.786} & 31.83 / 0.748 \\
\hline
Poisson$^\dagger$, $I_0 = 10^5$ & $46$ & 30.03 / 0.620 & 35.83 / 0.848 & 35.23 / 0.826 & 37.01 / 0.891 & \textbf{37.54} / \textbf{0.900} \\
Poisson, $I_0 = 3\times10^4$ & $41$ & 25.44 / 0.399 & 34.27 / 0.806 & 30.73 / 0.649 & \textbf{35.81} / \textbf{0.869} & 35.71 / 0.864 \\
Poisson$^\ddagger$, $I_0 = 10^4$ & $36$ & 20.87 / 0.217 & 32.08 / 0.739 & 26.25 / 0.434 & \textbf{33.95} / \textbf{0.829} & 33.33 / 0.805 \\
\hline
\end{tabular}}
\label{tab: denoising_strong}
\end{table}

\begin{figure*}[tb]\normalcolor %
  \centering
  \newcommand{\cw}{0.155\textwidth}
  \begin{minipage}[c]{0.03\textwidth}\centering\rotatebox{90}{\scriptsize\textbf{Gaussian ($35$\,dB})}\end{minipage}\hfill
  \begin{subfigure}[c]{\cw}\centering%
  \begin{tikzpicture}[spy using outlines={circle,orange,line width=0.5pt,magnification=3,size=0.7327cm, connect spies}]
    \node[name=img, inner sep=0pt, line width=0.05mm, draw=white] at (0,0) {\includegraphics[width=\textwidth]{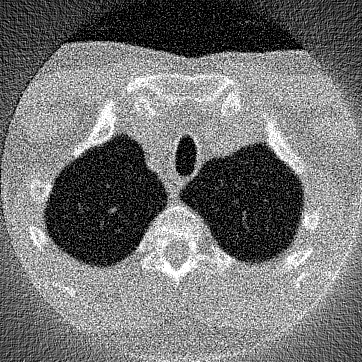}};
    \node[anchor=south west, draw=yellow, line width=0.35pt, fill=black, fill opacity=0.55, text=yellow, text opacity=1, inner sep=1.2pt, font=\tiny] at (img.south west) {19.62\,dB};
    \spy on ($(img.center)+1.181818*(0,0.6)$) in node at ($(img.north west)+(0.39,-0.39)$);
  \end{tikzpicture}\end{subfigure}\hfill
  \begin{subfigure}[c]{\cw}\centering%
  \begin{tikzpicture}[spy using outlines={circle,orange,line width=0.5pt,magnification=3,size=0.7327cm, connect spies}]
    \node[name=img, inner sep=0pt, line width=0.05mm, draw=white] at (0,0) {\includegraphics[width=\textwidth]{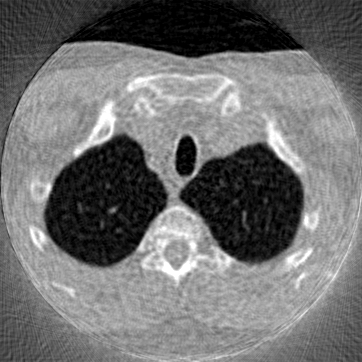}};
    \node[anchor=south west, draw=yellow, line width=0.35pt, fill=black, fill opacity=0.55, text=yellow, text opacity=1, inner sep=1.2pt, font=\tiny] at (img.south west) {30.56\,dB};
    \spy on ($(img.center)+1.181818*(0,0.6)$) in node at ($(img.north west)+(0.39,-0.39)$);
  \end{tikzpicture}\end{subfigure}\hfill
  \begin{subfigure}[c]{\cw}\centering%
  \begin{tikzpicture}[spy using outlines={circle,orange,line width=0.5pt,magnification=3,size=0.7327cm, connect spies}]
    \node[name=img, inner sep=0pt, line width=0.05mm, draw=white] at (0,0) {\includegraphics[width=\textwidth]{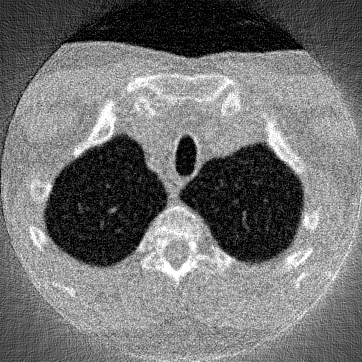}};
    \node[anchor=south west, draw=yellow, line width=0.35pt, fill=black, fill opacity=0.55, text=yellow, text opacity=1, inner sep=1.2pt, font=\tiny] at (img.south west) {24.88\,dB};
    \spy on ($(img.center)+1.181818*(0,0.6)$) in node at ($(img.north west)+(0.39,-0.39)$);
  \end{tikzpicture}\end{subfigure}\hfill
  \begin{subfigure}[c]{\cw}\centering%
  \begin{tikzpicture}[spy using outlines={circle,orange,line width=0.5pt,magnification=3,size=0.7327cm, connect spies}]
    \node[name=img, inner sep=0pt, line width=0.05mm, draw=white] at (0,0) {\includegraphics[width=\textwidth]{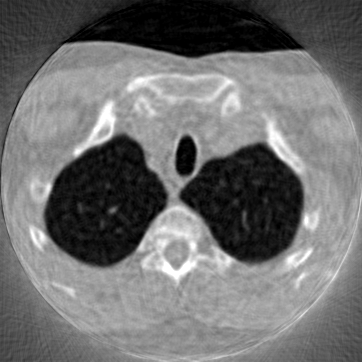}};
    \node[anchor=south west, draw=yellow, line width=0.35pt, fill=black, fill opacity=0.55, text=yellow, text opacity=1, inner sep=1.2pt, font=\tiny] at (img.south west) {\textbf{32.71\,dB}};
    \spy on ($(img.center)+1.181818*(0,0.6)$) in node at ($(img.north west)+(0.39,-0.39)$);
  \end{tikzpicture}\end{subfigure}\hfill
  \begin{subfigure}[c]{\cw}\centering%
  \begin{tikzpicture}[spy using outlines={circle,orange,line width=0.5pt,magnification=3,size=0.7327cm, connect spies}]
    \node[name=img, inner sep=0pt, line width=0.05mm, draw=white] at (0,0) {\includegraphics[width=\textwidth]{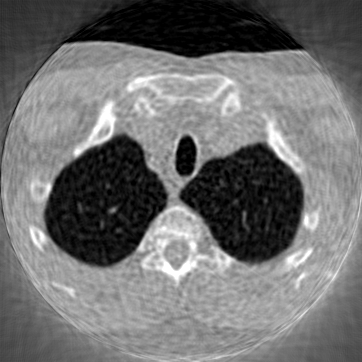}};
    \node[anchor=south west, draw=yellow, line width=0.35pt, fill=black, fill opacity=0.55, text=yellow, text opacity=1, inner sep=1.2pt, font=\tiny] at (img.south west) {31.83\,dB};
    \spy on ($(img.center)+1.181818*(0,0.6)$) in node at ($(img.north west)+(0.39,-0.39)$);
  \end{tikzpicture}\end{subfigure}\hfill
  \begin{subfigure}[c]{\cw}\centering%
  \begin{tikzpicture}[spy using outlines={circle,orange,line width=0.5pt,magnification=3,size=0.7327cm, connect spies}]
    \node[name=img, inner sep=0pt, line width=0.05mm, draw=white] at (0,0) {\includegraphics[width=\textwidth]{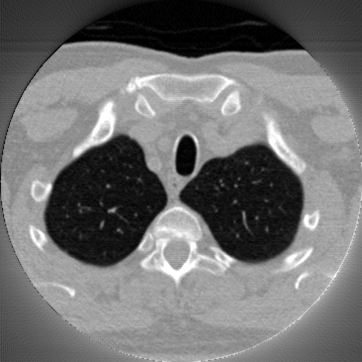}};
    \spy on ($(img.center)+1.181818*(0,0.6)$) in node at ($(img.north west)+(0.39,-0.39)$);
  \end{tikzpicture}\end{subfigure}\par\vspace{2pt}
  \begin{minipage}[c]{0.03\textwidth}\centering\rotatebox{90}{\scriptsize\textbf{Poisson ($I_0\!=\!10^4$})}\end{minipage}\hfill
  \begin{subfigure}[c]{\cw}\centering%
  \begin{tikzpicture}[spy using outlines={circle,orange,line width=0.5pt,magnification=3,size=0.7327cm, connect spies}]
    \node[name=img, inner sep=0pt, line width=0.05mm, draw=white] at (0,0) {\includegraphics[width=\textwidth]{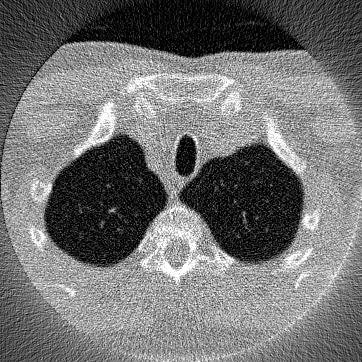}};
    \node[anchor=south west, draw=yellow, line width=0.35pt, fill=black, fill opacity=0.55, text=yellow, text opacity=1, inner sep=1.2pt, font=\tiny] at (img.south west) {20.87\,dB};
    \spy on ($(img.center)+1.181818*(0,0.6)$) in node at ($(img.north west)+(0.39,-0.39)$);
  \end{tikzpicture}\end{subfigure}\hfill
  \begin{subfigure}[c]{\cw}\centering%
  \begin{tikzpicture}[spy using outlines={circle,orange,line width=0.5pt,magnification=3,size=0.7327cm, connect spies}]
    \node[name=img, inner sep=0pt, line width=0.05mm, draw=white] at (0,0) {\includegraphics[width=\textwidth]{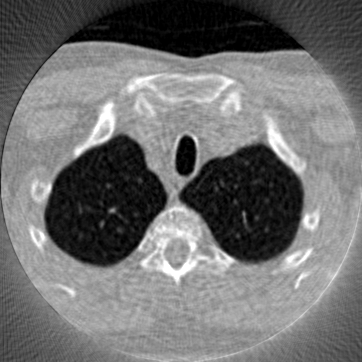}};
    \node[anchor=south west, draw=yellow, line width=0.35pt, fill=black, fill opacity=0.55, text=yellow, text opacity=1, inner sep=1.2pt, font=\tiny] at (img.south west) {32.08\,dB};
    \spy on ($(img.center)+1.181818*(0,0.6)$) in node at ($(img.north west)+(0.39,-0.39)$);
  \end{tikzpicture}\end{subfigure}\hfill
  \begin{subfigure}[c]{\cw}\centering%
  \begin{tikzpicture}[spy using outlines={circle,orange,line width=0.5pt,magnification=3,size=0.7327cm, connect spies}]
    \node[name=img, inner sep=0pt, line width=0.05mm, draw=white] at (0,0) {\includegraphics[width=\textwidth]{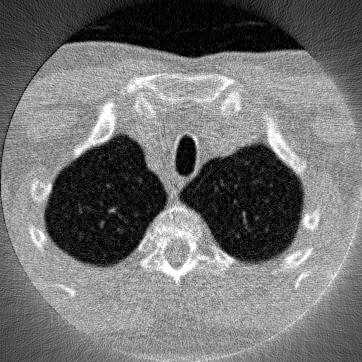}};
    \node[anchor=south west, draw=yellow, line width=0.35pt, fill=black, fill opacity=0.55, text=yellow, text opacity=1, inner sep=1.2pt, font=\tiny] at (img.south west) {26.25\,dB};
    \spy on ($(img.center)+1.181818*(0,0.6)$) in node at ($(img.north west)+(0.39,-0.39)$);
  \end{tikzpicture}\end{subfigure}\hfill
  \begin{subfigure}[c]{\cw}\centering%
  \begin{tikzpicture}[spy using outlines={circle,orange,line width=0.5pt,magnification=3,size=0.7327cm, connect spies}]
    \node[name=img, inner sep=0pt, line width=0.05mm, draw=white] at (0,0) {\includegraphics[width=\textwidth]{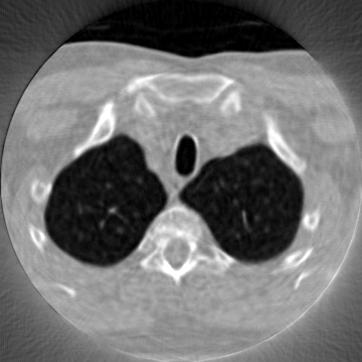}};
    \node[anchor=south west, draw=yellow, line width=0.35pt, fill=black, fill opacity=0.55, text=yellow, text opacity=1, inner sep=1.2pt, font=\tiny] at (img.south west) {\textbf{33.95\,dB}};
    \spy on ($(img.center)+1.181818*(0,0.6)$) in node at ($(img.north west)+(0.39,-0.39)$);
  \end{tikzpicture}\end{subfigure}\hfill
  \begin{subfigure}[c]{\cw}\centering%
  \begin{tikzpicture}[spy using outlines={circle,orange,line width=0.5pt,magnification=3,size=0.7327cm, connect spies}]
    \node[name=img, inner sep=0pt, line width=0.05mm, draw=white] at (0,0) {\includegraphics[width=\textwidth]{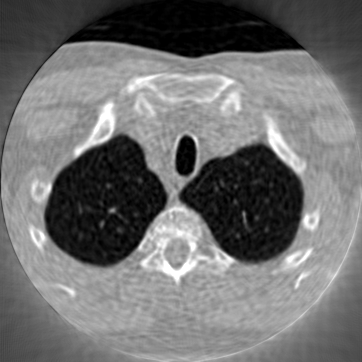}};
    \node[anchor=south west, draw=yellow, line width=0.35pt, fill=black, fill opacity=0.55, text=yellow, text opacity=1, inner sep=1.2pt, font=\tiny] at (img.south west) {33.33\,dB};
    \spy on ($(img.center)+1.181818*(0,0.6)$) in node at ($(img.north west)+(0.39,-0.39)$);
  \end{tikzpicture}\end{subfigure}\hfill
  \begin{subfigure}[c]{\cw}\centering%
  \begin{tikzpicture}[spy using outlines={circle,orange,line width=0.5pt,magnification=3,size=0.7327cm, connect spies}]
    \node[name=img, inner sep=0pt, line width=0.05mm, draw=white] at (0,0) {\includegraphics[width=\textwidth]{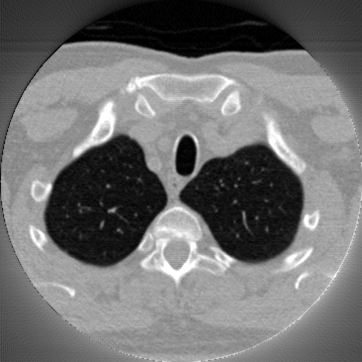}};
    \spy on ($(img.center)+1.181818*(0,0.6)$) in node at ($(img.north west)+(0.39,-0.39)$);
  \end{tikzpicture}\end{subfigure}\par\vspace{2pt}
  \makebox[0.03\textwidth]{}\hfill
  \makebox[\cw]{\scriptsize\shortstack{FBP input\\($128$ noisy views)}}\hfill
  \makebox[\cw]{\scriptsize\shortstack{only predicted \\ views}}\hfill
  \makebox[\cw]{\scriptsize\shortstack{predicted $+$ \\ measured views}}\hfill
  \makebox[\cw]{\scriptsize\shortstack{predicted $+$ \\ blind-spot (BS)}}\hfill
  \makebox[\cw]{\scriptsize\shortstack{predicted $+$ \\ complementary}}\hfill
  \makebox[\cw]{\scriptsize\shortstack{Full-view FBP\\($256$, noise-free)}}
  \caption{\rev{\textbf{Zero-shot denoising under strong noise} (Appendix~\ref{sec: denoising_strong}). As Figure~\ref{fig: noise_denoising}, at $35$\,dB Gaussian noise and $I_0 = 10^4$ photons per bin, the strongest noise levels of Table~\ref{tab: denoising_strong}. The same TomoTransformer, pretrained on noiseless data, is applied with no retraining. Boxes report PSNR against the noise-free $256$-view FBP (right column).}}
  \label{fig: noise_denoising_strong}
\end{figure*}

\subsection{More experiments on real experimental data}
\label{sec: more_experiments_real}

\paragraph{Fine-tuning details:}
For the real scans of Section~\ref{sec: real data}, we fine-tune the pre-trained TomoTransformer with the objective of Section~\ref{sec: ssl}, using only the measured views of that scan: $3{,}000$ steps of AdamW with learning rate $10^{-5}$, linear warm-up and cosine decay, and $16$ patches per step. The fine-tuning uses all slices that are evaluated ($448$ for the brain, the $24$ structured slices for the chip). We report the model at the last step; neither the reference nor any held-out data is used for model selection. Figure~\ref{fig: ssl_curves} shows the image quality along the fine-tuning.

\begin{figure}[t]\normalcolor %
  \centering
  \includegraphics[width=\linewidth]{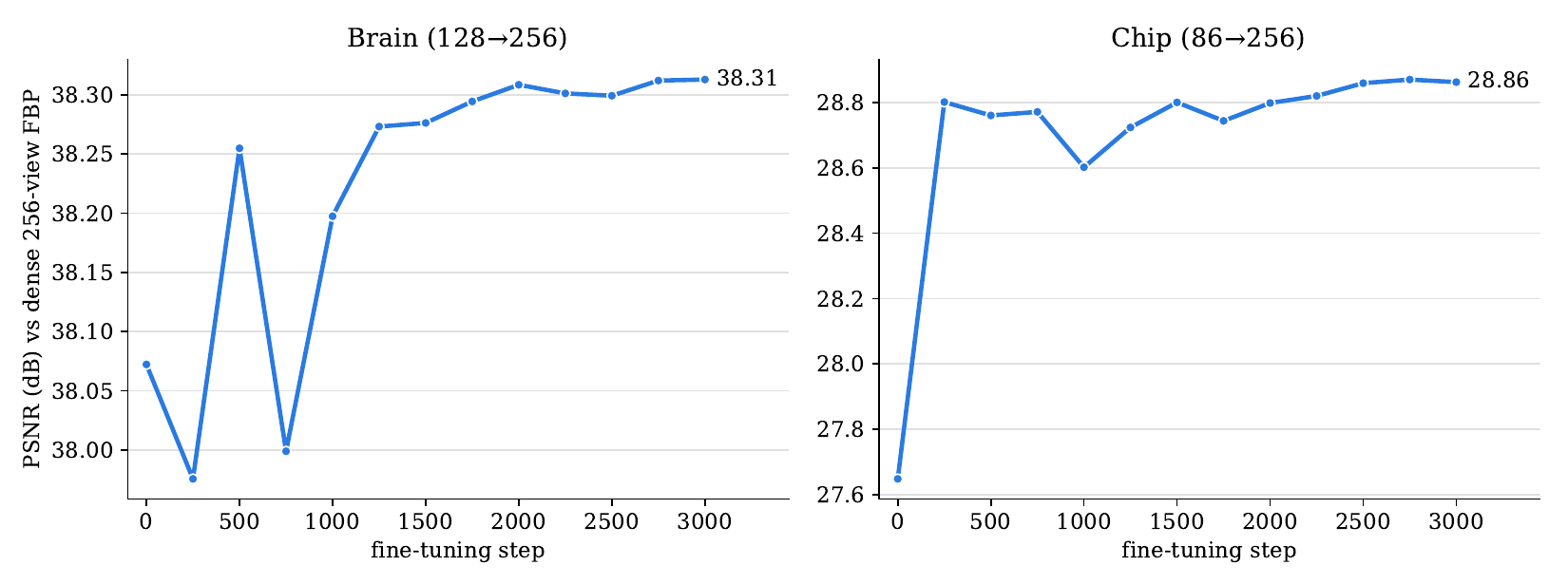}
  \caption{\rev{Image quality along the self-supervised fine-tuning (Section~\ref{sec: ssl}), brain $128\!\rightarrow\!256$ and chip $86\!\rightarrow\!256$, PSNR against the dense $256$-view FBP, evaluated every $250$ steps; step $0$ is the pre-trained (zero-shot) model.}}
  \label{fig: ssl_curves}
\end{figure}

\paragraph{Fine-tuning results:}
Here we provide additional experimental results on the two real PXCT scans of Section~\ref{sec: real data}, the brain and the chip. Table~\ref{tab: ssl_real} reports the results against the dense $512$-view FBP, for completions of the same measured views to $256$ and to $512$ views. Fine-tuning improves on the zero-shot model in every setting, much more on the chip than on the brain ($+1.41$ vs.\ $+0.35$\,dB at ${\rightarrow}256$), and ray-consistent stitching (Section~\ref{sec: sino_aggregation}) adds a further $0.08$--$0.21$\,dB on top. On the brain, completing to $512$ instead of $256$ views raises the PSNR from $36.29$ to $37.85$\,dB. On the chip, it raises SSIM ($0.798 \to 0.815$) but not PSNR ($28.35 \to 28.06$\,dB). Figure~\ref{fig: brain_real} shows the effect of querying more views from the same measurements.

\begin{table}[t]\normalcolor %
\caption{\rev{Self-supervised test-time fine-tuning on two real experimental data of PXCT scans (PSNR in dB\,/\,SSIM, mean over slices). Every row is scored against the dense $512$-view FBP of the same measurements. \emph{Zero-shot} is the pre-trained model; \emph{SSL} is the same model fine-tuned on the measured views of that scan alone; \emph{SSL + RCS} replaces its patch-wise averaging by RCS. The FBP column is the FBP of the measured views that were given to the model. Best PSNR and SSIM per row in bold.}}
\centering
\small
\renewcommand\arraystretch{1.25}\setlength{\tabcolsep}{4pt}
\begin{tabular}{ll|c|c|c|c}
\hline
Sample & Views & FBP & Zero-shot & SSL & SSL + RCS \\
\hline
\multirow{2}{*}{Brain, $448$ slices}
 & $128\!\rightarrow\!256$ & 31.34 / 0.687 & 35.78 / 0.837 & 36.13 / 0.845 & \textbf{36.29} / \textbf{0.849} \\
 & $128\!\rightarrow\!512$ & 31.34 / 0.687 & 37.56 / 0.889 & 37.77 / 0.894 & \textbf{37.85} / \textbf{0.894} \\
\hline
\multirow{2}{*}{\shortstack[l]{Chip, structured\\layers, $24$ slices}}
 & $86\!\rightarrow\!256$ & 22.00 / 0.529 & 26.73 / 0.747 & 28.14 / 0.789 & \textbf{28.35} / \textbf{0.798} \\
 & $86\!\rightarrow\!512$ & 22.00 / 0.529 & 26.86 / 0.758 & 27.96 / 0.790 & \textbf{28.06} / \textbf{0.815} \\
\hline
\end{tabular}
\label{tab: ssl_real}
\end{table}

\begin{figure*}[tb]\normalcolor %
  \centering
  \begin{minipage}[c]{0.03\textwidth}\centering\rotatebox{90}{\small\textbf{Brain ($z\!=\!103$)}}\end{minipage}\hfill
  \begin{subfigure}[c]{0.235\textwidth}\centering%
  \begin{tikzpicture}[spy using outlines={circle,orange,line width=0.5pt,magnification=2,size=0.8568cm, connect spies}]
    \node[name=img, inner sep=0pt, line width=0.05mm, draw=white] at (0,0) {\includegraphics[width=\textwidth]{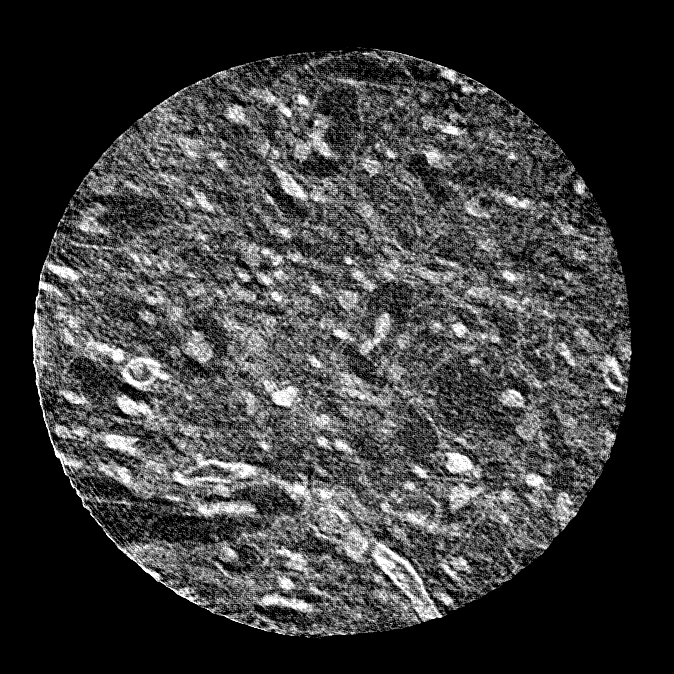}};
    \node[anchor=south west, draw=yellow, line width=0.35pt, fill=black, fill opacity=0.55, text=yellow, text opacity=1, inner sep=1.6pt, font=\footnotesize] at (img.south west) {31.4\,dB};
    \spy on ($(img.center)+1.181818*(-0.10,0.95)$) in node at ($(img.north west)+(0.4432,-0.4432)$);
  \end{tikzpicture}\end{subfigure}\hfill
  \begin{subfigure}[c]{0.235\textwidth}\centering%
  \begin{tikzpicture}[spy using outlines={circle,orange,line width=0.5pt,magnification=2,size=0.8568cm, connect spies}]
    \node[name=img, inner sep=0pt, line width=0.05mm, draw=white] at (0,0) {\includegraphics[width=\textwidth]{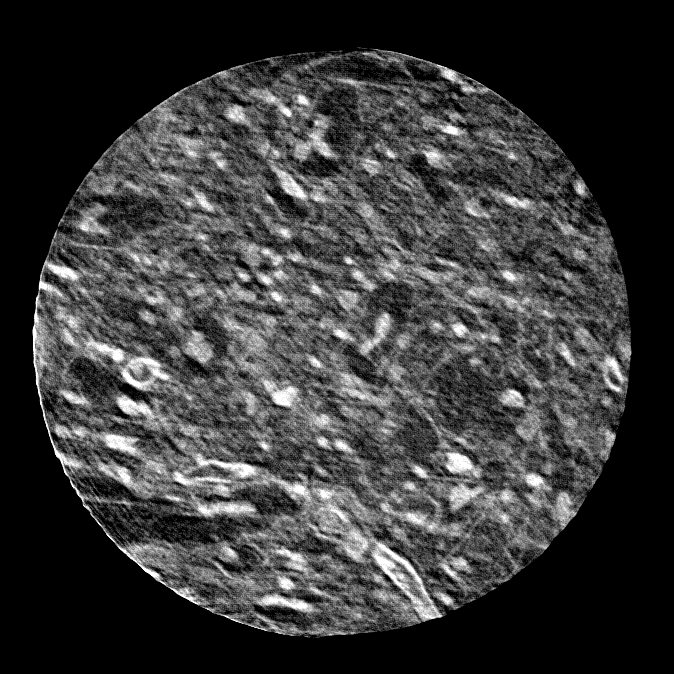}};
    \node[anchor=south west, draw=yellow, line width=0.35pt, fill=black, fill opacity=0.55, text=yellow, text opacity=1, inner sep=1.6pt, font=\footnotesize] at (img.south west) {36.4\,dB};
    \spy on ($(img.center)+1.181818*(-0.10,0.95)$) in node at ($(img.north west)+(0.4432,-0.4432)$);
  \end{tikzpicture}\end{subfigure}\hfill
  \begin{subfigure}[c]{0.235\textwidth}\centering%
  \begin{tikzpicture}[spy using outlines={circle,orange,line width=0.5pt,magnification=2,size=0.8568cm, connect spies}]
    \node[name=img, inner sep=0pt, line width=0.05mm, draw=white] at (0,0) {\includegraphics[width=\textwidth]{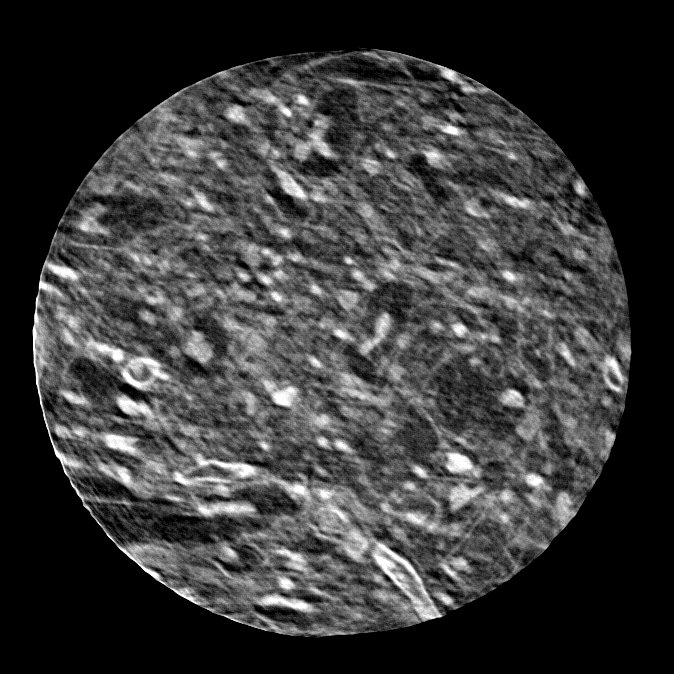}};
    \node[anchor=south west, draw=yellow, line width=0.35pt, fill=black, fill opacity=0.55, text=yellow, text opacity=1, inner sep=1.6pt, font=\footnotesize] at (img.south west) {37.9\,dB};
    \spy on ($(img.center)+1.181818*(-0.10,0.95)$) in node at ($(img.north west)+(0.4432,-0.4432)$);
  \end{tikzpicture}\end{subfigure}\hfill
  \begin{subfigure}[c]{0.235\textwidth}\centering%
  \begin{tikzpicture}[spy using outlines={circle,orange,line width=0.5pt,magnification=2,size=0.8568cm, connect spies}]
    \node[name=img, inner sep=0pt, line width=0.05mm, draw=white] at (0,0) {\includegraphics[width=\textwidth]{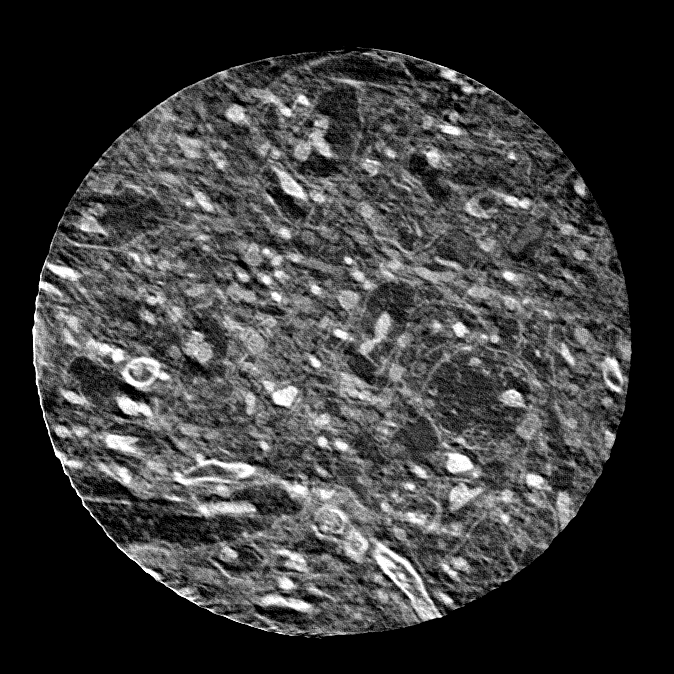}};
    \spy on ($(img.center)+1.181818*(-0.10,0.95)$) in node at ($(img.north west)+(0.4432,-0.4432)$);
  \end{tikzpicture}\end{subfigure}\par\vspace{2pt}
  \begin{minipage}[c]{0.03\textwidth}\centering\rotatebox{90}{\small\textbf{Chip ($z\!=\!435$)}}\end{minipage}\hfill
  \begin{subfigure}[c]{0.235\textwidth}\centering%
  \begin{tikzpicture}[spy using outlines={circle,orange,line width=0.5pt,magnification=3,size=1.2852cm, connect spies}]
    \node[name=img, inner sep=0pt, line width=0.05mm, draw=white] at (0,0) {\includegraphics[width=\textwidth]{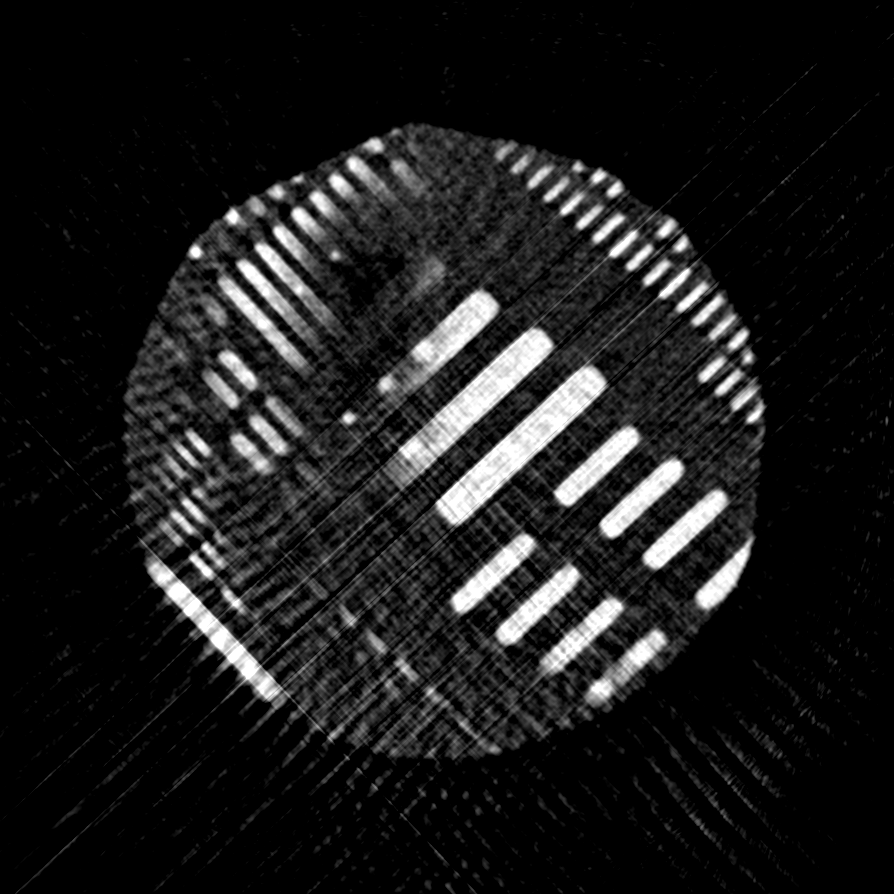}};
    \node[anchor=south west, draw=yellow, line width=0.35pt, fill=black, fill opacity=0.55, text=yellow, text opacity=1, inner sep=1.6pt, font=\footnotesize] at (img.south west) {23.4\,dB};
    \spy on ($(img.center)+1.181818*(0.20,-0.60)$) in node at ($(img.north west)+(0.6648,-0.6648)$);
  \end{tikzpicture}\end{subfigure}\hfill
  \begin{subfigure}[c]{0.235\textwidth}\centering%
  \begin{tikzpicture}[spy using outlines={circle,orange,line width=0.5pt,magnification=3,size=1.2852cm, connect spies}]
    \node[name=img, inner sep=0pt, line width=0.05mm, draw=white] at (0,0) {\includegraphics[width=\textwidth]{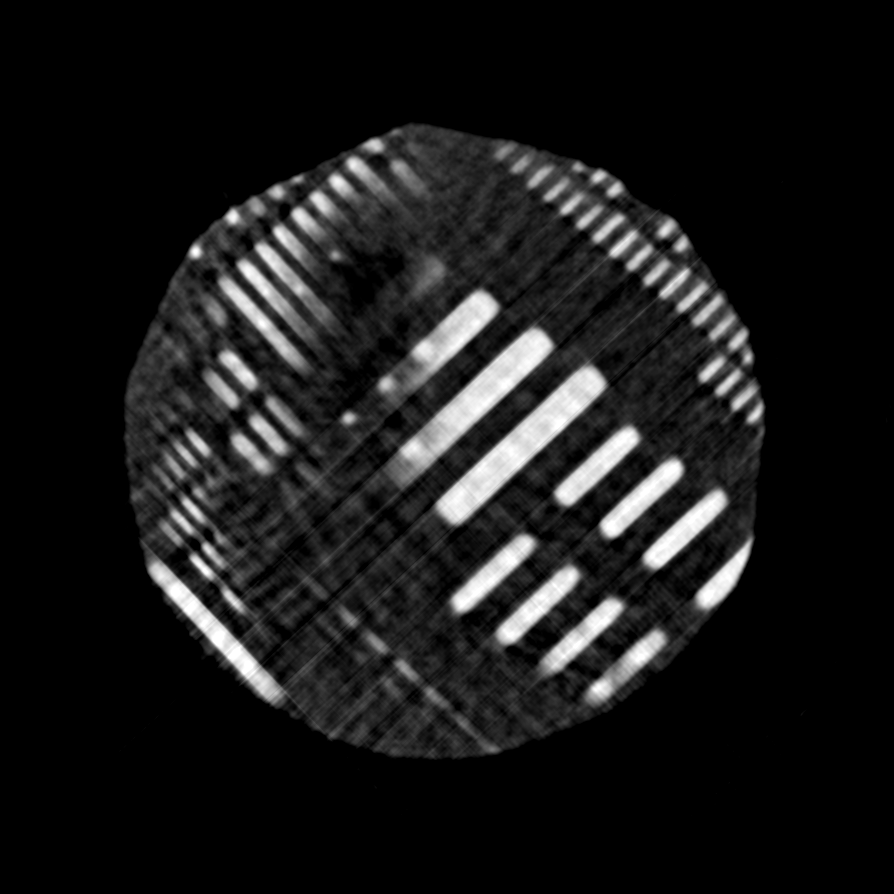}};
    \node[anchor=south west, draw=yellow, line width=0.35pt, fill=black, fill opacity=0.55, text=yellow, text opacity=1, inner sep=1.6pt, font=\footnotesize] at (img.south west) {29.6\,dB};
    \spy on ($(img.center)+1.181818*(0.20,-0.60)$) in node at ($(img.north west)+(0.6648,-0.6648)$);
  \end{tikzpicture}\end{subfigure}\hfill
  \begin{subfigure}[c]{0.235\textwidth}\centering%
  \begin{tikzpicture}[spy using outlines={circle,orange,line width=0.5pt,magnification=3,size=1.2852cm, connect spies}]
    \node[name=img, inner sep=0pt, line width=0.05mm, draw=white] at (0,0) {\includegraphics[width=\textwidth]{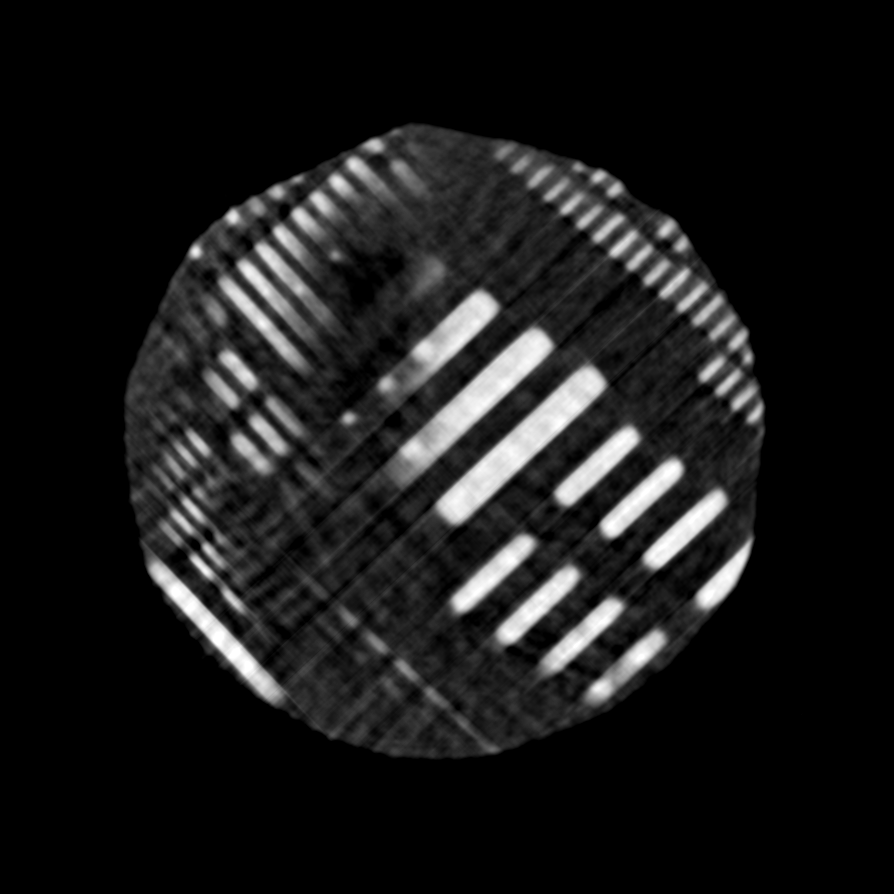}};
    \node[anchor=south west, draw=yellow, line width=0.35pt, fill=black, fill opacity=0.55, text=yellow, text opacity=1, inner sep=1.6pt, font=\footnotesize] at (img.south west) {29.5\,dB};
    \spy on ($(img.center)+1.181818*(0.20,-0.60)$) in node at ($(img.north west)+(0.6648,-0.6648)$);
  \end{tikzpicture}\end{subfigure}\hfill
  \begin{subfigure}[c]{0.235\textwidth}\centering%
  \begin{tikzpicture}[spy using outlines={circle,orange,line width=0.5pt,magnification=3,size=1.2852cm, connect spies}]
    \node[name=img, inner sep=0pt, line width=0.05mm, draw=white] at (0,0) {\includegraphics[width=\textwidth]{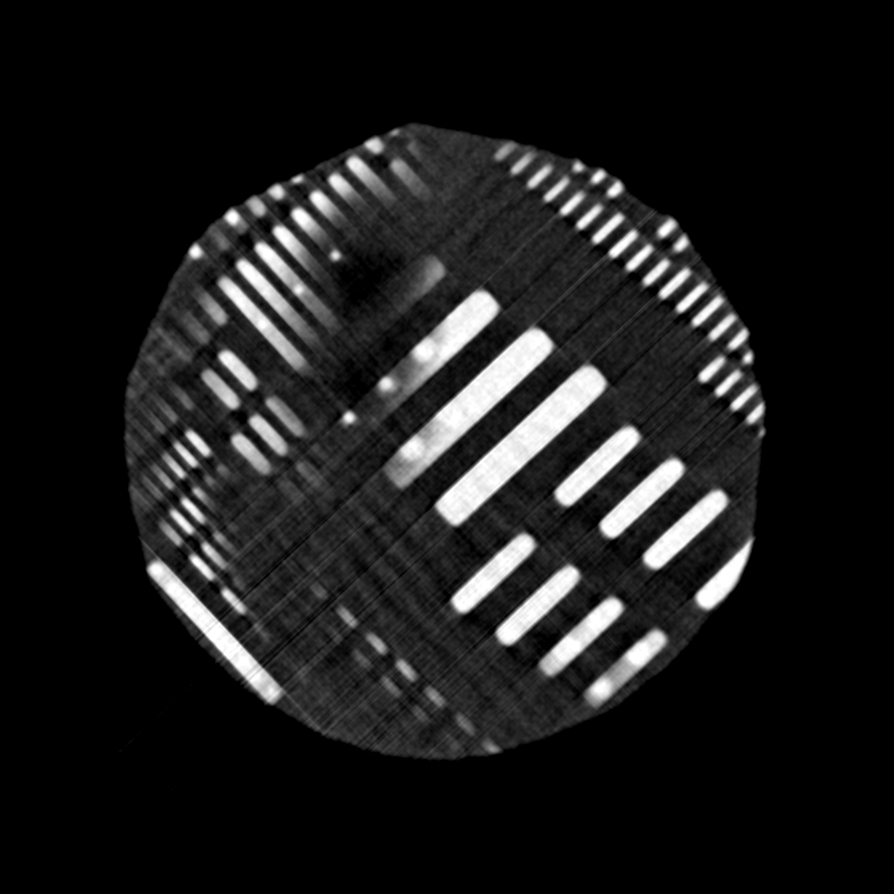}};
    \spy on ($(img.center)+1.181818*(0.20,-0.60)$) in node at ($(img.north west)+(0.6648,-0.6648)$);
  \end{tikzpicture}\end{subfigure}\par\vspace{2pt}
  \makebox[0.03\textwidth]{}\hfill
  \makebox[0.235\textwidth]{\small\shortstack{FBP input\\($128$ / $86$} views)}\hfill
  \makebox[0.235\textwidth]{\small\shortstack{TomoTransformer\\(${\rightarrow}256$)}}\hfill
  \makebox[0.235\textwidth]{\small\shortstack{TomoTransformer\\(${\rightarrow}512$)}}\hfill
  \makebox[0.235\textwidth]{\small\shortstack{Full-view FBP\\($512$)}}
  \caption{\rev{Querying more views from the same measurements, on the two real PXCT scans of Section~\ref{sec: real data}: the brain (top, $128$ measured views) and the chip (bottom, $86$ measured views), same slices as Figure~\ref{fig: ssl_real}. The box in the bottom-left reports \emph{per-slice} PSNR (dB) against the dense $512$-view FBP of the same measurements.}}
  \label{fig: brain_real}
\end{figure*}

\paragraph{Non-uniform angular sampling:}
In Section~\ref{sec: real data}, the measured views of the chip are chosen as close as possible to a uniform grid. The acquisition itself, however, is not uniform~\citep{aidukas2024high}: its angular density varies along the scan and rises sharply around $44^\circ$ and $133^\circ$, up to $29$ views per degree against a median of $8$. Such irregular angular sampling is common in practice, but is not covered by models trained on a fixed grid. Here we keep the acquired angles exactly as they are: we take $256$ of the acquired views, evenly spaced in acquisition order, and use every third one as input ($86$ views, with gaps ranging from $0.54^\circ$ to $3.06^\circ$ instead of a constant $2.1^\circ$; Figure~\ref{fig: nonuniform_real}, left). The pre-trained TomoTransformer, without fine-tuning, predicts the remaining $170$ views, again at their acquired angles. Since the model conditions each token on its actual angle, the non-uniform views need no resampling to a grid. The only change is in the final back-projection, where each view of the completed sinogram is weighted by its angular spacing. The reference is the FBP of all $256$ views with the same weights. Averaged over the $24$ slices, TomoTransformer with RCS reaches $28.3$--$28.5$\,dB, against $24.0$--$24.2$\,dB for the weighted FBP of the $86$ views (ranges over the three choices of which third of the views is measured), close to the $28.8$\,dB it reaches with near-uniform angles. Figure~\ref{fig: nonuniform_real} shows the FBP and TomoTransformer reconstructions. This experiment shows that TomoTransformer can reconstruct a real scan whose angular sampling is not uniform, without retraining and without resampling the data to a uniform grid.

\begin{figure*}[tb]\normalcolor %
  \centering
  \begin{subfigure}[t]{0.188\textwidth}\centering\includegraphics[width=\textwidth]{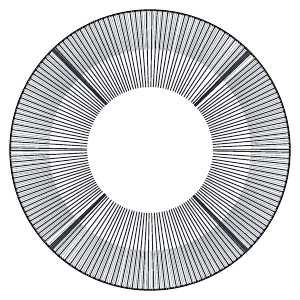}\caption*{Projection angles}\end{subfigure}\hfill
  \begin{subfigure}[t]{0.188\textwidth}\centering%
  \begin{tikzpicture}[spy using outlines={circle,orange,line width=0.5pt,magnification=3,size=1.0282cm, connect spies}]
    \node[name=img, inner sep=0pt, line width=0.05mm, draw=white] at (0,0) {\includegraphics[width=\textwidth]{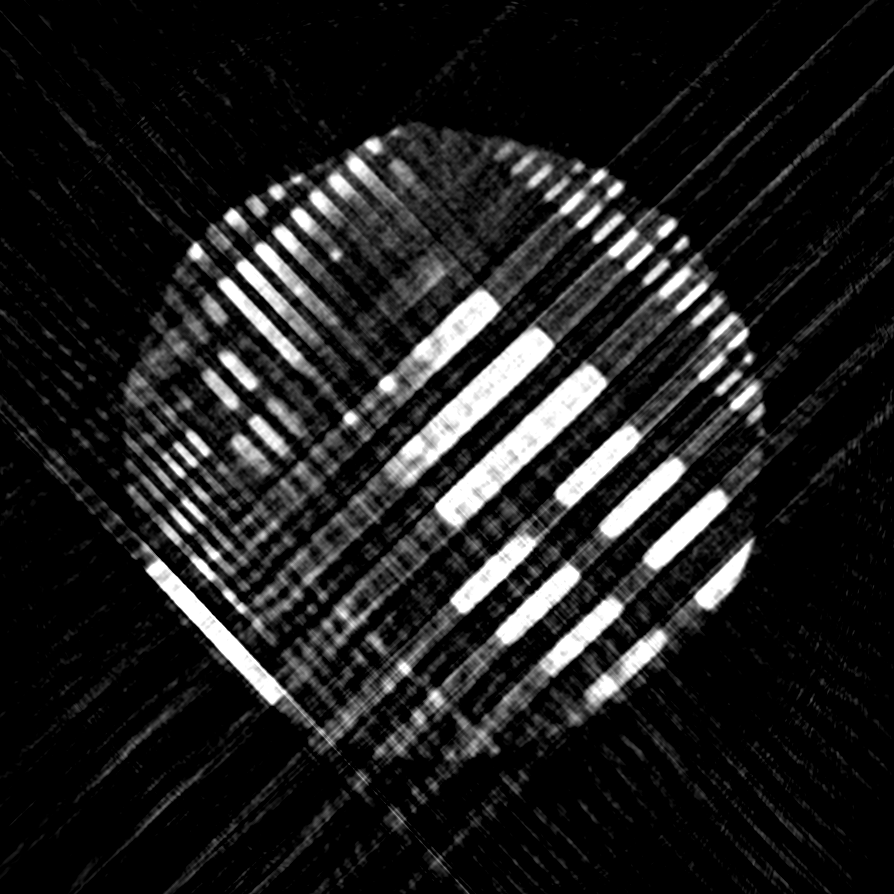}};
    \node[anchor=south west, draw=yellow, line width=0.35pt, fill=black, fill opacity=0.55, text=yellow, text opacity=1, inner sep=1.4pt, font=\scriptsize] at (img.south west) {20.7\,dB};
    \spy on ($(img.center)+1.181818*(0.16,-0.48)$) in node at ($(img.north west)+(0.5318,-0.5318)$);
  \end{tikzpicture}\caption*{FBP ($86$ views)}\end{subfigure}\hfill
  \begin{subfigure}[t]{0.188\textwidth}\centering%
  \begin{tikzpicture}[spy using outlines={circle,orange,line width=0.5pt,magnification=3,size=1.0282cm, connect spies}]
    \node[name=img, inner sep=0pt, line width=0.05mm, draw=white] at (0,0) {\includegraphics[width=\textwidth]{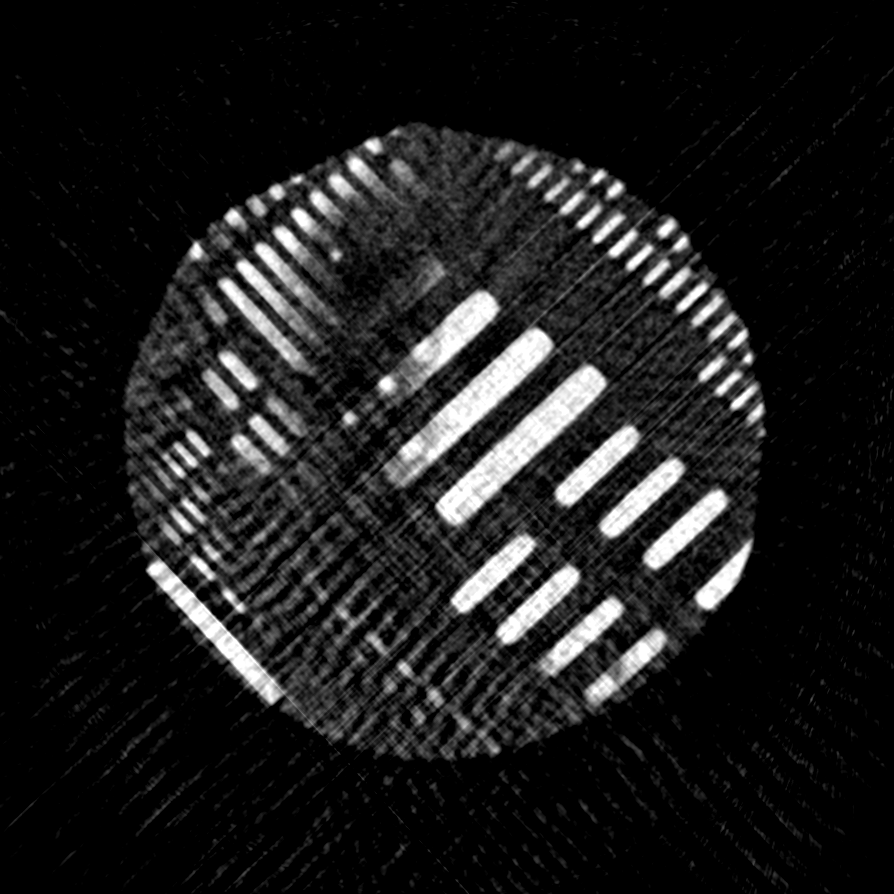}};
    \node[anchor=south west, draw=yellow, line width=0.35pt, fill=black, fill opacity=0.55, text=yellow, text opacity=1, inner sep=1.4pt, font=\scriptsize] at (img.south west) {24.6\,dB};
    \spy on ($(img.center)+1.181818*(0.16,-0.48)$) in node at ($(img.north west)+(0.5318,-0.5318)$);
  \end{tikzpicture}\caption*{Weighted FBP ($86$)}\end{subfigure}\hfill
  \begin{subfigure}[t]{0.188\textwidth}\centering%
  \begin{tikzpicture}[spy using outlines={circle,orange,line width=0.5pt,magnification=3,size=1.0282cm, connect spies}]
    \node[name=img, inner sep=0pt, line width=0.05mm, draw=white] at (0,0) {\includegraphics[width=\textwidth]{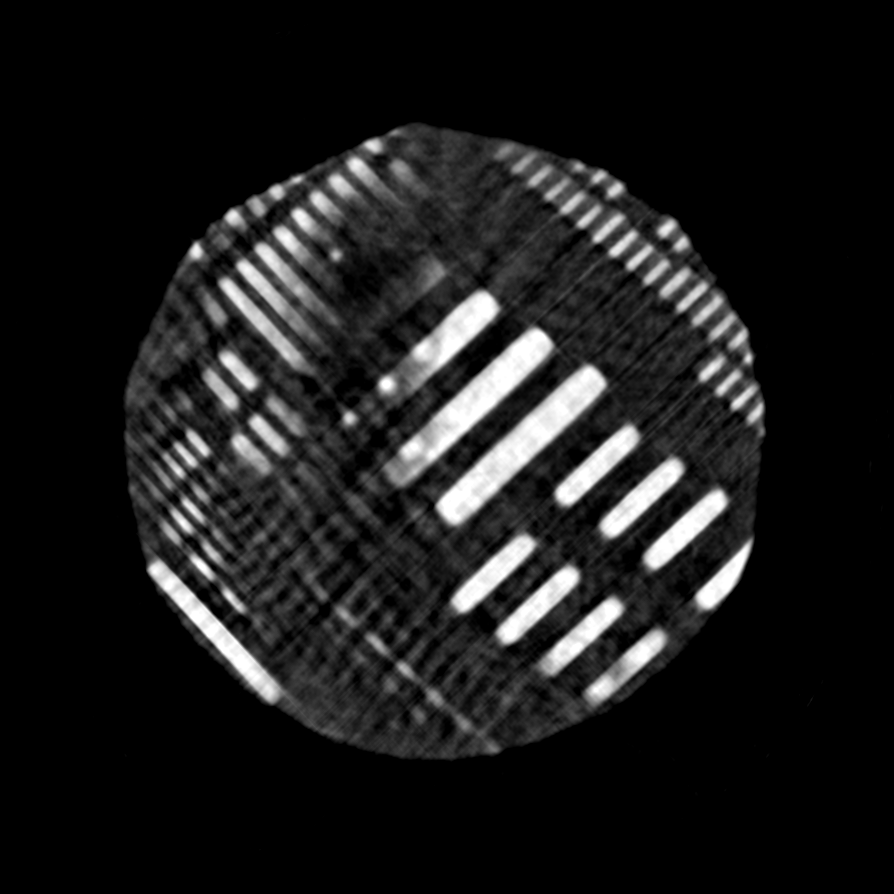}};
    \node[anchor=south west, draw=yellow, line width=0.35pt, fill=black, fill opacity=0.55, text=yellow, text opacity=1, inner sep=1.4pt, font=\scriptsize] at (img.south west) {30.0\,dB};
    \spy on ($(img.center)+1.181818*(0.16,-0.48)$) in node at ($(img.north west)+(0.5318,-0.5318)$);
  \end{tikzpicture}\caption*{TomoTransformer}\end{subfigure}\hfill
  \begin{subfigure}[t]{0.188\textwidth}\centering%
  \begin{tikzpicture}[spy using outlines={circle,orange,line width=0.5pt,magnification=3,size=1.0282cm, connect spies}]
    \node[name=img, inner sep=0pt, line width=0.05mm, draw=white] at (0,0) {\includegraphics[width=\textwidth]{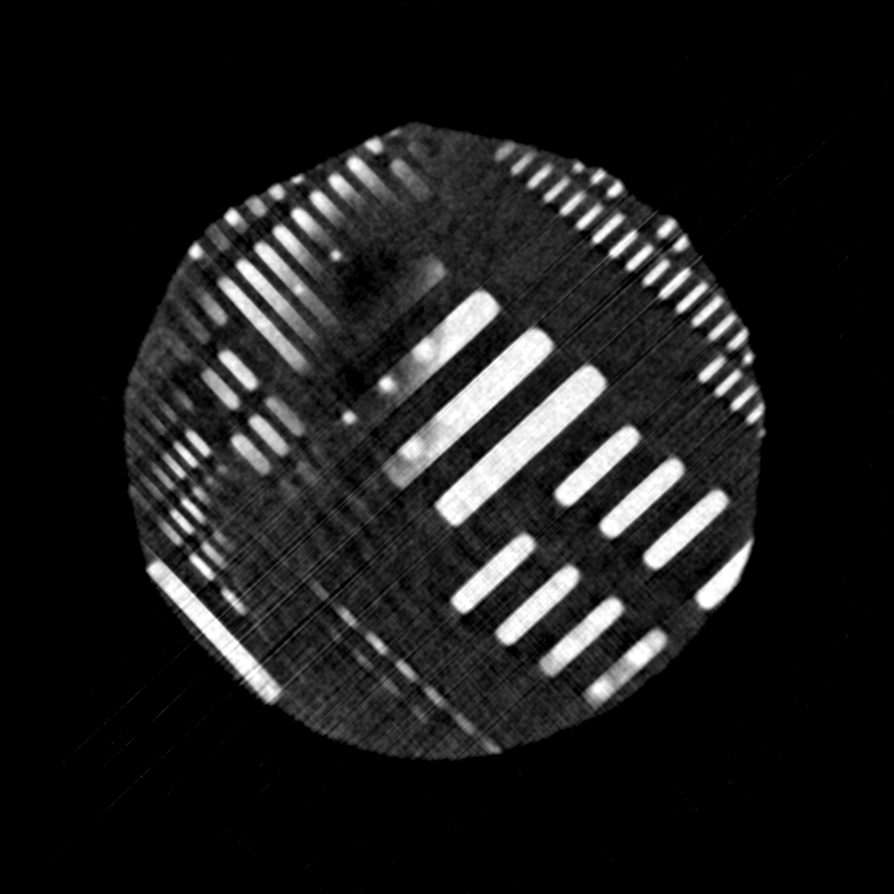}};
    \spy on ($(img.center)+1.181818*(0.16,-0.48)$) in node at ($(img.north west)+(0.5318,-0.5318)$);
  \end{tikzpicture}\caption*{Weighted FBP ($256$)}\end{subfigure}
  \caption{\rev{Non-uniform angular sampling on the real chip scan (same slice as Figure~\ref{fig: ssl_real}), $86\!\rightarrow\!256$ views at the acquired, non-uniformly spaced angles. Left: the $86$ measured views (black) and the $170$ views predicted by TomoTransformer (gray); each line is one view, and the views cluster around $44^\circ$ and $133^\circ$. Weighted FBP weights each view by its angular spacing; TomoTransformer is the pre-trained model without fine-tuning, assembled with RCS (Section~\ref{sec: sino_aggregation}) and back-projected with the same weights. The box in the bottom-left reports per-slice PSNR (dB) against the weighted FBP of all $256$ views (right).}}
  \label{fig: nonuniform_real}
\end{figure*}

\paragraph{Baselines on the brain:}
We also apply the baselines to the real brain projections of Section~\ref{sec: real data}. The pre-trained ViewTrans of Section~\ref{sec: variable_sparsity} cannot be applied to these measurements, since its architecture is tied to the detector size and image resolution it was trained on (Appendix~\ref{sec: viewtrans_details}). The image-domain baselines of Section~\ref{sec: fixed_sparsity} are likewise protocol-bound, but can at least be run in the single setting they were trained for, $64\!\rightarrow\!256$ views, albeit at projection angles that still differ from those seen during training, since the measured views are not uniformly spaced. Figure~\ref{fig: brain_baselines} compares them with TomoTransformer (varied) of Section~\ref{sec: fixed_sparsity}, which has the same number of parameters. TomoTransformer outperforms all baselines, while they hallucinate texture absent from the reference and leave severe blocking artifacts. \rev{With overlapping patches (Figure~\ref{fig: brain_baselines_overlap}; $16$ patches per pixel, Hann-weighted), the blocking artifacts of the baselines disappear and their PSNR rises by $0.2$--$0.4$\,dB, yet TomoTransformer, which uses no overlapping patches thanks to RCS, remains clearly better ($34.97$\,dB over the $24$ slices vs.\ at most $34.14$\,dB for Restormer).}

\begin{figure*}[p]\normalcolor %
  \centering
  \captionsetup[subfigure]{skip=1pt}

 \begin{subfigure}[t]{0.235\textwidth}
  \captionsetup{labelformat=empty,font=sevenpt}
  \centering
  \begin{tikzpicture}[spy using outlines={circle,orange,line width=0.5pt,magnification=3,size=1.1818cm, connect spies}]
    \node[name=img, inner sep=0pt, line width=0.05mm, draw=white] at (0,0) {\includegraphics[width=\textwidth]{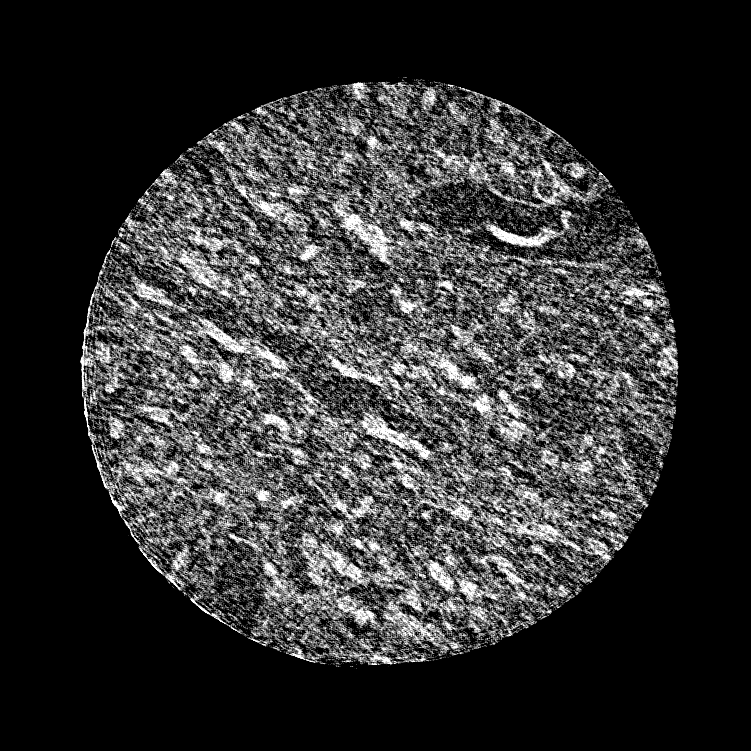}};
    \node[anchor=south west, draw=yellow, line width=0.35pt, fill=black, fill opacity=0.55, text=yellow, text opacity=1, inner sep=1.6pt, font=\footnotesize]
          at ( img.south west) {27.0\,dB};
    \spy on ($(img.center)+(0.6736,0.8864)$) in node at ($(img.north west)+(0.65,-0.65)$);
  \end{tikzpicture}
  \caption*{FBP (64 projs)}
 \end{subfigure}\hfill
 \begin{subfigure}[t]{0.235\textwidth}
  \captionsetup{labelformat=empty,font=sevenpt}
  \centering
  \begin{tikzpicture}[spy using outlines={circle,orange,line width=0.5pt,magnification=3,size=1.1818cm, connect spies}]
    \node[name=img, inner sep=0pt, line width=0.05mm, draw=white] at (0,0) {\includegraphics[width=\textwidth]{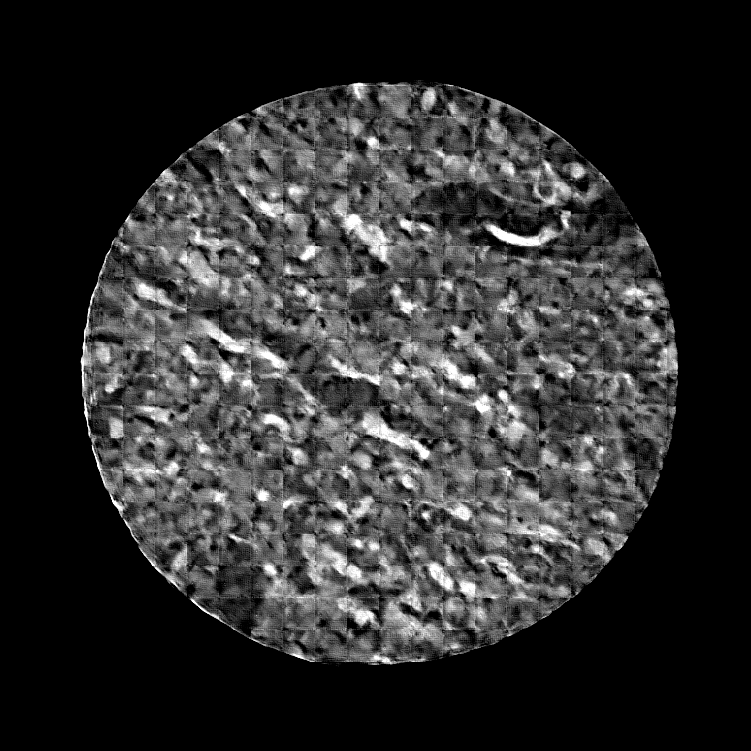}};
    \node[anchor=south west, draw=yellow, line width=0.35pt, fill=black, fill opacity=0.55, text=yellow, text opacity=1, inner sep=1.6pt, font=\footnotesize]
          at ( img.south west) {31.7\,dB};
    \spy on ($(img.center)+(0.6736,0.8864)$) in node at ($(img.north west)+(0.65,-0.65)$);
  \end{tikzpicture}
  \caption*{U-Net}
 \end{subfigure}\hfill
 \begin{subfigure}[t]{0.235\textwidth}
  \captionsetup{labelformat=empty,font=sevenpt}
  \centering
  \begin{tikzpicture}[spy using outlines={circle,orange,line width=0.5pt,magnification=3,size=1.1818cm, connect spies}]
    \node[name=img, inner sep=0pt, line width=0.05mm, draw=white] at (0,0) {\includegraphics[width=\textwidth]{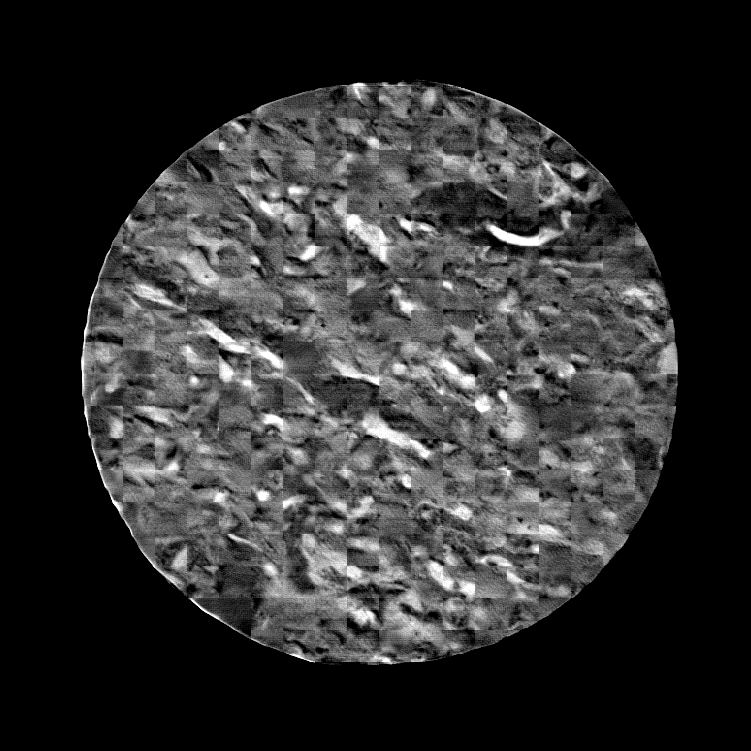}};
    \node[anchor=south west, draw=yellow, line width=0.35pt, fill=black, fill opacity=0.55, text=yellow, text opacity=1, inner sep=1.6pt, font=\footnotesize]
          at ( img.south west) {32.4\,dB};
    \spy on ($(img.center)+(0.6736,0.8864)$) in node at ($(img.north west)+(0.65,-0.65)$);
  \end{tikzpicture}
  \caption*{DRUNet}
 \end{subfigure}\hfill
 \begin{subfigure}[t]{0.235\textwidth}
  \captionsetup{labelformat=empty,font=sevenpt}
  \centering
  \begin{tikzpicture}[spy using outlines={circle,orange,line width=0.5pt,magnification=3,size=1.1818cm, connect spies}]
    \node[name=img, inner sep=0pt, line width=0.05mm, draw=white] at (0,0) {\includegraphics[width=\textwidth]{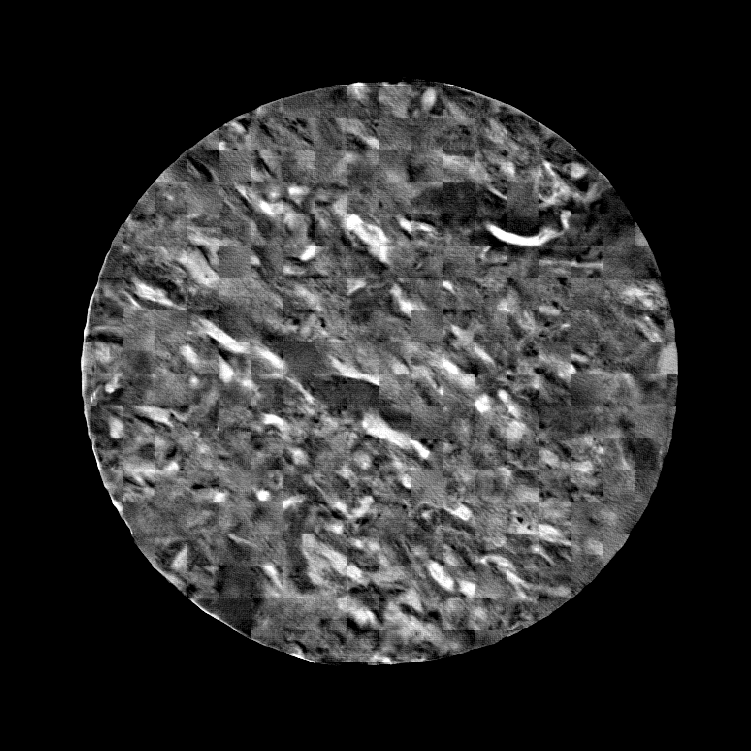}};
    \node[anchor=south west, draw=yellow, line width=0.35pt, fill=black, fill opacity=0.55, text=yellow, text opacity=1, inner sep=1.6pt, font=\footnotesize]
          at ( img.south west) {32.7\,dB};
    \spy on ($(img.center)+(0.6736,0.8864)$) in node at ($(img.north west)+(0.65,-0.65)$);
  \end{tikzpicture}
  \caption*{NAFNet}
 \end{subfigure}\par\vspace{2pt}
 \begin{subfigure}[t]{0.235\textwidth}
  \captionsetup{labelformat=empty,font=sevenpt}
  \centering
  \begin{tikzpicture}[spy using outlines={circle,orange,line width=0.5pt,magnification=3,size=1.1818cm, connect spies}]
    \node[name=img, inner sep=0pt, line width=0.05mm, draw=white] at (0,0) {\includegraphics[width=\textwidth]{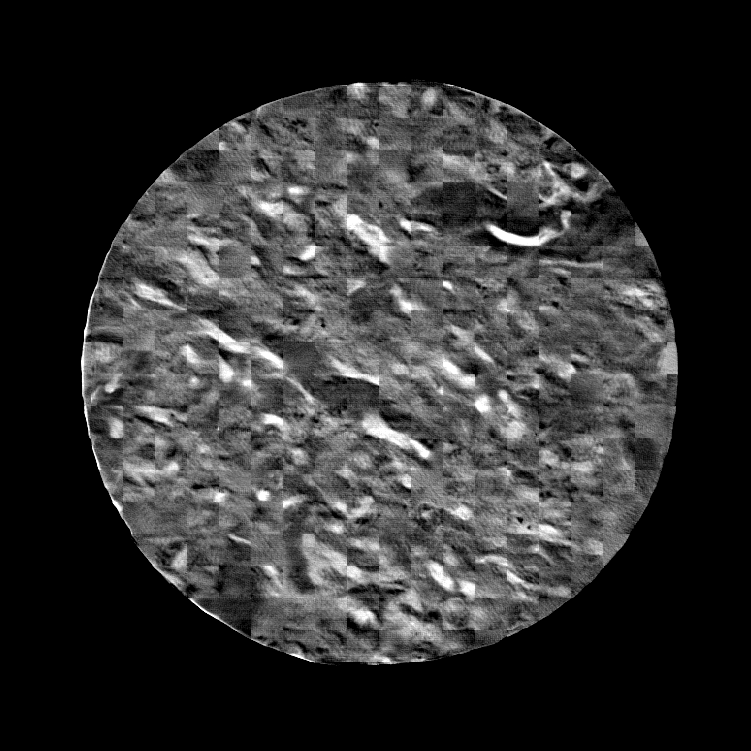}};
    \node[anchor=south west, draw=yellow, line width=0.35pt, fill=black, fill opacity=0.55, text=yellow, text opacity=1, inner sep=1.6pt, font=\footnotesize]
          at ( img.south west) {32.9\,dB};
    \spy on ($(img.center)+(0.6736,0.8864)$) in node at ($(img.north west)+(0.65,-0.65)$);
  \end{tikzpicture}
  \caption*{Restormer}
 \end{subfigure}\hfill
 \begin{subfigure}[t]{0.235\textwidth}
  \captionsetup{labelformat=empty,font=sevenpt}
  \centering
  \begin{tikzpicture}[spy using outlines={circle,orange,line width=0.5pt,magnification=3,size=1.1818cm, connect spies}]
    \node[name=img, inner sep=0pt, line width=0.05mm, draw=white] at (0,0) {\includegraphics[width=\textwidth]{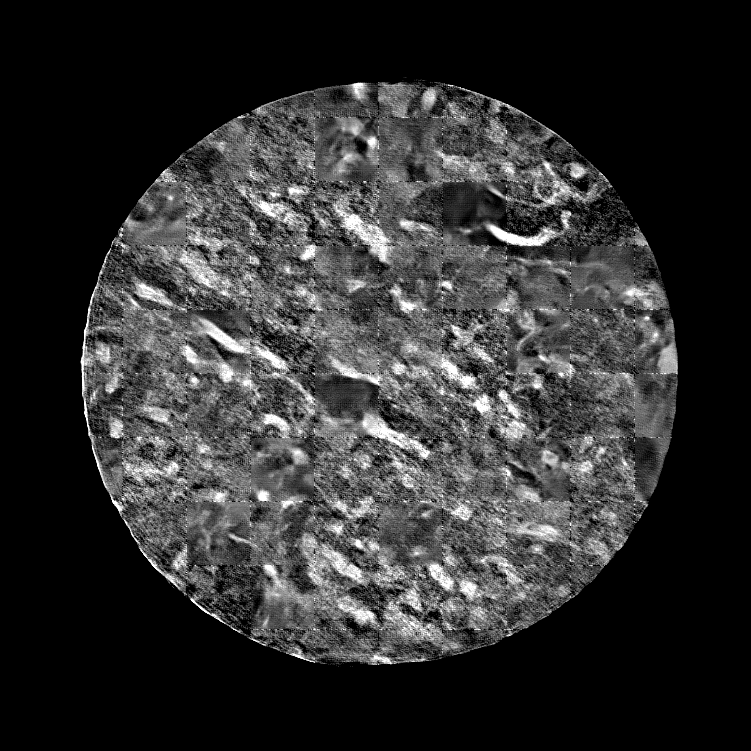}};
    \node[anchor=south west, draw=yellow, line width=0.35pt, fill=black, fill opacity=0.55, text=yellow, text opacity=1, inner sep=1.6pt, font=\footnotesize]
          at ( img.south west) {31.0\,dB};
    \spy on ($(img.center)+(0.6736,0.8864)$) in node at ($(img.north west)+(0.65,-0.65)$);
  \end{tikzpicture}
  \caption*{CTformer}
 \end{subfigure}\hfill
 \begin{subfigure}[t]{0.235\textwidth}
  \captionsetup{labelformat=empty,font=sevenpt}
  \centering
  \begin{tikzpicture}[spy using outlines={circle,orange,line width=0.5pt,magnification=3,size=1.1818cm, connect spies}]
    \node[name=img, inner sep=0pt, line width=0.05mm, draw=white] at (0,0) {\includegraphics[width=\textwidth]{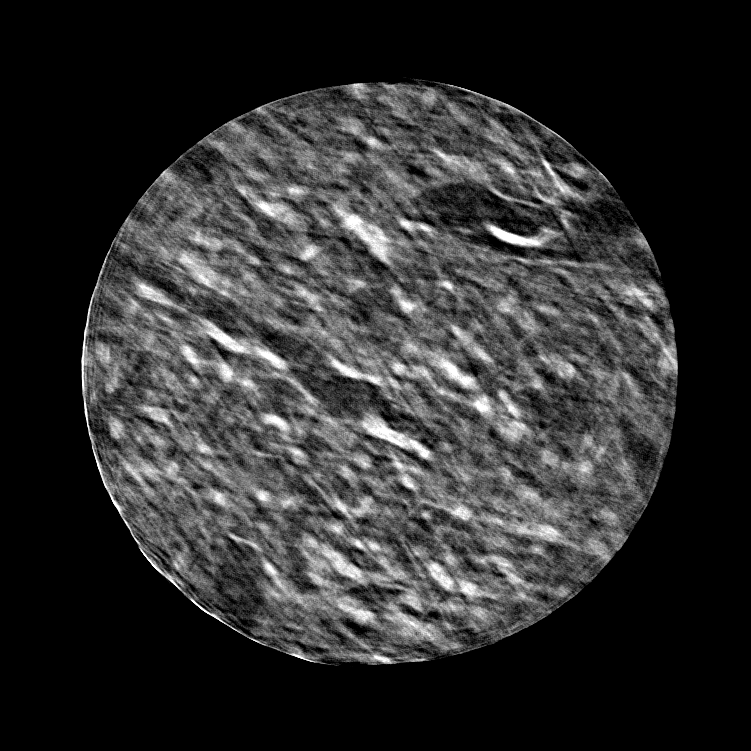}};
    \node[anchor=south west, draw=yellow, line width=0.35pt, fill=black, fill opacity=0.55, text=yellow, text opacity=1, inner sep=1.6pt, font=\footnotesize]
          at ( img.south west) {33.7\,dB};
    \spy on ($(img.center)+(0.6736,0.8864)$) in node at ($(img.north west)+(0.65,-0.65)$);
  \end{tikzpicture}
  \caption*{TomoTransformer \rev{(RCS)}}
 \end{subfigure}\hfill
 \begin{subfigure}[t]{0.235\textwidth}
  \captionsetup{labelformat=empty,font=sevenpt}
  \centering
  \begin{tikzpicture}[spy using outlines={circle,orange,line width=0.5pt,magnification=3,size=1.1818cm, connect spies}]
    \node[name=img, inner sep=0pt, line width=0.05mm, draw=white] at (0,0) {\includegraphics[width=\textwidth]{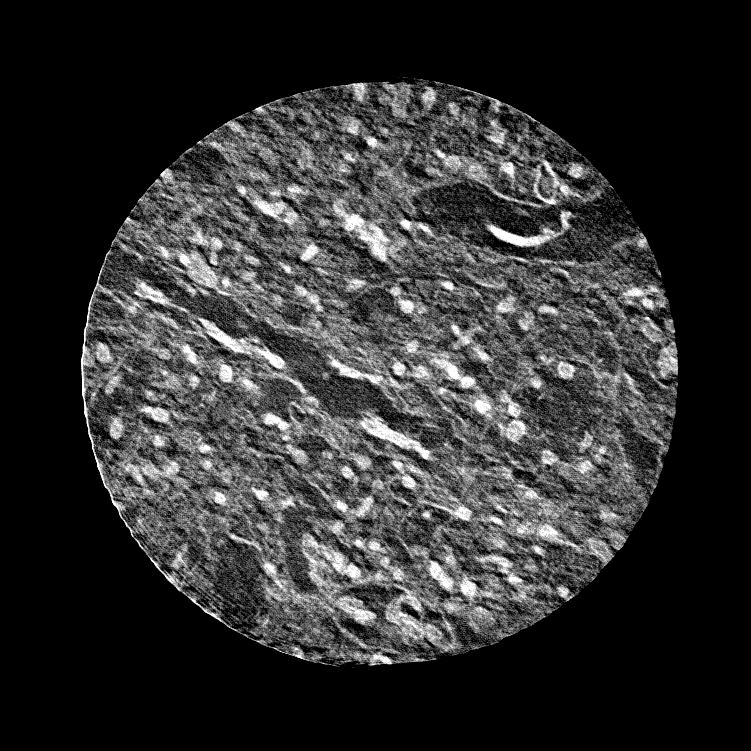}};
    \spy on ($(img.center)+(0.6736,0.8864)$) in node at ($(img.north west)+(0.65,-0.65)$);
  \end{tikzpicture}
  \caption*{FBP (256 projs)}
 \end{subfigure}\hfill

  \caption{Pre-trained single-purpose baselines of Section \ref{sec: fixed_sparsity} applied to the \emph{real experimental} brain projections ($64\!\rightarrow\!256$). Each baseline was trained on one fixed geometry, reconstructing a $256$-view FBP from $64$ \emph{uniformly} distributed views. However, the $617$ real experimental projections are not uniformly spaced. TomoTransformer is trained on varied sparsity patterns and conditions on each token's actual angle, and can conveniently handle this real scenario\rev{; its patch predictions are assembled with ray-consistent stitching (RCS, Section~\ref{sec: sino_aggregation})}. The box in the bottom-left reports PSNR (dB) against the dense $256$-view FBP of the same measurements.}
  \label{fig: brain_baselines}
\end{figure*}

\begin{figure*}[p]\normalcolor %
  \centering
  \captionsetup[subfigure]{skip=1pt}

 \begin{subfigure}[t]{0.235\textwidth}
  \captionsetup{labelformat=empty,font=sevenpt}
  \centering
  \begin{tikzpicture}[spy using outlines={circle,orange,line width=0.5pt,magnification=3,size=1.1818cm, connect spies}]
    \node[name=img, inner sep=0pt, line width=0.05mm, draw=white] at (0,0) {\includegraphics[width=\textwidth]{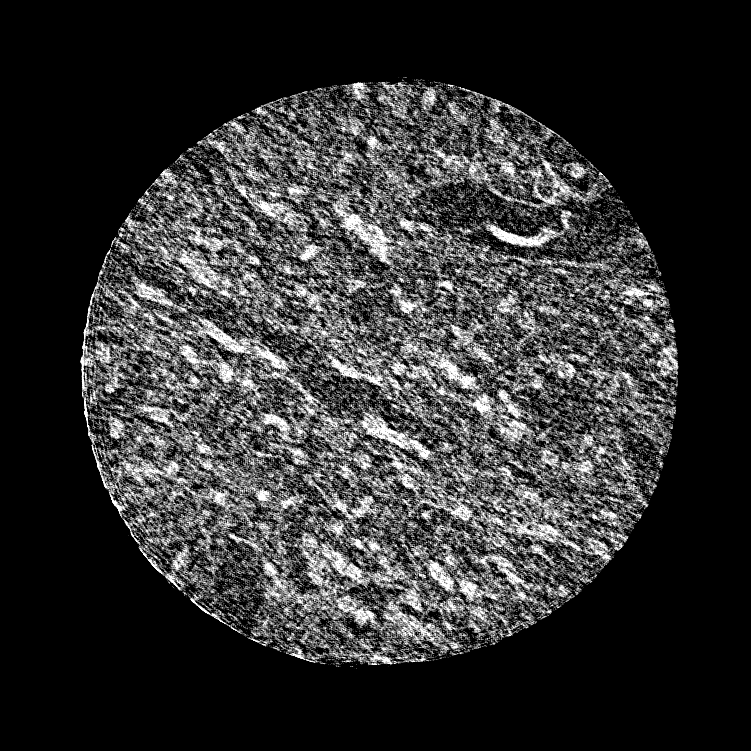}};
    \node[anchor=south west, draw=yellow, line width=0.35pt, fill=black, fill opacity=0.55, text=yellow, text opacity=1, inner sep=1.6pt, font=\footnotesize]
          at ( img.south west) {27.0\,dB};
    \spy on ($(img.center)+(0.6736,0.8864)$) in node at ($(img.north west)+(0.65,-0.65)$);
  \end{tikzpicture}
  \caption*{FBP (64 projs)}
 \end{subfigure}\hfill
 \begin{subfigure}[t]{0.235\textwidth}
  \captionsetup{labelformat=empty,font=sevenpt}
  \centering
  \begin{tikzpicture}[spy using outlines={circle,orange,line width=0.5pt,magnification=3,size=1.1818cm, connect spies}]
    \node[name=img, inner sep=0pt, line width=0.05mm, draw=white] at (0,0) {\includegraphics[width=\textwidth]{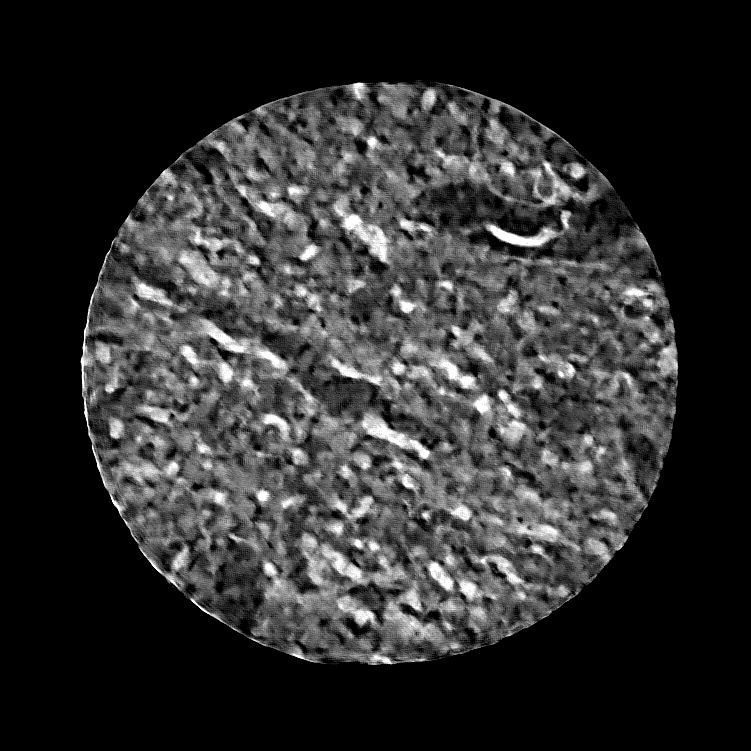}};
    \node[anchor=south west, draw=yellow, line width=0.35pt, fill=black, fill opacity=0.55, text=yellow, text opacity=1, inner sep=1.6pt, font=\footnotesize]
          at ( img.south west) {32.0\,dB};
    \spy on ($(img.center)+(0.6736,0.8864)$) in node at ($(img.north west)+(0.65,-0.65)$);
  \end{tikzpicture}
  \caption*{U-Net}
 \end{subfigure}\hfill
 \begin{subfigure}[t]{0.235\textwidth}
  \captionsetup{labelformat=empty,font=sevenpt}
  \centering
  \begin{tikzpicture}[spy using outlines={circle,orange,line width=0.5pt,magnification=3,size=1.1818cm, connect spies}]
    \node[name=img, inner sep=0pt, line width=0.05mm, draw=white] at (0,0) {\includegraphics[width=\textwidth]{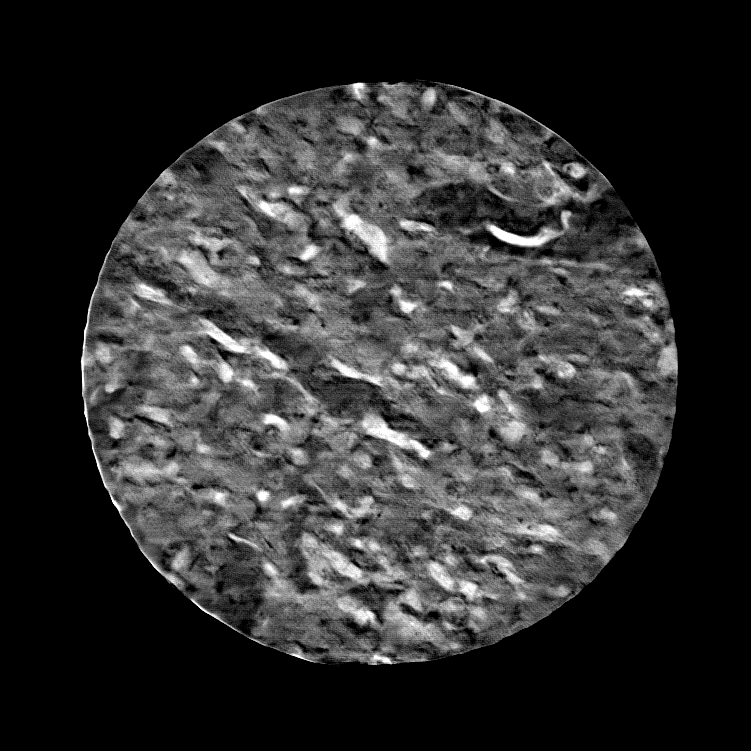}};
    \node[anchor=south west, draw=yellow, line width=0.35pt, fill=black, fill opacity=0.55, text=yellow, text opacity=1, inner sep=1.6pt, font=\footnotesize]
          at ( img.south west) {32.8\,dB};
    \spy on ($(img.center)+(0.6736,0.8864)$) in node at ($(img.north west)+(0.65,-0.65)$);
  \end{tikzpicture}
  \caption*{DRUNet}
 \end{subfigure}\hfill
 \begin{subfigure}[t]{0.235\textwidth}
  \captionsetup{labelformat=empty,font=sevenpt}
  \centering
  \begin{tikzpicture}[spy using outlines={circle,orange,line width=0.5pt,magnification=3,size=1.1818cm, connect spies}]
    \node[name=img, inner sep=0pt, line width=0.05mm, draw=white] at (0,0) {\includegraphics[width=\textwidth]{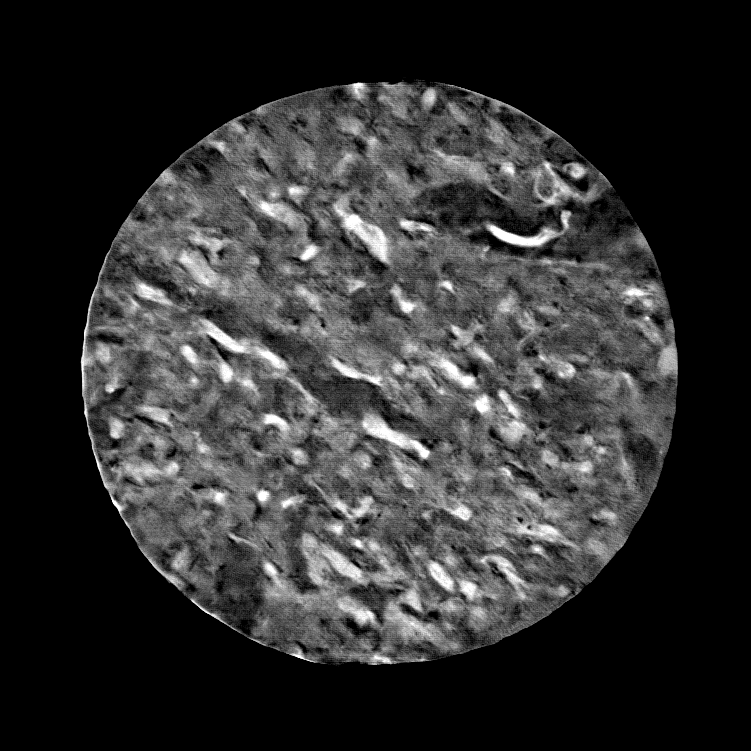}};
    \node[anchor=south west, draw=yellow, line width=0.35pt, fill=black, fill opacity=0.55, text=yellow, text opacity=1, inner sep=1.6pt, font=\footnotesize]
          at ( img.south west) {33.1\,dB};
    \spy on ($(img.center)+(0.6736,0.8864)$) in node at ($(img.north west)+(0.65,-0.65)$);
  \end{tikzpicture}
  \caption*{NAFNet}
 \end{subfigure}\par\vspace{2pt}
 \begin{subfigure}[t]{0.235\textwidth}
  \captionsetup{labelformat=empty,font=sevenpt}
  \centering
  \begin{tikzpicture}[spy using outlines={circle,orange,line width=0.5pt,magnification=3,size=1.1818cm, connect spies}]
    \node[name=img, inner sep=0pt, line width=0.05mm, draw=white] at (0,0) {\includegraphics[width=\textwidth]{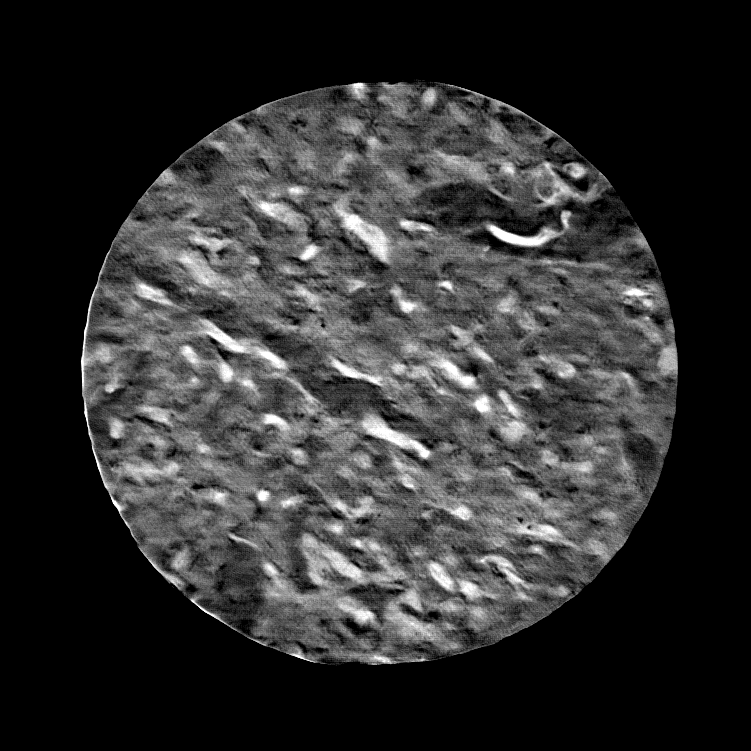}};
    \node[anchor=south west, draw=yellow, line width=0.35pt, fill=black, fill opacity=0.55, text=yellow, text opacity=1, inner sep=1.6pt, font=\footnotesize]
          at ( img.south west) {33.1\,dB};
    \spy on ($(img.center)+(0.6736,0.8864)$) in node at ($(img.north west)+(0.65,-0.65)$);
  \end{tikzpicture}
  \caption*{Restormer}
 \end{subfigure}\hfill
 \begin{subfigure}[t]{0.235\textwidth}
  \captionsetup{labelformat=empty,font=sevenpt}
  \centering
  \begin{tikzpicture}[spy using outlines={circle,orange,line width=0.5pt,magnification=3,size=1.1818cm, connect spies}]
    \node[name=img, inner sep=0pt, line width=0.05mm, draw=white] at (0,0) {\includegraphics[width=\textwidth]{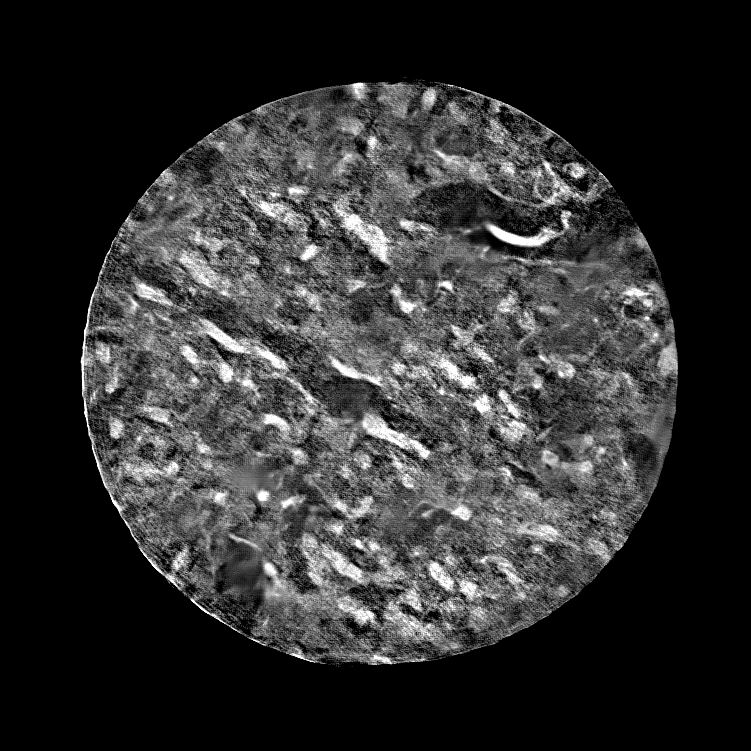}};
    \node[anchor=south west, draw=yellow, line width=0.35pt, fill=black, fill opacity=0.55, text=yellow, text opacity=1, inner sep=1.6pt, font=\footnotesize]
          at ( img.south west) {31.3\,dB};
    \spy on ($(img.center)+(0.6736,0.8864)$) in node at ($(img.north west)+(0.65,-0.65)$);
  \end{tikzpicture}
  \caption*{CTformer}
 \end{subfigure}\hfill
 \begin{subfigure}[t]{0.235\textwidth}
  \captionsetup{labelformat=empty,font=sevenpt}
  \centering
  \begin{tikzpicture}[spy using outlines={circle,orange,line width=0.5pt,magnification=3,size=1.1818cm, connect spies}]
    \node[name=img, inner sep=0pt, line width=0.05mm, draw=white] at (0,0) {\includegraphics[width=\textwidth]{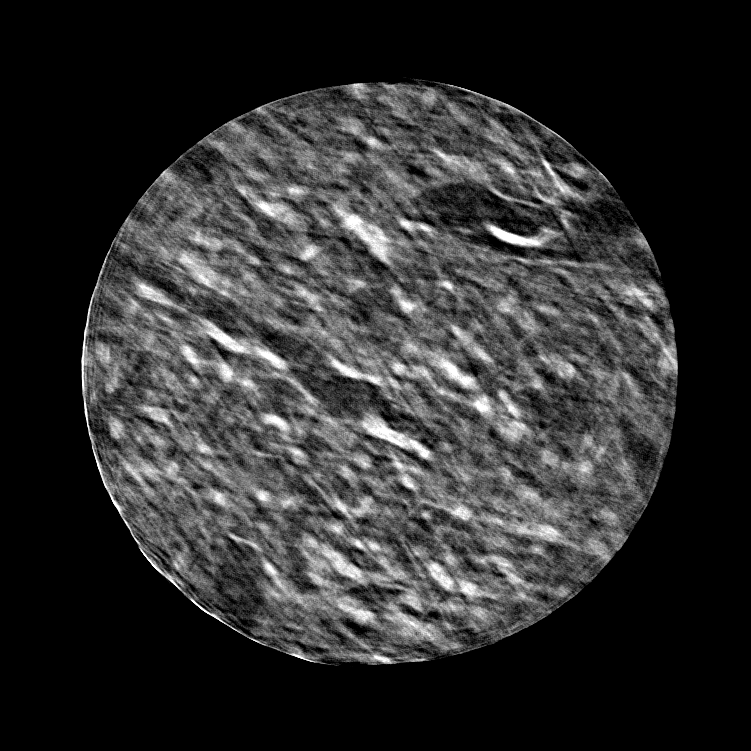}};
    \node[anchor=south west, draw=yellow, line width=0.35pt, fill=black, fill opacity=0.55, text=yellow, text opacity=1, inner sep=1.6pt, font=\footnotesize]
          at ( img.south west) {33.7\,dB};
    \spy on ($(img.center)+(0.6736,0.8864)$) in node at ($(img.north west)+(0.65,-0.65)$);
  \end{tikzpicture}
  \caption*{TomoTransformer (RCS)}
 \end{subfigure}\hfill
 \begin{subfigure}[t]{0.235\textwidth}
  \captionsetup{labelformat=empty,font=sevenpt}
  \centering
  \begin{tikzpicture}[spy using outlines={circle,orange,line width=0.5pt,magnification=3,size=1.1818cm, connect spies}]
    \node[name=img, inner sep=0pt, line width=0.05mm, draw=white] at (0,0) {\includegraphics[width=\textwidth]{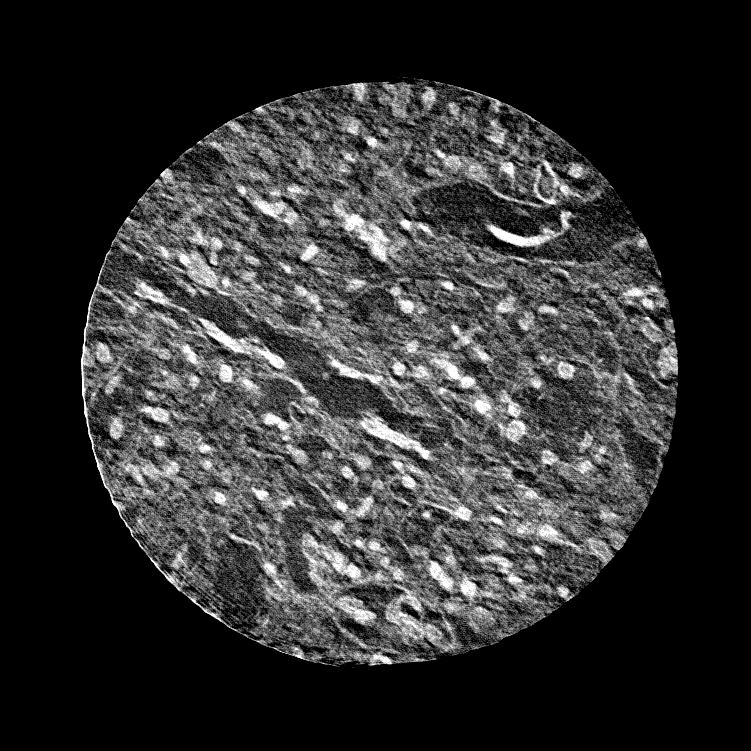}};
    \spy on ($(img.center)+(0.6736,0.8864)$) in node at ($(img.north west)+(0.65,-0.65)$);
  \end{tikzpicture}
  \caption*{FBP (256 projs)}
 \end{subfigure}\hfill

  \caption{\rev{The comparison of Figure~\ref{fig: brain_baselines} with overlapping patches for the image-domain baselines: every pixel is covered by $16$ overlapping patches (stride $8$ for the $32$-px U-Net, DRUNet, NAFNet and Restormer; stride $16$ for the $64$-px CTformer), averaged with a Hann window, which removes their blocking artifacts at the cost of 16$\times$ higher inference time. In contrast, TomoTransformer uses ray-consistent stitching (RCS) without overlapping patches, as shown in Figure~\ref{fig: brain_baselines}. The box in the bottom-left reports PSNR (dB) against the dense $256$-view FBP of the same measurements.}}
  \label{fig: brain_baselines_overlap}
\end{figure*}

\subsection{Additional analysis on HNSCC dataset}
\label{sec: hnscc}
In Table~\ref{tab: fixed_sparsity_psnr}, the gap between TomoTransformer (fixed) and TomoTransformer (varied) at $64\!\rightarrow\!256$ is much larger on HNSCC than on the other datasets; with RCS the gap is $5.11$\,dB on HNSCC, against $0.49$--$1.18$\,dB on the other seven datasets. Here we analyze the cause of this gap and show that it is due to the geometry of the HNSCC volumes. As shown in Figure~\ref{fig: hnscc}, the HNSCC volumes (HNSCC-3DCT-RT, part of CTSpine1K~\citep{deng2021ctspine1k}) are radiotherapy CTs: below the patient lies a long, flat, horizontal couch top and a head-rest, visible in almost all test slices. For projections whose rays are nearly parallel to the couch, its line integrals form a narrow, very bright peak whose position and height change quickly with the projection angle. These missing views are the hardest to predict, and their errors are back-projected as horizontal streaks across the whole slice, including the patient (Figure~\ref{fig: hnscc}, bottom row). While both fixed and varied models have the same architecture and the same training data, they differ only in the measured views sampled during training. The fixed model is trained only on the $64$-of-$256$ sampling of this protocol and can specialize to this difficult geometry, whereas the varied model is trained on varied numbers of measured views. The gap therefore reflects specialization to one protocol: supervised fine-tuning of the varied model on $400$ training slices at $64\!\rightarrow\!256$ (Section~\ref{sec: ssl}, last row of Table~\ref{tab: fixed_sparsity_psnr}) raises its HNSCC PSNR to $43.20$\,dB and removes about four fifths of its additional horizontal error.

\begin{figure*}[tb]\normalcolor %
\centering
\captionsetup[subfigure]{labelformat=empty,font=sevenpt}
\begin{subfigure}[t]{0.245\textwidth}\centering\includegraphics[width=\textwidth]{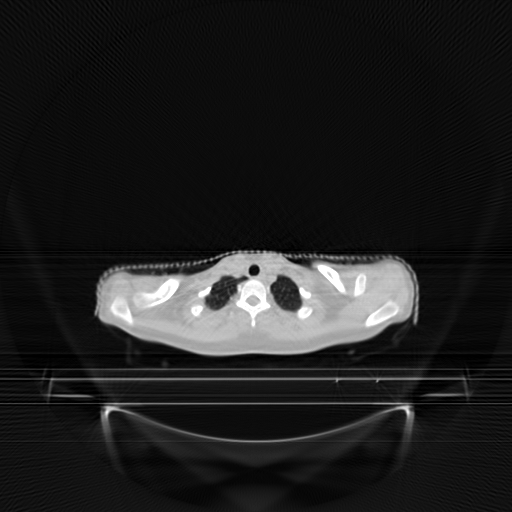}\caption*{Fixed, RCS ($44.33$\,dB)}\end{subfigure}\hfill
\begin{subfigure}[t]{0.245\textwidth}\centering\includegraphics[width=\textwidth]{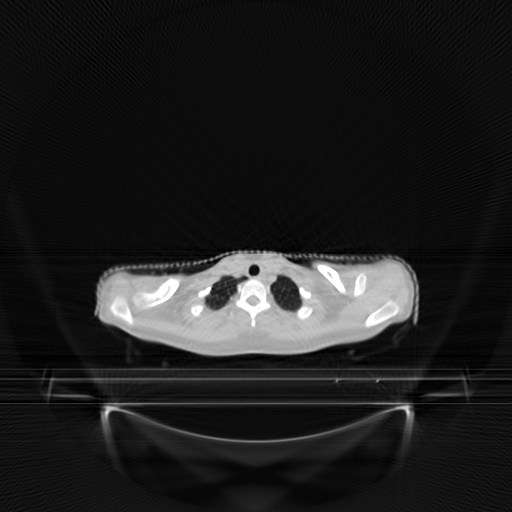}\caption*{Varied, RCS ($39.55$\,dB)}\end{subfigure}\hfill
\begin{subfigure}[t]{0.245\textwidth}\centering\includegraphics[width=\textwidth]{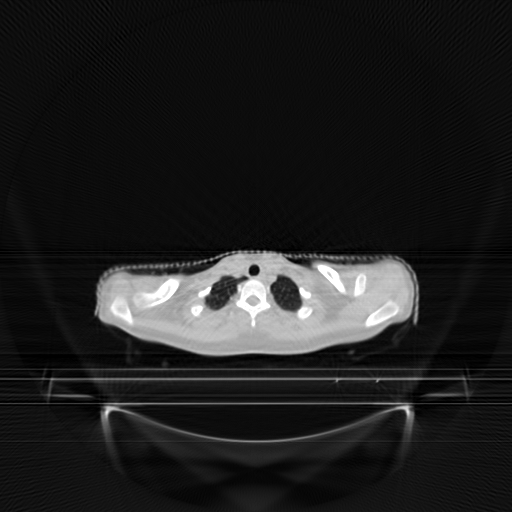}\caption*{Fine-tuned, RCS ($42.85$\,dB)}\end{subfigure}\hfill
\begin{subfigure}[t]{0.245\textwidth}\centering\includegraphics[width=\textwidth]{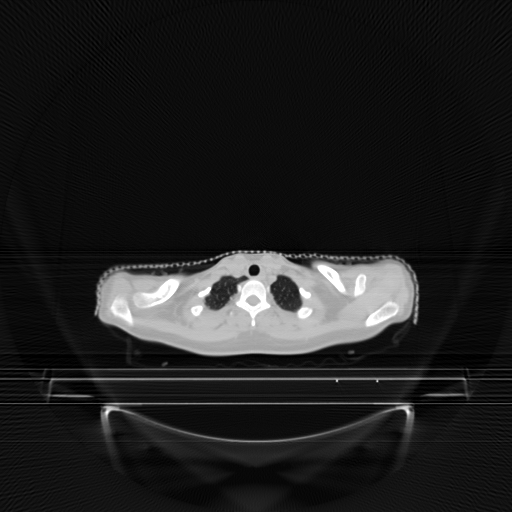}\caption*{Reference ($256$-view FBP)}\end{subfigure}\par\vspace{2pt}
\begin{subfigure}[t]{0.245\textwidth}\centering\includegraphics[width=\textwidth]{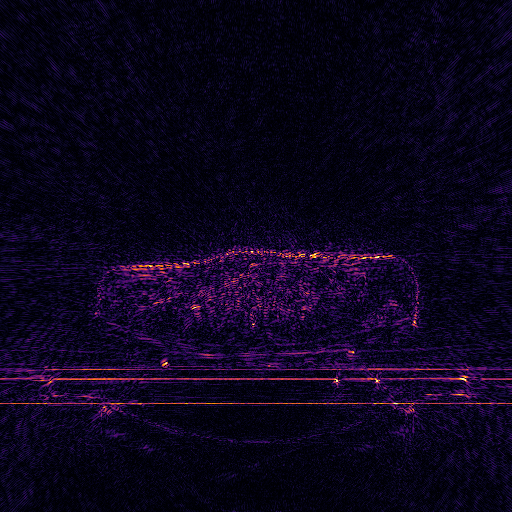}\end{subfigure}\hfill
\begin{subfigure}[t]{0.245\textwidth}\centering\includegraphics[width=\textwidth]{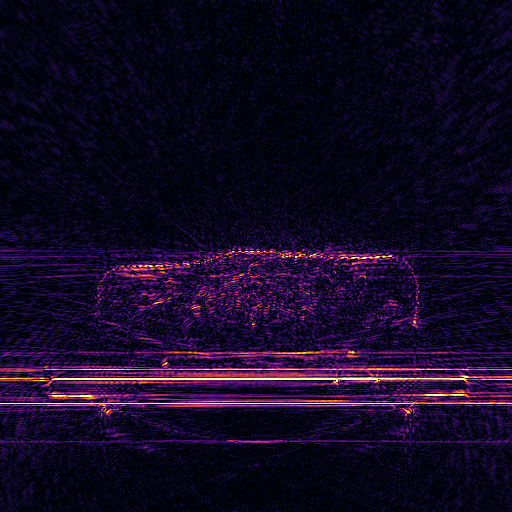}\end{subfigure}\hfill
\begin{subfigure}[t]{0.245\textwidth}\centering\includegraphics[width=\textwidth]{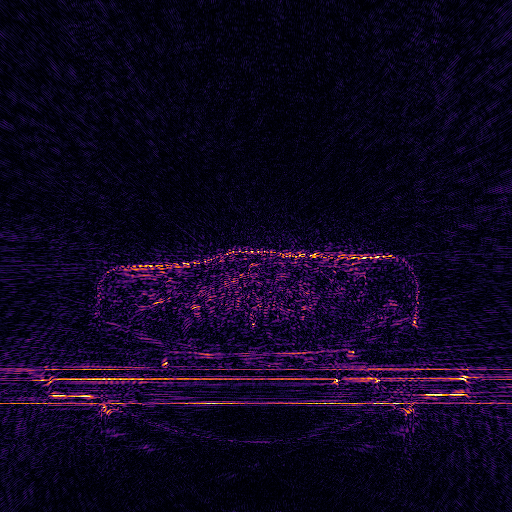}\end{subfigure}\hfill
\begin{subfigure}[t]{0.245\textwidth}\centering\mbox{}\end{subfigure}
\caption{\rev{\textbf{TomoTransformer (fixed) and TomoTransformer (varied) on HNSCC}, $64\!\rightarrow\!256$, both with RCS (Appendix~\ref{sec: hnscc}). Top: TomoTransformer (fixed, RCS), TomoTransformer (varied, RCS), the varied model after supervised fine-tuning, with RCS, and the dense $256$-view FBP reference, on one test slice. Bottom: absolute error against the reference, on a common color scale. The error of the varied model is dominated by horizontal streaks, caused by the views nearly parallel to the couch. PSNR of the full slice in parentheses.}}
\label{fig: hnscc}
\end{figure*}

\end{revblock}

\end{document}